\documentclass[preprint, 12pt]{elsarticle}

\usepackage{lineno,hyperref}
\modulolinenumbers[5]

\journal{Journal of Computational Physics}

\usepackage{amsmath,amssymb}
\usepackage{graphicx}
\usepackage{subfig}

\usepackage{floatrow}
\usepackage{placeins}

\usepackage{caption}

\usepackage{verbatim}
\usepackage{amsthm}
\usepackage{mathtools}

\usepackage{enumerate}
\usepackage{color}

\usepackage{algorithm}%,algcompatible}
\usepackage{algorithmic}
\usepackage{anyfontsize}
\usepackage{pgfplots}
\pgfplotsset{ % Here we specify options for all figures in the document
  compat=1.5, % Which version of pgfplots do we want to use?
  legend style = {font=\scriptsize\sffamily},
  label style = {font=\scriptsize\sffamily},
  tick style = {font=\scriptsize\sffamily},
  every tick label/.append style={font=\scriptsize}
  }
  
\newlength\figureheight 
\newlength\figurewidth 

\graphicspath{{../}}  
  
\usepackage{tikz}
\usepackage{tkz-tab}
\usetikzlibrary{bayesnet}

\newcommand\be{\begin{equation}}
\newcommand\ee{\end{equation}}
\def\RR{ \mathbb R}
\newcommand{\refeqq}[1]{Eq.~(\ref{#1})}
\newcommand{\reffig}[1]{Fig.~\ref{#1}}

\newcommand\bea{\begin{eqnarray*}}
\newcommand\eea{\end{eqnarray*}}

\newcommand{\bd}{\begin{description}}
\newcommand{\ed}{\end{description}}
\newcommand{\bi}{\begin{itemize}}
\newcommand{\ei}{\end{itemize}}
\newcommand{\bb}{\begin{block}}
\newcommand{\eb}{\end{block}}
\newcommand{\pa}{\partial}

\newcommand{\bs}{\boldsymbol}

\newcommand\cg{_\text{c}}
\newcommand\cf{_\text{cf}}

\newcommand\aaa{_\text{f}}

\newcommand\pf{p_\text{f}}
\newcommand\pc{p_\text{c}}
\newcommand\pcf{p_\text{cf}}

\newcommand\calL{\mathcal{L}}

\newcommand\calM{\mathcal{M}}

\newcommand\R{\mathbb{R}}

\newcommand\E{\mathbb{E}}

\newcommand\bq{\boldsymbol{x}}
\newcommand{\bqd}{\boldsymbol{x}^{(1:N)}}

\newcommand{\bqi}{\boldsymbol{x}^{(i)}}

\newcommand\btheta{\boldsymbol{\theta}}

\newcommand{\bQ}{\boldsymbol{X}}

\newcommand{\calR}{\mathcal{R}}

\newcommand{\kl}{\operatorname{KL}}
\newcommand{\bgam}{\boldsymbol{\theta}}

\newcommand\diag{\operatorname{diag}}

\newcommand\Cov{\operatorname{Cov}}

\DeclareMathAlphabet\mathbfcal{OMS}{cmsy}{b}{n}

\begin{document}

\begin{frontmatter}

%\title{Elsevier \LaTeX\ template\tnoteref{mytitlenote}}
\title{{\em Predictive} Coarse-Graining}

% \address{Radarweg 29, Amsterdam}
\author[TUM]{Markus Sch\"oberl}
\ead{m.schoeberl@tum.de}
\author[TUMIAS,ND]{Nicholas Zabaras}
\ead{nzabaras@gmail.com}
\ead[url]{www.zabaras.com}
\author[TUM]{Phaedon-Stelios Koutsourelakis\corref{mycorrespondingauthor}}
\ead{p.s.koutsourelakis@tum.de}
\ead[url]{www.contmech.mw.tum.de}


%\ead[url]{www.contmech.mw.tum.de}

\address[TUM]{Continuum Mechanics Group,\\ Technical University of Munich, Boltzmannstra{\ss}e 15, 85748 Garching, Germany}
\address[TUMIAS]{Institute for Advanced Study,\\ Technical University of Munich, Lichtenbergstra{\ss}e 2a, 85748 Garching, Germany}
\address[ND]{Department of Aerospace and Mechanical Engineering, 
	University of Notre Dame, \\
	365 Fitzpatrick Hall, 
	Notre Dame, IN 46556, USA}

\cortext[mycorrespondingauthor]{Corresponding author}

\begin{abstract}
We propose a data-driven, coarse-graining formulation
in the context of equilibrium statistical mechanics. 
In contrast to existing techniques which  are based on a  fine-to-coarse map, we adopt the opposite strategy by prescribing a {\em probabilistic coarse-to-fine} map. 
This corresponds to a directed probabilistic model where the coarse variables play the role of latent generators of the fine scale (all-atom) data.  
From an information-theoretic perspective, the  framework proposed provides an improvement upon  the  relative entropy method \cite{{shell2008}} and is capable of  quantifying   the  uncertainty due to the information loss that unavoidably takes place during the CG process.
Furthermore, it can be readily extended to a  fully Bayesian model where various sources of uncertainties   are reflected in the posterior of the model parameters.
The latter can be used to produce not only point estimates of fine-scale reconstructions or macroscopic observables, but more importantly, predictive posterior distributions on these quantities. Predictive posterior distributions reflect the confidence of the model as a function of the amount of data and the level of coarse-graining.
The issues of model complexity and  model selection are seamlessly addressed by employing a hierarchical prior that favors  the discovery of sparse solutions, revealing the most
prominent features in the coarse-grained model. 
A flexible and parallelizable Monte Carlo - Expectation-Maximization (MC-EM) scheme is proposed for carrying out inference and learning tasks.
A comparative assessment of  the proposed methodology is presented for  a lattice spin system and the SPC/E water model. 
\end{abstract}

\begin{keyword}
Coarse-Graining, Generative models, Bayesian, Uncertainty quantification,
SPC/E water, Lattice systems
% \texttt{elsarticle.cls}\sep \LaTeX\sep Elsevier \sep template
% \MSC[2010] 00-01\sep  99-00
\end{keyword}

\end{frontmatter}

\linenumbers

 \section{Introduction}

Molecular dynamics simulations~\cite{alder_studies_1959} are nowadays commonplace in physics,  chemistry and engineering and represent one of the most reliable tools in the analysis of complex processes and the design of new materials~\cite{karplus_molecular_2002,buehler_atomistic_2008,kremer2010}.
Direct simulations are hampered by the gigantic number of degrees of freedom, complex,  potentially long-range and  high-order interactions, and as a result,  are limited to small  spatio-temporal scales with current and foreseeable  computational resources. 

% 
% Predicting folding processes, systems must even
% be simulated in the timescale of milliseconds.~\cite{shell2008peptide, shell2009}
% Such complex many-body systems are highly demanding in computation and become often intractable for 
% systems of interest.

An approach towards making complex simulations practicable over  extended time/space  scales is coarse-graining (CG)~\cite{voth_coarse-graining_2008}.
Coarse-graining methods attempt to summarize  the atomistic detail
 in much fewer degrees of freedom which in turn lead to shorter simulation times, with  potentially larger time-steps and enable the analysis of systems that occupy larger spatial domains. Furthermore, from a reductionist's  point of view, they can provide insight into the fundamental components or processes associated with the macroscopic behavior and properties of molecular ensembles.
%  Not only the degrees of freedom are of interest
% but also coarse-graining the complexity or order of interactions
% in the potential would increase the
% efficiency of simulations. Less computational effort enables us to study systems of larger scales.

A systematic strategy towards coarse-graining is offered in the context of free-energy computation methods~\cite{lelievere2010, bilionis_free_2012}. Nevertheless, their  primary goal is  to escape deep, free-energy wells and  are generally limited to a relatively small number of CG variables. A mathematically rigorous approach to coarse-graining  lattice systems and a rich set of multi-level, adaptive  algorithms for equilibrium  and nonequilibrium settings, has been developed   in~\cite{katsoulakis2003,chatterjee_spatially_2004,
katsoulakis_error_2006,katsoulakis_numerical_2008,kalligiannaki_multilevel_2011,
katsoulakis_information-theoretic_2013}. Inversion-based methods such as 
the Direct or Iterative Boltzmann Inversion~\cite{tschop_simulation_1998,Reith2003} and  Inverse Monte Carlo
\cite{PhysRevE.52.3730}, represent a popular strategy  where the parameters of the CG model are adjusted to reproduce macroscopic observables~\cite{rudzinski_generalized-yvon-born-green_2015}.
Molecular Renormalization Group CG~\cite{Savelyev20094044} is founded upon the ideas first presented
in~\cite{Swendsen1979} and is based on matching correlators, obtained from
atomistic and coarse-grained simulations, for observables that explicitly enter the coarse-grained Hamiltonian. 
% transforms the interactions for a specified
% group of atoms towards effective interactions based on the interactions within this group {\color{red} This sentence needs revision}.
% Depending on the
% transformations employed,
% problems might arise for recovering the cross interactions between the different groups.
Data-driven, variational CG methods such as  Multiscale CG~\cite{izvekov:134105, noid2007},   Relative Entropy~\cite{shell2008}, Ultra GG~\cite{dama2013}, offer a rigorous way   of learning CG models  by approximating the Potential of Mean Force (PMF) \cite{Leach2001} with respect to the CG variables on the basis of appropriate functionals. 
% For a broad overview on coarse-graining methods we refer to~\cite{noid2013}.

It is obvious that unless there are known redundancies in the all-atom or fine-grained (FG) description, any coarse-graining scheme will result in information loss~\cite{katsoulakis_information_2006,foley_impact_2015}.   A manifestation of this can be seen if one attempts to reconstruct 
the microscopic, FG configurations from the  CG states~\cite{katsoulakis2006,trashorras2010}. Discrepancies will appear not only because the CG statistics are not captured correctly, but because the CG variables do not encode  all the details needed to reproduce the FG picture.
Despite this, predictions generated by existing CG schemes are always in the form of {\em point estimates} that do not reflect any of the predictive uncertainty which the aforementioned information loss induces.
It is also reasonable to expect that this information loss increases the larger  the difference between the dimension of fine and coarse descriptions becomes. Nevertheless given two competing CG descriptions of the same dimension, it is unlikely that both will capture the FG picture equally well. The discovery of a good set of CG variables (analogous to finding good reaction coordinates or collective variables in free energy computations~\cite{rohrdanz_discovering_2013})  is,
on one hand, a function of the macroscopic quantities of interest but more importantly of the complex structure of inter-dependencies in the FG model.

% from noid 2013:
% In essence, the outstanding challenge for “bottom-up”
% coarse-graining strategies is to determine approximations to
% W that are tractable to calculate, efficient to simulate, suffi-
% ciently accurate for describing a given phenomena, and, ide-
% ally, useful (i.e., transferable) for modeling molecular systems
% and thermodynamic state points other than the one for which
% they were parameterized. At present, no single method com-
% pletely addresses this challenge. Rather, a vast array of diverse

% 
% The relative entropy formulation~\cite{shell2008} provides such a connection between coarse-graining and information-theoretic concepts and tools.  
% It is based on the Kullback-Leibler (KL) divergence ~\cite{kullback1951} as
% a measure of distance between the  distributions describing the 
% fine- and the coarse-scale statistics.

The starting point of all CG schemes is the prescription of the coarse variables through a many-to-one, {\em fine-to-coarse} map.
 Such maps are dictated by the analysis objectives but also by physical insight on which FG features might be important~\cite{noid2013}. For example several atoms/molecules can be lumped into a single, effective, pseudo-molecule with coordinates defined by considering the center of mass. 
 A central component of the present work is the implicit definition of the CG variables through a {\em coarse-to-fine} map. This is achieved by a {\em probabilistic generative model} that treats the CG degrees of freedom as latent variables and explicitly quantifies the uncertainty in the reconstruction of the FG states from the CG description.
The model is complemented with a distribution for the CG variables. Both densities are parametrized  and the optimal values are determined on the basis of an information-theoretic objective (e.g. minimizing a Kullback-Leibler divergence as in~\cite{shell2008}) which is shown to be a special case of a more general, Bayesian framework. 
The latter offers a critical advantage over existing techniques as it enables the prediction of macroscopic observables not only in the form of point estimates, but by providing whole distributions. These  reflect the uncertainty due the aforementioned information loss as well as the fact that finite amounts of training data were used. 

The emphasis on this amplified predictive ability of the proposed framework is the reason behind the title chosen for the present paper {\em predictive coarse-graining} (PCG). 
The Bayesian framework advocated offers a superior setting for model selection. We make use of hierarchical prior models that promote the discovery of a sparse set of features in the aforementioned model components. This enables the search to be carried out over a very large set of feature functions for the CG potential which naturally amplifies the expressivity of the model~\cite{noid2013}.
  We note that 
a  Bayesian framework towards uncertainty quantification 
for force field parameters in molecular dynamics was introduced in
\cite{Angelikopoulos2012,Angelikopoulos2013}. Other Bayesian formulations of coarse-graining problems using macroscopic observables were presented in~\cite{Farrell2014, Farrell2015} where also the issues of model calibration and validation were discussed. 
The structure of the rest of the paper is as follows. Section \ref{sec:methodology} presents the basic model components, compares them with other CG schemes (primarily the relative entropy method), provides details on the exponential family of distributions employed for which uniqueness of solution can be proven  and discusses in detail algorithmic and computational aspects.  Numerical evidence of the capabilities of the proposed framework is provided in Section \ref{sec:numericIl} where coarse-graining efforts for a Ising  lattice system as well as for the SPC/E water model are documented.  
% We support mentioned advantages and capabilities of predictive coarse-graining
% in .
% The proposed scheme is used for coarse-graining a binary two-dimensional lattice-system.
% Potential observables of such systems are the the
% magnetization or coverage and the correlation between the lattice sites.
In all numerical examples, we report results on the {\em predictive uncertainty}  as a function of  the level of coarse graining, and the amount of  data available.
%We find sparse solutions which accelerate predictions next to
% coarse graining the degrees of freedom.
Finally, Section \ref{sec:summary}, summarizes the main contributions and discusses natural extensions of the proposed framework. 
% To summarize the differences in the presented coarse-graining scheme is the
% introduction of
% a Bayesian formalism, with defining a probabilistic mapping form the coarse-scale to the fine-scale.
% The mapping is parametrized and its parameters are optimized during the learning. With this formalism
% we account for uncertainty induced by insufficient coarse-models, restricted availability
% of fine-scale data, and the mapping uncertainty.

% WHY the PREDICTIVE IN THE TITLE

%{\color{red}[ Make reference to Texas group publications, particularly the Bayesian aspects and their relation to ours]}

\section{Methodology}
\label{sec:methodology}

This section introduces the notational conventions adopted and presents the proposed modeling  and computational frameworks. We frequently draw comparisons with the relative entropy method introduced  in~\cite{shell2008} and further expanded and studied in~\cite{chaimovich2011,bilionis2013} in order to  shed light on the aspects related to  information loss and to emphasize the need for quantifying  the resulting   uncertainty in the predictions.
\subsection{Equilibrium statistical mechanics}
\label{sec:esm}
We consider molecular ensembles in equilibrium described by an $n\aaa$-dimensional vector denoted by  $\bq \in \calM\aaa \subset \R^{n\aaa}$. This generally consists of the coordinates of the atoms which follow the Boltzmann-Gibbs density\footnote{In the following,  we assume all probability measures are absolutely continuous with the Lebesgue measure and therefore work exclusively with the corresponding probability density functions.}:
\be
\label{eqn:pf}
p\aaa(\bq| \beta) = \frac{\exp\left\{-\beta U\aaa(\bq)\right\}}{Z\aaa(\beta)},
\ee
where $U\aaa(\bq)$ is the all-atom (fine-grained) potential, $\beta = \frac{1}{k_b T}$ where $k_b$ is the Boltzmann constant and $T$ is the temperature, and $Z\aaa(\beta)$ is the normalization constant (partition function) given by:
\be
 Z\aaa(\beta) = \int_{\calM\aaa} \exp \{ -\beta U\aaa(\bq) \} d \bq.
\ee

In the following, we assume that the temperature $T$ (or equivalently $\beta$) is constant as it is commonly done in coarse-graining literature, even though it is generally of interest to derive coarse-grained descriptions that are suitable for all (or at least a wide range) of temperatures~\cite{noid2013}. 
In this setting and in order to simplify the notation, we drop the temperature dependence.

If  $a(\bq): \calM\aaa \to \R$   denotes an observable (e.g. magnetization in Ising models), then the  corresponding macroscopic properties can be computed as an expectation with respect to to $p\aaa(\bq)$ as follows:
\be
 \label{eqn:predictivePf}
 \E_{p\aaa(\bq)}[a(\bq)] = \int_{\calM\aaa} a(\bq) p\aaa(\bq) d \bq.
\ee
Such expectations are (approximately) computed using long and cumbersome simulations as explained in the introduction e.g. by a long MCMC run~\cite{cances_theoretical_2005}.
Our goal is two-fold. Firstly,  to construct a coarse-grained description of the system that would be easier and faster to simulate, and secondly to use this in order to predict  expectations of any observable as in~\refeqq{eqn:predictivePf}.
A distinguishing aspect of the proposed PCG  framework is that we also   compute  quantitative metrics of the predictive uncertainty in those estimates.  
At a third level, one would also want the coarse-grained description to provide a decomposition  of the original, all-atom ensemble into physically interpretable terms and interactions. We defer such a discussion on how the proposed model can achieve this goal for the conclusions.

We denote by $\bQ$  the  coarse-grained  variables  and assume that they take values in $\calM\cg \subset \R^{n\cg}$. It is obviously desirable  that $n\cg \ll n\aaa$. Let also  $U\cg(\bQ)$ denote the potential associated with $\bQ$ and  $p\cg(\bQ)$ the corresponding density:
\be
\label{eqn:pc}
p\cg(\bQ) = \frac{\exp\left\{-\beta U\cg(\bQ)\right\}}{Z\cg},
\ee
with the normalization constant,
\be
 Z\cg = \int_{\calM\cg} \exp \{ -\beta U\cg(\bQ) \} d \bQ.
\ee

In existing coarse-graining formulations, the coarse variables $\bQ$ are defined
 using a restriction, fine-to-coarse map  $\calR: \calM\aaa \rightarrow \calM\cg$ i.e. $\bQ=\calR(\bq)$.  As this is generally a many-to-one map, it is not invertible
\cite{bilionis2013}. If the observables of interest actually depend on $\bQ$ i.e. if $a(\bq)=A(\calR(\bq))=A(\bQ)$, then one can readily show that it suffices that $p\cg(\bQ)$ is equal to the marginal of $\bQ$ with respect to $p\aaa(\bq)$, or equivalently that $U\cg(\bQ) = U\cg^{\mathrm{opt}}(\bQ)$ where:
\be
 U\cg^{\mathrm{opt}}(\bQ)=-\beta^{-1} \log \int \delta(\bQ-\calR(\bq)) ~p\aaa(\bq) d \bq.
 \label{eq:ucopt}
\ee
That is the coarse-scale potential $U\cg(\bq)$ coincides with the potential of mean-force of $\bQ$. This is a consequence of the following equalities:
%\be
\begin{align}%{ll}
 \E_{p\aaa}[a] & =\int_{\calM\aaa} a(\bq) ~p\aaa(\bq)~d\bq \nonumber \\
  & = \int_{\calM\aaa} A(\calR(\bq)) ~ p\aaa(\bq)~d\bq \nonumber \\
   & = \int_{\calM\aaa} \left( \int_{\calM\cg} A(\bQ) \delta(\bQ-\calR(\bq))~d\bQ \right) p\aaa(\bq)~d\bq \nonumber \\
   & = \int_{\calM\cg} A(\bQ) \left( \int_{\calM\aaa}  \delta(\bQ-\calR(\bq))~p\aaa(\bq)~d\bq  \right) d\bQ \nonumber \\
   & = \int_{\calM\cg} A(\bQ) ~p_c(\bQ) ~d\bQ.    \nonumber
\end{align}
%\ee
Nevertheless, even if one is able to compute or approximate sufficiently well $U\cg^{\mathrm{opt}}(\bQ)$, there is no guarantee that expectations of other observables that do not solely depend on $\bQ$ can be accurately computed. Consistent reconstructions of the all-atom configurations $\bq$, given $\bQ$ samples from $p\cg(\bQ)$,  can be obtained from the conditional:
\be
p_{\calR}(\bq | \bQ)= \frac{ \delta( \bQ-\calR(\bq))}{Z_{\calR}(\bQ)},  
\label{eq:cond}
\ee
i.e. the uniform density on the manifold in $\calM\aaa$ implied by the map $\calR$\footnote{In~\cite{rudzinski2011} this is further generalized by introducing an additional, weighting density.}, where:
\be
Z_{\calR}(\bQ)=\int \delta( \bQ-\calR(\bq))~d\bq.
\ee
Given a coarse-grained potential $U\cg$ (not necessarily the optimal as in~\refeqq{eq:ucopt})  and the density $p\cg(\bQ)$ 
in~\refeqq{eqn:pc}, the corresponding reconstruction density of the all-atom description 
consistent with the map $p_{\calR}(\bq | \bQ)$ (\refeqq{eq:cond}) is given by:
%\be
\begin{align}%{ll}
   {p}_{\calR}(\bq) & = \int p_{\calR}(\bq | \bQ)~p\cg(\bQ)~d\bQ \nonumber \\
   & = \int \frac{ \delta( \bQ-\calR(\bq))}{Z_{\calR}(\bQ)} ~p\cg(\bQ) ~d\bQ \nonumber \\
   & = \frac{p\cg(\calR(\bq))}{Z_{\calR}(\calR(\bq))}.
\label{eq:pR}
\end{align}
%\ee

We note that in the context of the relative entropy method~\cite{shell2008}, which like ours,  is data-driven and has an information-theoretic underpinning,
the goal is to identify the $U\cg$ (within a certain class) that brings ${p}_{\calR}(\bq)$ (\refeqq{eq:pR}) as close as possible to the reference, FG density 
 $p\aaa(\bq)$ (\refeqq{eqn:pf}). For that purpose the Kullback-Leibler (KL) divergence~\cite{cover_elements_1991}  $\kl(p\aaa(\bq) || {p}_{\calR}(\bq))$
 is employed as the objective which, based on \refeqq{eq:pR}, is given by:
%\be
\begin{align}%{ll}
 0 \le \kl(p\aaa(\bq) || {p}_{\calR}(\bq)) & = - \int p\aaa(\bq) \log \frac{ {p}_{\calR}(\bq)}{p\aaa(\bq)}~d\bq \nonumber \\
 & = - \E_{p\aaa(\bq)}[ \log ~ p\cg(\calR(\bq)] + \E_{p\aaa(\bq)}[\log Z_{\calR}(\calR(\bq))] -H(p\aaa), 
 \label{eq:klre}
\end{align}
%\ee
where $H(p\aaa)$ is the entropy of $p\aaa(\bq)$, which is independent of $U\cg$ and can be ignored in the minimization. 
As it has been identified in several investigations~\cite{chaimovich2011, bilionis2013, rudzinski2011}, while the first term can be reduced by adjusting $U\cg$ (it can be shown that the minimum is attained when  $U\cg(\bQ) = U\cg^{\mathrm{opt}}(\bQ)$), the second term is fixed once the restriction map $\calR$  that defines the coarse-grained variables has been selected. It represents a constant penalty  reflecting  the information loss that takes place due to the coarse-grained (and generally lower-dimensional) description adopted. Our goal is to reduce this component of information loss.

\subsection{Probabilistic generative model}

We propose a {\em probabilistic, generative model}~\cite{bishop_latent_1999} in which the coarse description is treated as a latent (hidden) state.   In particular, we define a {\em joint} density $\bar{p}(\bQ,\bq)$ for $\bQ$ and $\bq$ as follows:
\be
\bar{p}(\bQ,\bq)=p\cf(\bq|\bQ)~p\cg(\bQ). 
\label{eq:joint}
\ee
This consists of two components i.e.:
\begin{enumerate}[(i)]
 \item a density $p\cg(\bQ)$  describing the statistics of the coarse-grained description $\bQ$,
 \item a {\bf  probabilistic, coarse-to-fine mapping} implied by the conditional density $p\cf(\bq|\bQ)$.
\end{enumerate}
We discuss the form and parametrization of the aforementioned densities in the sequel. 
We emphasize at this stage the different definition of the coarse-grained variables as latent  generators  that give rise to the observables through the probabilistic {\em lifting} operator   implied by  $p\cf$~\cite{katsoulakis2003},  in contrast to the restriction operators employed in other schemes explained previously. 
 Such mappings can take various forms (e.g. local
or global, linear or nonlinear) and can be extended to many hierarchical levels,
as it will be shown. Understanding the meaning of the latent
variables can only be done through the prism of this generative mapping. According to this, each FG configuration $\bq^{(i)}$ is generated as follows:  
\bi
\item Draw a CG configuration $\bQ^{(i)}$ from $p\cg(\bQ )$.
\item Draw $\bq^{(i)}$ from $p\cf(\bq| \bQ^{(i)})$.
\ei

As we will show, an advantage of the proposed framework is that it readily provides  a (predictive) probability
density for the observables of interest.  
The marginal density of the FG description $\bq$ is given from~\refeqq{eq:joint} by integrating out $\bQ$:
\be
 \label{eqn:predictiveCg}
 \bar{p}\aaa (\bq) = \int_{\calM\cg} p\cf(\bq|\bQ)~p\cg(\bQ) d\bQ.
\ee

Suppose the aforementioned component densities are parametrized by $\btheta=(\bgam\cg,\bgam\cf)$ i.e.  $p\cg(\bQ|\bgam\cg)$  and  $p\cf(\bq|\bQ, \bgam\cf)$, and we attempt to minimize the KL-divergence between the reference density $p\aaa(\bq)$ and the marginal
$\bar{p}\aaa (\bq| \btheta)$ implied by the generative model proposed :
%\be
\begin{align}%{ll}
 \kl ( p\aaa(\bq) || \bar p\aaa(\bq) ) &= - \int_{\calM\aaa} p\aaa(\bq)
 \log \frac{ \bar p\aaa (\bq) }{ p\aaa(\bq) } d \bq \nonumber \\
 & =- \int p\aaa(\bq) \log \bar{p}\aaa (\bq | \btheta) ~d\bq+\int p\aaa(\bq) \log p\aaa(\bq) d \bq.
% &= - \int p\aaa(\bq) \log \left(\int_{\calM\cg} p\cf(\bq|\bQ) p\cg(\bQ) d \bQ\right) ~d\bq +  \int p\aaa(\bq) \log  p\aaa(\bq) d \bq.
\label{eqn:KLpredictive}
\end{align}
%\ee
This is equivalent to  {\em maximizing} $\int p\aaa(\bq) \log \bar{p}\aaa (\bq | \btheta) ~d\bq$ which, given samples $\{\bq^{(i)}\}_{i=1}^N$ from $p\aaa(\bq)$ is approximated by the {\em log-likelihood} of $\bar{p}\aaa (\bq | \btheta)$ \footnote{This  result can be obtained (up to $1/N$) by substituting $p\aaa(\bq)$ in~\refeqq{eqn:KLpredictive} by the empirical measure $\frac{1}{N} \sum_{i=1}^N \delta(\bq-\bq^{(i)} )$. The likelihood of $N$ samples drawn from $p\aaa(\bq)$ is trivially $\prod_{i=1}^N \bar{p}\aaa (\bq^{(i)} |\btheta)$. }:
%\be
\begin{align}%{ll}
 \calL(\btheta) & = \sum_{i=1}^N \log \bar{p}\aaa (\bq^{(i)} |\btheta) \nonumber \\
 & = \sum_{i=1}^N \log \left(\int_{} p\cf(\bq^{(i)} |\bQ^{(i)},\bgam\cf )~ p\cg(\bQ^{(i)} | \bgam\cg) d \bQ^{(i)} \right).
\label{eq:loglike}
\end{align}
%\ee
We note  in the expression above that we associate a latent, coarse configuration $\bQ^{(i)}$ to each sample $\bq^{(i)}$ which is effectively its pre-image.  
More importantly, the objective in the aforementioned expression accounts  for both the density of the coarse-grained description as well as the reconstruction (lifting) of the all-atom configuration from the (latent) coarse-grained one.
Maximizing $\calL(\btheta)$ naturally leads to the Maximum Likelihood estimate $\btheta_{\mathrm{MLE}}$.

\begin{figure}
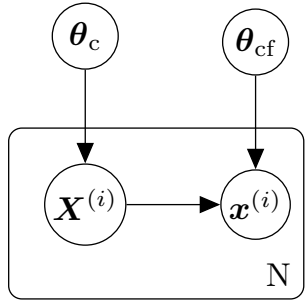

\centering
\resizebox{0.3\textwidth}{!}{%
      \tikz{ %
        \node[latent] (Xr) {$\bQ^{(i)}$} ; %
        \node[latent, above=of Xr] (thetacg) {$\bgam\cg$} ; 
        \node[latent, right=of Xr] (xrs) {$\bq^{(i)}$} ; %
        \node[latent, above=of xrs] (thetacf) {$\bgam\cf$} ; 
        %\plate[inner sep=0.25cm, xshift=-0.12cm, yshift=0.12cm] {plate1} {(xrs)} {S}; %
        %\plate[inner sep=0.25cm, xshift=-0.12cm, yshift=0.12cm] {plate2} {(Xr) (plate1)} {R}; %
        \plate[inner sep=0.25cm, xshift=-0.12cm, yshift=0.12cm] {plate3} {(Xr) (xrs)} {N}; %
        \edge[] {Xr} {xrs} ; %
        \edge {thetacg} {Xr} ; %
        \edge {thetacf} {xrs} ; %
}
}
\caption{Probabilistic graphical model representation.}
\label{fig:graph}
\end{figure}

Furthermore the interpretation of the objective  as the log-likelihood makes the progression into Bayesian formulations much more straightforward. If for example we define a prior density $p(\btheta)$
then maximizing:
\be
\arg \max_{\btheta}~ \left\{ \calL(\btheta) + \log p(\btheta) \right\}, 
\label{eq:map}
\ee
is equivalent to obtaining a Maximum a Posteriori (MAP) estimate $\btheta_{\mathrm{MAP}}$~\cite{bishop2006}.  The next step from  point estimates for the model parameters is of course obtaining the full posterior $p(\btheta | \bq^{(1:N)})$ using Bayes formula as:
%\be
\begin{align}%{ll}
 p(\btheta | \bq^{(1:N)}) & \propto p(\bq^{(1:N)} | \btheta) ~p(\btheta) \nonumber \\
 & \propto e^{\calL(\btheta)} ~p(\btheta) \nonumber \\
%  & \propto \prod_{i=1}^N p(\bq^{(i)} |\btheta) ~p(\btheta) \\
 & \propto \prod_{i=1}^N \left(\int_{} p\cf(\bq^{(i)} |\bQ^{(i)},\bgam\cf )~ p\cg(\bQ^{(i)} | \bgam\cg) d \bQ^{(i)} \right) p(\btheta).
\label{eq:post}
\end{align}
%\ee
The aforementioned relationship can be concretely represented in the form of a directed graphical model as depicted in Fig.~\ref{fig:graph}.

We discuss a  strategy for approximating this posterior in the next subsections. It is more important to emphasize at this stage that given this posterior,
we can produce not just point  estimates of the expectation of any observable $a(\bq)$, but also compute its predictive posterior. 
For that purpose we make use of the predictive posterior $p( \bq | \bq^{(1:N)} )$ of our model  which is determined by marginalizing the latent variables $\bQ$ and the model parameters  $\btheta$:
%\be
\begin{align}%{ll}
 p( \bq | \bq^{(1:N)} ) & = \int p(\bq, \bQ, \btheta | \bq^{(1:N)} )~d\bQ d\btheta \nonumber \\
 & = \int p(\bq, \bQ | \btheta , \bq^{(1:N)} )~p(\btheta| \bq^{(1:N)} )~d\bQ d\btheta.
\end{align}
%\ee
%arising from the generative model proposed
By replacing the joint density with the proposed generative model in \refeqq{eq:joint}, the predictive posterior $p( \bq | \bq^{(1:N)} )$ becomes:
\be
   p( \bq | \bq^{(1:N)} ) = \int p\cf(\bq | \bQ, \bgam\cf)~p\cg(\bQ |  \bgam\cg) ~p(\btheta | \bq^{(1:N)})~d\bQ d\btheta.
\label{eq:predPost}
\ee
The latter can be used in place of the FG distribution $p\aaa(\bq)$ in \refeqq{eqn:predictivePf}, to obtain approximations to the  expectation of any observable $a(\bq)$ as follows:
% The latter can be used in conjunction with any observable $a(\bq)$ as in~\refeqq{eqn:predictivePf}, to make {\em probabilistic} predictions as follows:
%\be
\begin{align}%{ll}
 \E_{\pf(\bq)} [a(\bq)] &\approx \E_{p(\bq|\bqd)} [a(\bq)] \nonumber \\
 & = \int a(\bq)~p(\bq|\bqd)~d\bq  \nonumber \\
 & = \int a(\bq)~\left( \int p\cf(\bq | \bQ, \bgam\cf)~p\cg(\bQ |  \bgam\cg) ~p(\btheta | \bq^{(1:N)}) ~d\bQ~d\btheta  \right) d\bq  \nonumber \\
 & = \int  \underbrace{ \left( \int a(\bq)~p\cf(\bq | \bQ, \bgam\cf)~p\cg(\bQ |  \bgam\cg)~d\bQ ~ d\bq \right) }_{\hat{a}(\btheta )} ~p(\btheta | \bq^{(1:N)}) ~d\btheta  \nonumber \\
 & = \int  \hat{a}(\btheta )~p(\btheta | \bq^{(1:N)}) ~d\btheta.
 \label{eqn:propPrediction}
\end{align}
%\ee
The approximation in the first line reflects the quality of the model as well as the uncertainty arising from the finite  data $\bqd$ that were used to calibrate it.
This derivation suggests that $\hat{a}(\btheta )$ represents the predictive estimate of the expectation of $a(\bq)$ for a given value $\btheta$ of the model's parameters.
Averaging over the posterior of the latter provides the expected (a posteriori) value of this quantity.
More importantly though by propagating the (posterior) uncertainty of $\btheta$ through $ \hat{a}(\btheta )$, one can readily obtain the predictive distribution of the observable.
In the numerical examples we frequently plot such posterior statistics, usually in the form of credible intervals (see also~\ref{app:credInt}).
% that reflects the epistemic uncertainty in the probabilistic, generative model.
% \red{
% The aforementioned epistemic uncertainty stems from the information loss due to  the lower-dimensional CG description and the the finite amount of training data. It is reflected 
% which is reflected in the extent of credible intervals\footnote{\red{Details on estimating
% credible intervals are denoted in \ref{app:credInt}.}}
% of observables such as magnetization and correlation for the Ising model,
% respectively for SPC/E water the radial and angular distribution in Sections~\ref{sec:num_ising} and~\ref{sec:num_water}.
% }
Point estimates can be easily recovered if the analyst wishes to do so by employing for example the  MAP (or MLE) estimate $\btheta_{\mathrm{MAP}}$
in the aforementioned equation  i.e. if $p(\btheta | \bq^{(1:N)}) \equiv \delta(\btheta-\btheta_{\mathrm{MAP}})$.

\subsection{Inference and learning (point estimates)}
\label{sec:inference_learning}
This section is concerned with the computational aspects of training the proposed model. We pay particular attention to distributions in the exponential family for which the concavity of the maximum-likelihood problem can be analytically shown. Furthermore, we  discuss strategies for parallelizing these tasks and improving the computational efficiency. We finally discuss particular prior specifications that are suitable for sparse feature recovery and model selection. 

We begin our discussion with a strategy for  obtaining point estimates for the model parameters $\btheta$ by maximizing the log-likelihood (or the log-posterior)
as given in~\refeqq{eq:loglike} (or~\refeqq{eq:map}).
The difficulty in the optimization problem stems from the intractability of the log-likelihood
due to the integration with respect to the latent variables $\bQ^{(i)}$ (except for trivial cases for $p\cg, p\cf$).
%  the marginalization of which leads to an intractable expression 
% \red{, hence direct optimization of the log-likelihood is impractical}.
To address this we employ an {\em  Expectation-Maximization} (EM) scheme~\cite{dempster_maximum_1977,neal_view_1998}
where MCMC is used to approximate the E-step (MCEM)~\cite{wei_monte_1990}
and stochastic approximations to handle the Monte Carlo noise in the gradient
estimates of the M-Step~\cite{robbins1951,cappe_inference_2005}.
The EM algorithm allows the maximization of the log-likelihood by
circumventing the need for repeated evaluations of the aforementioned intractable integrals
and normalization constants.
%According \cite{neal_view_1998}, constructing a lower bound of the log-likelihood the optimization is performed on the joint data space ($\bq$,$\bQ$) leading to tractable expressions in the E- and M-step.
% }
To motivate the derivation, we note that for an arbitrary set of 
densities $q_i(\bQ^{(i)})$ we can construct  lower bounds, denoted by  $\mathcal{F}^{(i)} (q_i(\bQ^{(i)}),~\btheta)$,  {\em for each term}
in the sum that makes up the log-likelihood as follows:
%\be
\begin{align}%{ll}
  \calL(\btheta) &  = \sum_{i=1}^N ~\log \left(\int_{} p\cf(\bq^{(i)} |\bQ^{(i)},\bgam\cf )~ p\cg(\bQ^{(i)} | \bgam\cg) d \bQ^{(i)} \right) \nonumber \\ 
  & = \sum_{i=1}^N ~  \log \left( \int_{} \frac{ p\cf(\bq^{(i)} |\bQ^{(i)},\bgam\cf )~ p\cg(\bQ^{(i)} | \bgam\cg) }{ q_i(\bQ^{(i)})} q_i(\bQ^{(i)}) ~ d \bQ^{(i)} \right) \nonumber \\
  & \ge \sum_{i=1}^N ~  \underbrace{ \left(  \int_{}  q_i(\bQ^{(i)}) \log \frac{ p\cf(\bq^{(i)} |\bQ^{(i)},\bgam\cf )~ p\cg(\bQ^{(i)} | \bgam\cg) }{ q_i(\bQ^{(i)})} ~ d \bQ^{(i)} \right) }_{:= \mathcal{F}^{(i)} (q_i(\bQ^{(i)}), ~\btheta) } \nonumber  \\
  & = \sum_{i=1}^N ~ \mathcal{F}^{(i)} (q_i(\bQ^{(i)}), ~\btheta) \nonumber \\
  & = \mathcal{F}(\bs{q}(\bQ), ~\btheta), 
\end{align}
%\label{eq:lowerbound}
%\ee
% \red{
% where $\mathcal{F}^{(i)} (q_i(\bQ^{(i)}), ~\btheta)$ denotes the contribution to the lower-bound of the log-likelihood from the corresponding data-point $(i)$,
% }
where $\bs{q}(\bQ)=\prod_{i=1}^N q_i(\bQ^{(i)})$, and the result in the third step is a consequence of Jensen's inequality.  We note that the optimal $q_i^{\mathrm{opt}}(\bQ^{(i)})$ for each of the aforementioned terms is:
\be
q_i^{\mathrm{opt}}(\bQ^{(i)}) = q_i(\bQ^{(i)}|\bqi, \btheta) \propto p\cf(\bq^{(i)} |\bQ^{(i)},\bgam\cf )~ p\cg(\bQ^{(i)} | \bgam\cg), 
\label{eq:qopt}
\ee
i.e. the conditional posterior of the latent variables $\bQ^{(i)}$ given $\bqi$ and $\btheta$. This is optimal in the sense that the inequality becomes an equality~\cite{bishop2006} i.e.:
\be
\mathcal{F}^{(i)} (q_i^{\mathrm{opt}}(\bQ^{(i)}), ~\btheta)=\log \left(\int p\cf(\bq^{(i)} |\bQ^{(i)},\bgam\cf )~ p\cg(\bQ^{(i)} | \bgam\cg) d \bQ^{(i)} \right).
\ee
All other $q_i$'s lead to suboptimal schemes that fall under the category of Variational Bayesian Expectation-Maximization (VB-EM,~\cite{beal_variational_2003}).
More importantly, the aforementioned derivation suggests an iterative algorithm where one alternates (until convergence) between the following two steps, i.e. at each iteration $t$:
\bi
\item[{\bf E-step}:] Given the current estimate of $\btheta \equiv\btheta^{(t)}$, evaluate:
\be
 \mathcal{F}(\bs{q}^{\mathrm{opt},\,t}(\bQ), ~\btheta^{(t)})=\sum_{i=1}^N ~ \mathcal{F}^{(i)} (q_i^{\mathrm{opt},\,t}(\bQ^{(i)}), ~\btheta^{(t)}), 
\label{eq:eme}
\ee
where $q_i^{\mathrm{opt},\,t}$ is given in~\refeqq{eq:qopt} for $\btheta \equiv\btheta^{(t)}$.

\item[{\bf M-step}:] Given the current $q_i^{\mathrm{opt},\,t}(\bQ^{(i)})$, find:
%\be
\begin{align}%{ll}
 \btheta^{(t+1)} & =\arg \max_{\btheta} \sum_{i=1}^N ~ \mathcal{F}^{(i)} (q_i^{\mathrm{opt},\,t}(\bQ^{(i)}), ~\btheta^{(t)}) \nonumber \\
 & =\arg \max_{\btheta} \sum_{i=1}^N ~  \left(  \int  q_i^{\mathrm{opt},\,t}(\bQ^{(i)}) \log  \left( p\cf(\bq^{(i)} |\bQ^{(i)},\bgam\cf^{(t)} )~ p\cg(\bQ^{(i)} | \bgam\cg^{(t)}) \right)  ~ d \bQ^{(i)} \right).
 \label{eq:emm}
\end{align}
%\ee
\ei

We discuss in detail each of the two steps. 
\bi
\item The E-step of the algorithm requires computing expectations with respect to the intractable distributions in~\refeqq{eq:qopt}. As it can be seen in~\refeqq{eq:emm} only the terms in  $\mathcal{F}^{(i)}$ that depends on $\btheta$ needs to be computed which we approximate by a Monte Carlo estimator: 
%\be
\begin{align}%{l}
\int  q_i^{\mathrm{opt},\,t}(\bQ^{(i)}) \log  \left( p\cf(\bq^{(i)} |\bQ^{(i)},\bgam\cf^{(t)} )~ p\cg(\bQ^{(i)} | \bgam\cg^{(t)}) \right) d\bQ^{(i)}  \approx \nonumber \\
\approx \frac{1}{m_t} \sum_{j=1}^{m_t}\left( \log  p\cf(\bq^{(i)} |\bQ^{(i)}_j,\bgam\cf^{(t)} )~ p\cg(\bQ^{(i)}_j | \bgam\cg^{(t)}) \right).
\label{eq:gradientmcest}
\end{align}
%\ee
The $m_t$ samples  used at each iteration $t$ are drawn using MCMC from $q_i^{\mathrm{opt},\,t}(\bQ^{(i)})$. 
Compared to i.i.d. Monte Carlo samples, the use of MCMC introduces theoretical complications with regards to the stability and the error in the approximation~\cite{younes_convergence_1999,andrieu_stability_2005}. A  recent treatment of the convergence conditions for such schemes is contained in~\cite{fort_convergence_2016}. The obvious error source arises from the bias in the  MCMC samples which are {\em approximately} distributed according to the target density. In addition the samples generated are correlated. Such errors can be subdued by increasing the sample size $m_t$. 
Heuristically speaking, at the first few iterations $t$, even a crude estimate of the objective generally suffices to drive the parameter $\btheta$-updates toward the region of interest. As the EM iterations proceed, the number of samples should increase in order to zoom-in at the optimum and  minimize the oscillatory behavior due to the noise in the estimates.
Several strategies have been proposed to optimize $m_t$ or even devise an  automatic schedule  
 by making use of  error estimates 
\cite{booth_maximizing_1999,levine_implementations_2001,
fort_convergence_2003,levine_automated_2004}. In this work, we used a constant sample size i.e. $m_t=m, \forall t$ that we report in the numerical examples. We found through several cross-validation runs that this had no noticeable effect to the optima identified. 
We note finally that other Monte Carlo schemes can be utilized. One would expect that Importance Sampling~\cite{liu_monte_2008}, where previously generated samples are re-weighted and re-used, could be quite effective particularly when $\btheta^{(t)}$ do not change much and the corresponding $q_i^{\mathrm{opt},\,t}$ are quite similar. A more potent alternative is offered by Sequential Monte Carlo schemes (SMC)~\cite{bilionis_free_2012,del_moral_feynman-kac_2004}  which combine the benefits of MCMC and Importance Sampling. 

\item The maximization of the lower bound with respect to $\btheta$ is not analytically tractable even when a Monte Carlo approximation of the objective, as discussed previously, is used.
For that purpose, we make use of  a gradient ascent scheme that employs the partial derivatives of $\mathcal{F}$:
%\be
\begin{align}%{ll}
\mathcal{G}(\btheta)=\nabla_{\btheta} ~\mathcal{F} & = \sum_{i=1}^N \nabla_{\btheta} \mathcal{F}^{(i)} ~~(=\sum_{i=1}^N \mathcal{G}^{(i)}(\btheta)) \nonumber \\
& = \sum_{i=1}^N \nabla_{\btheta} \left(  \int  q_i^{\mathrm{opt},\,t}(\bQ^{(i)}) \log  \left( p\cf(\bq^{(i)} |\bQ^{(i)},\bgam\cf^{(t)} )~ p\cg(\bQ^{(i)} | \bgam\cg^{(t)}) \right) ~ d \bQ^{(i)} \right), 
\end{align}
%\ee
where at each iteration $t$, each term $\mathcal{G}^{(i)}(\btheta)$ is approximated by a Monte Carlo estimate (see discussion before) as:
%\be
\begin{align}%{ll}
  \mathcal{G}^{(i)}(\btheta)  & = \nabla_{\btheta}  \int  q_i^{\mathrm{opt},\,t}(\bQ^{(i)}) \log \left( p\cf(\bq^{(i)} |\bQ^{(i)},\bgam\cf^{(t)} )~ p\cg(\bQ^{(i)} | \bgam\cg^{(t)}) \right) ~ d \bQ^{(i)}  \nonumber \\
    & \approx \frac{1}{m_t} \sum_{j=1}^{m_t} \nabla_{\btheta}   \log  \left( p\cf(\bq^{(i)} |\bQ^{(i)}_j,\bgam\cf^{(t)} )~ p\cg(\bQ^{(i)}_j | \bgam\cg^{(t)}) \right) \nonumber \\
    & =\hat{\mathcal{G}}^{(i)}_t.
\label{eq:gradmc}
\end{align}
%\ee
The latter are used to update $\btheta$ as follows\footnote{As discussed in the seminal work of Neal and Hinton~\cite{neal_view_1998}, more than one updates of $\btheta$ per EM iteration can be performed.}:
\be
  \bgam^{t+1} = \bgam^{t} + \eta_t \sum_{i=1}^N \hat{\mathcal{G}}^{(i)}_t.
  \label{eqn:RMupdate}
 \ee
The step sizes $\eta_t$ are defined in the context of the Robbins-Monro scheme~\cite{robbins1951} which is designed to handle the unavoidable Monte Carlo noise in the gradient estimates. They should satisfy the following conditions~\cite{spall2003}:
 \be
 \sum_{t=1}^\infty \eta_t = + \infty, \text{ and } \sum_{t=1}^\infty \eta_t^2 < \infty.
 \ee
 In this work, we employ~\cite{bilionis2013}:
 \begin{equation}
  \eta_{t} = \frac{\alpha}{ (A + t)^\rho},
  \label{eq:rmparam}
 \end{equation}
 with $\rho \in (0.5, 1]$.
The choice for the values $\alpha$, $\rho$, and $A$ is problem dependent
and is explicitly given in Sections~\ref{sec:num_ising} and~\ref{sec:num_water}
for the Ising and water problems, respectively.

\item We note finally that the gradient needed for  the $\btheta-$updates, involves the sum of $N$ independent terms, one for each datum (i.e. FG configuration) available. Apart from the obvious opportunity for parallelization that this offers, it also suggests that fine-scale data can be successively added. Hence the optimization can be initiated with a small number of data points $N$ and the changes in the optimal $\btheta$  identified can be monitored as more fine-scale data are generated/added to ensure that convergence is achieved with the smallest such effort. Another strategy for reducing the computational effort is to perform the E-step i.e. sample from $q_i^{\mathrm{opt},\,t}$ only for a subset of the data $i=1,\ldots, N$ at a time.
While this has the potential of reducing the overall number of MCMC steps needed, convergence is still guaranteed~\cite{neal_view_1998}.
%This will not prevent convergence~\cite{neal_view_1998}, but has the potential of reducing the overall number of MCMC steps needed. 
%
\ei

\subsection{Exponential family densities - Uniqueness of solution}
\label{sec:expon}
In order to provide some insight to the log-likelihood maximization, we consider the case of model densities belong to the exponential family~\cite{bishop2006,mohamed_bayesian_2008}.
As it will be shown in the numerical illustrations, this represents a very large set of flexible densities where by appropriate selection of the feature functions $\bs{\phi}$ and $\bs{\psi}$ in the equations below one can capture interactions of various order (e.g. $2^{nd}, 3^{rd}$)~\cite{bilionis2013, rudzinski2011}. Such densities have the form:
\be
 \pc(\bQ|\bgam\cg) = \exp\{ \bgam\cg^\text{T} \bs{\phi}(\bQ) - A(\bgam\cg) \},
 \label{eq:pcexp}
\ee
and:
\be
 \pcf(\bq|\bQ,\bgam\cf) = \exp\{ \bgam\cf^\text{T} \bs{\psi}(\bq,\bQ) - B(\bQ, \bgam\cf) \},
\ee
where $A(\bgam\cg)$ and $B(\bQ, \bgam\cf)$ are the log-partition functions given by:
%\be
\begin{align}%{ll}
  A(\bgam\cg) = \log \int e^{   \bgam\cg^\text{T} \phi(\bQ) } d\bQ, \nonumber \\
  B(\bQ, \bgam\cf)  = \log \int e^{   \bgam\cf^\text{T} \psi(\bq,\bQ) } d\bq.
\end{align}
%\ee
One can readily show that:
%\be
\begin{align}%{ll}
\frac{\pa  A(\bgam\cg)}{\pa \theta_{\mathrm{c},k}} & = <\phi_k(\bQ)>_{ \pc(\bQ|\bgam\cg) }, \nonumber \\
\frac{\pa^2  A(\bgam\cg)}{\pa \theta_{\mathrm{c},k} \pa \theta_{\mathrm{c},l} }& = \Cov_{\pc(\bQ|\bgam\cg) } [\phi_k(\bQ),\phi_l(\bQ)],
\end{align}
%\ee
and:
%\be
\begin{align}%{ll}
\frac{\pa  B(\bQ,\bgam\cf)}{\pa \theta_{\mathrm{cf},k}} & = <\psi_k(\bq,\bQ)>_{ \pcf(\bq|\bQ,\bgam\cf) }, \nonumber \\
\frac{\pa^2  B(\bQ,\bgam\cf)}{\pa \theta_{\mathrm{cf},k} \pa \theta_{\mathrm{cf},l} }& = \Cov_{\pcf(\bq|\bQ,\bgam\cf) } [\psi_k(\bq,\bQ),\psi_l(\bq,\bQ)],
\end{align}
%\ee
where $<\cdot>_{p}$ denotes the expectation with respect to the density $p$ and $\Cov_p[\cdot,\cdot]$ the covariance of the arguments with respect to $p$.
Hence, for $\pc$ and $\pcf$ as above, the gradient of the objective $\mathcal{F}$ in~\refeqq{eq:emm} is given by
\footnote{We compare gradients of PCG with the relative entropy mehtod in \ref{app:compare_gradients_relEntr}.}:
%\be
\begin{align}%{ll}
  & \frac{\pa \mathcal{F} }{ \pa \theta_{\mathrm{c},k} }= \sum_{i=1}^N ~   \left( <\phi_k(\bQ^{(i)})>_{ q_i(\bQ^{(i)}) } -<\phi_k(\bQ)>_{ \pc(\bQ|\bgam\cg) }  \right), \nonumber \\
  \text{and} ~& \frac{\pa \mathcal{F} }{ \pa \theta_{\mathrm{cf},k} }= \sum_{i=1}^N ~   \left( <\psi_k(\bq^{(i)} ,\bQ^{(i)})>_{ q_i(\bQ^{(i)}) } -<\psi_k(\bq,\bQ^{(i)})>_{ \pcf(\bq|\bQ^{(i)},\bgam\cf) q_i(\bQ^{(i)}) }  \right).
\end{align}
%\ee
Furthermore, the Hessian is:
%\be
\begin{align}%{ll}
 \frac{\pa^2 \mathcal{F} }{ \pa \theta_{\mathrm{c},k} \theta_{\mathrm{c},l}} & = -N \Cov_{\pc(\bQ|\bgam\cg) } [\phi_k(\bQ),\phi_l(\bQ)], \nonumber \\
 \frac{\pa^2 \mathcal{F} }{ \pa \theta_{\mathrm{c},k} \theta_{\mathrm{cf},l}} & = 0, \nonumber \\
 \frac{\pa^2 \mathcal{F} }{ \pa \theta_{\mathrm{cf},k} \theta_{\mathrm{cf},l}} & = -\sum_{i=1}^N \Cov_{\pcf(\bq|\bQ^{(i)},\bgam\cf)  q_i(\bQ^{(i)}) } [\psi_k(\bq,\bQ),\psi_l(\bq,\bQ)].
\label{eq:hessian}
\end{align}
%\ee

The block-diagonal Hessian is negative definite (at least when linearly independent feature functions are employed) which ensures that the objective is concave and has a unique maximum (whether  arbitrary $q_i$ are  employed or $q_i^{\mathrm{opt}}$ as in~\refeqq{eq:qopt}).
We note also that Monte Carlo estimates of the Hessian can also be obtained and used in the $\btheta-$updates. These however tend to be more noisy than the gradients and special treatment is needed unless one is willing to generate large numbers of  MCMC samples~\cite{bilionis2013}. Finally, there is a wealth of stochastic approximation
schemes that have been proposed and exhibit accelerated convergence
\cite{Moritz2015, Byrd2011, Chen2015, Kushner2003}.

\subsection{Prior specification}

The incorporation of priors for $\btheta$ does not pose any computational difficulties as their contribution is additive (see~\refeqq{eq:map}) to the log-likelihood and its partial derivatives.
While priors for $\bgam\cf$, i.e. the parameters in the coarse-to-fine map, are unavoidably problem-dependent due to their special physical meaning, a more general strategy can be adopted for the $\bgam\cg$, i.e. the parameters associated with the density of the coarse-grained variables $\bQ$. 
For exponential family distributions as in~\refeqq{eq:pcexp}, each $\theta_{\mathrm{c},k}$ is associated with a feature function $\phi_k(\bQ)$. As it will  become apparent in the numerical examples, each of these feature functions encapsulates low- or high-order dependencies (or components thereof) between $\bQ$. It is obviously impossible to know a priori which of the $\phi(\bQ)$ are relevant for a particular problem and how these depend on the dimension of $\bQ$ or the coarse-to-fine probabilistic map $p\cf$. This underpins an important {\em model selection} 
 issue that has been of concern in several coarse-graining studies~\cite{noid2013,Farrell2014, Farrell2015, rudzinski2011}.
One strategy to address this is to initiate the search with a small number of features $\phi(\bQ)$ and progressively add more. These can be selected from a pool of candidates by employing appropriate criteria. In~\cite{bilionis_free_2012, della_pietra_inducing_1997} for example, the feature function that causes the largest (expected) decrease (or increase) in the KL-divergence (or the log-likelihood) that we seek to minimize  (or maximize), is added at each step. In this work, we adopt a different  approach whereby {\em all} available $\phi(\bQ)$ contained in the  vocabulary of feature functions, are simultaneously considered. Consequently this leads to a vector of unknowns $\bgam\cg$ of very large dimension which not only impedes computations but can potentially lead to multiple local maxima, if the Hessian in~\refeqq{eq:hessian} becomes semi-negative definite i.e. if linear dependencies between the selected $\phi(\bQ)$ are present. 
More importantly though (at least when the number of data points $N$ is small), it can obstruct the identification of the most salient features of the coarse-grained model which provide valuable physical insight~\cite{noid2013}.

To address this, we propose the use of sparsity-enforcing priors that are capable of identifying solutions in  which  only a (small) subset of $\bgam\cg$ are non-zero and therefore only 
 the corresponding $\phi(\bQ)$ are active~\cite{figueiredo_adaptive_2003,west_bayesian_2003}.
A lot of the prior models that have been proposed along these lines can be readily cast in the context of {\em hierarchical Bayesian models} where {\em hyper-parameters} are introduced  in the prior. In this work, we adopt the Automatic Relevance Determination (ARD,~\cite{mackay1994}) model which consists of the following:
\be
 p(\bgam\cg|\bs{\tau}) \equiv  \prod_k \mathcal{N}(\theta_{\mathrm{c}, k}|0,\tau_k^{-1}) , \quad 
 \tau_k \sim  Gamma( \tau_k |a_0,b_0).
 \label{eqn:ardPrior}
\ee
This implies that each $\theta_{\mathrm{c},k}$ is modeled (a priori) with an independent, zero-mean, Gaussian, with a precision hyper-parameter $\tau_k$ which is in turn modeled (independently) with a (conjugate) Gamma density. 
We note that when $\tau_k \rightarrow \infty$, then $\theta_{\mathrm{c},k} \rightarrow 0$. The resulting prior for $\theta_{\mathrm{c},k}$ arising by marginalizing the hyper-parameter is a heavy-tailed,  Student's $t-$distribution.  For the purposes of  learning of $\bgam\cg$ and in order to compute derivatives of the log-prior, we retain the $\tau_k$'s and treat them  as latent variables in an inner-loop EM scheme~\cite{bishop_variational_2000}
(see derivation in \ref{app:ard}) which consists of:
\begin{itemize}
 \item E-step: evaluate:
 \be \left\langle \tau_k \right\rangle_{p(\tau_k|\theta_{\mathrm{c},k})} =
 \frac{ a_0 + \frac{1}{2} }{ b_0 + \frac{\theta_{\mathrm{c},k}^2}{2} }.
 \label{eq:ardPosterior}
 \ee
 \item M-step: evaluate:
\be
\frac{\partial \log p(\bgam\cg) }{\partial \theta_{\mathrm{c},k} } =
 - \left\langle \tau_k \right\rangle_{p(\tau_k|\theta_{\mathrm{c},k})}  \theta_{\mathrm{c},k}.
\label{eq:priorgrad}
\ee

 \end{itemize}
We note also that the second derivative of the log-prior with respect to $\bgam\cg$ can be  similarly obtained as: 
\be
\frac{\partial^2 \log p(\bgam\cg) }{\partial \theta_{\mathrm{c},k} \partial  \theta_{\mathrm{c},l}} 
  = \left\{ \begin{array}{cc}
             - \left\langle \tau_k \right\rangle_{p(\tau_k|\theta_{\mathrm{c},k})}, & \textrm{if $k=l$}\\
             0, & \textrm{ otherwise.}
            \end{array}
   \right.
\label{eq:priorhessian}
\ee

% \red{
% Does this contain new info?

% The ARD extension adds a hierarchical prior independently associated with each parameter $ \theta_{\mathrm{c},k}$ (respectively feature $\phi_k(\bQ)$) modeled as Gaussian with zero mean and
% precision $\tau_k$ as hyper-parameter following a Gamma distribution with the parameters $(a_0, b_0)$.
% We evaluate in the E-step (see \refeqq{eq:ardPosterior}) the posterior $p(\tau_k| \theta_{\mathrm{c}, k})$ while
% a small precision $\tau_k$ (or large variance) of the corresponding Gaussian prior $p(\theta_{\mathrm{c}, k}|0,\tau_k^{-1})$
% indicates likely large sensitivity of the model with respect to parameter
% $ \theta_{\mathrm{c}, k}$ or feature $\phi_k(\bQ)$ and likely indicates the feature es relevant
% \cite{Neal1996}. The contribution arising from ARD prior to the derivative ( M-step, \refeqq{eq:priorgrad}),
% drives likely irrelevant parameters  $\theta_{\mathrm{c}, k}$ and respectively features
% $\phi_k(\bQ)$ to zero.
% }

\subsection{Approximate Bayesian inference - Laplace's approximation}
\label{sec::approxBayInf}
\begin{figure}
 \includegraphics[width=0.4\textwidth]{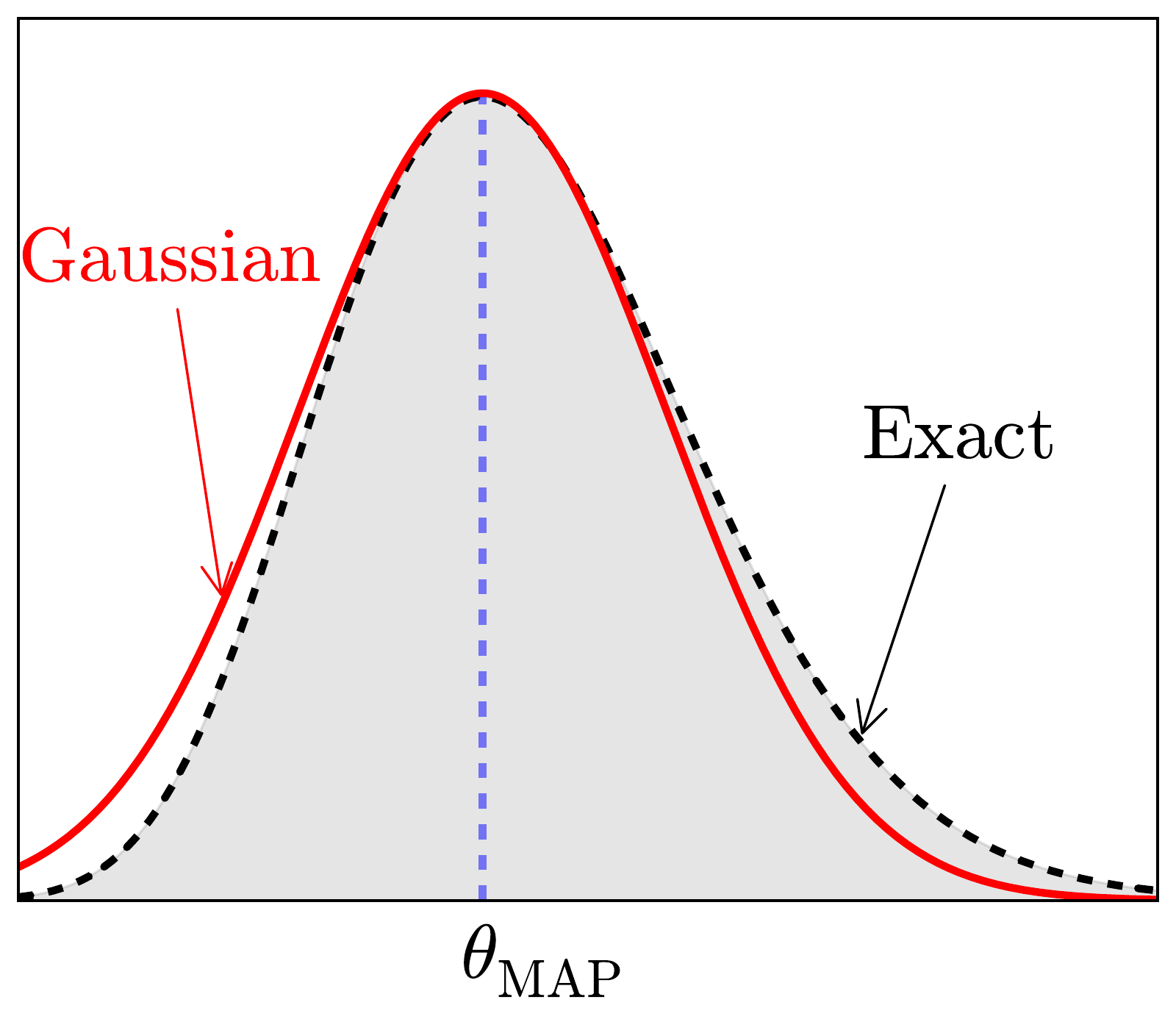}
 \caption{Schematic illustration of the Laplace's approximation.}
 \label{fig:laplace}
\end{figure}

The discussion thus far has been limited to point estimates for $\bgam$. A fully Bayesian treatment would pose significant computational challenges. These stem from the intractability of the log-partition function $A(\bgam\cg)$ of $p\cg$ in the exponential family of models (see~\refeqq{eq:pcexp}). Sampling or approximating the full posterior of $\bgam\cg$ would require repeated evaluations of this and potentially its derivatives, a difficulty which is only amplified when $\dim(\bgam\cg)\gg1$. %For that purpose we adopt a hybrid strategy where a point, MAP estimate for $\bs{\theta}_{c,MAP}$ is computed and the posterior of the remaining parameters $p(\bgam\cf | \bq^{(1:N)}, \bs{\theta}_{c,MAP})$ is approximated \eqref{eq:post}.
For that reason, we adopt  an approximation based on the Laplace's method~\cite{MacKay_2002}. According to this,  the target posterior $p(\bgam | \bq^{(1:N)} )$  is modeled with a Gaussian (\reffig{fig:laplace})
with mean equal to the MAP estimate $\bs{\theta}_{\mathrm{MAP}}$ and a covariance $\bs{S}$ equal to the inverse of the negative Hessian  of the log-posterior at $\bs{\theta}_{\mathrm{MAP}}$ (see Eqs. \eqref{eq:hessian} and \eqref{eq:priorhessian}).  These two quantities are readily obtained at the last iteration (upon convergence) of the  MC-EM scheme described previously.
Hence:
\be
\bs{S}^{-1} = \left[ \begin{array}{cc} 
 \bs{S}_{cc} & 0 \\
 0 & \bs{S}_{ff}
                     \end{array} \right],
                     \label{eq:laplacecovariance}
\ee
where the block-matrices above  are given by:
%\be
\begin{align}%{l}
 \bs{S}_{cc} &= N \Cov_{\pc(\bQ|\bgam\cg) } [\bs{\phi}(\bQ),\phi_l(\bQ)] + \diag( \left\langle \tau_k \right\rangle_{p(\tau_k|\theta_{\mathrm{c},k})}) \nonumber \\
 \bs{S}_{ff} &= \sum_{i=1}^N \Cov_{\pcf(\bq|\bQ^{(i)},\bgam\cf)  q_i(\bQ^{(i)}) } [\bs{\psi}(\bq,\bQ)].
\label{eq:hessian1}
\end{align}
%\ee
Laplace's approximation can also be interpreted as a second-order Taylor series expansion of the log-posterior at $\bs{\theta}_{\mathrm{MAP}}$.
Some remarks:
\bi
\item For $\theta_{\mathrm{c},k}$ that are effectively turned off when using the ARD prior (i.e.  $\theta_{\mathrm{c},k,\mathrm{MAP}}=0$), $\left\langle \tau_k \right\rangle_{p(\tau_k|\theta_{\mathrm{c},k})} \to \infty$  and thus dominate the corresponding terms in $\bs{S}^{-1}$. As a result, the (approximate) posterior covariance of these $\theta_{\mathrm{c},k}$ approaches $0$.
\item We note that when the number of data points $N \to \infty$, the corresponding terms in $\bs{S}^{-1}$ increase and as a result the (approximate) posterior covariance goes to $0$, as one would expect.  
\ei

Algorithm \ref{alg:bayesianCG} summarizes the basic steps of the scheme advocated.

\begin{algorithm}[H]
\caption{\small Proposed MC-EM scheme}
\begin{algorithmic}[1]
%\algsetup{linenosize=\tiny}
\small
\STATE Initialize $\btheta^0=\{ {\bgam\cg}^0, {\bgam\cf}^0 \}$.
\STATE Select parameters $\{a,\rho,A\}$ for the Robbins-Monro optimization algorithm (\refeqq{eq:rmparam}).
\STATE Step $t=0$
\WHILE{$(not~ \text{converged})$}
\STATE MC-E-step:
\FOR{all $i= 1,\dots,N$}
 \STATE Generate MCMC samples from the (conditional) posterior distribution $q_i(\bQ^{(i)})$ in~\refeqq{eq:qopt}
\ENDFOR
\STATE M-step:
\STATE Construct Monte Carlo gradient estimators $\hat{\mathcal{G}}^{(i)}$ (\refeqq{eq:gradmc})
augmented by the prior gradient (\refeqq{eq:priorgrad}).
% $\calL(q(\bQi|\bqi,\bgam\cg^t,\bgam\cf^t),\bgam\cg^t,\bgam\cf^t)$ with respect to $\bgam\cg$
% (see Equation \eqref{eqn:gradThetaCg}) and $\bgam\cf$ (see Equation \eqref{eqn:gradThetaCf}):
% $\nabla_{\bgam} \calL(q(\bQi|\bqi,\bgam\cg^t,\bgam\cf^t),\bgam\cg^t,\bgam\cf^t)$
\STATE Update the parameters $\btheta$ based on~\refeqq{eqn:RMupdate})
\STATE $t \leftarrow t+1$
\ENDWHILE
\STATE Compute Hessian of the log-posterior~\refeqq{eq:map} at $\bgam_{\mathrm{MAP}}$ (Eqs. \eqref{eq:hessian}, \eqref{eq:priorhessian}) to construct  Laplace's approximation of the posterior $p(\bgam | \bq^{(1:N)})$ (\refeqq{eq:laplacecovariance}).
\end{algorithmic}
\label{alg:bayesianCG}
\end{algorithm}

\section{Numerical Illustrations}
\label{sec:numericIl}
We illustrate the proposed PCG framework in two examples. We particularize the definition of coarse-grained variables $\bQ$ which unavoidably differs from problem to problem. We emphasize   through several illustrations the ability of the proposed method to produce predictive estimates of various macroscopic observables  as well as quantify the predictive uncertainty as a function of the amount of training data $N$ used and the level of coarse-graining i.e. the ratio of fine/coarse variables. We also provide comparisons with the results obtained by employing the relative entropy method.
Finally, we demonstrate how the ARD prior advocated can lead to the discovery of sparse solutions revealing the most prominent feature functions in the coarse potential and  possibly  the most significant types of interactions that this should contain.
Whenever such a  hierarchical prior (ARD) is employed (\refeqq{eqn:ardPrior}) for the parameters $\bgam\cg$ in the coarse potential, the following values were used  for the hyperparameters: $a_0=b_0=10^{-5}$.

% The following section shows two examples how the proposed coarse-graining
% methodology could be applied.
% We first coarse-grain the Ising model considered
% as a two dimensional lattice of magnetic moments in this work.
% % The system is attractive because of low
% % computational cost which allows us to systematically investigate various settings.
% The second example targets the coarse-graining of a water system.
% The general focus of this paper is the predictability of observables
% which are in the case of the Ising model the system's magnetization and 
% correlation between specific lattice sites within the system.
% For the water problem, the focus in on the predictive 
% radial distribution function (RDF) and the angular distribution function (ADF),
% as defined in~\cite{hansen2006}.
% Water serves as solvent in various chemical systems and
% often a tremendous amount of computation time is spent on simulating water
%~\cite{noid2013}. Coarse-graining would increase efficiency
% and with the proposed framework it will allow to quantifying predictive uncertainty.

\subsection{Ising model}
\label{sec:num_ising}
The Ising model serves as abstraction of various physical problems, e.g. for
modeling electromagnetism or lattice gas
systems~\cite{Selinger2015, Ashcroft1976}. It has been the subject of detailed studies and several strategies for coarse-graining in equilibrium~\cite{katsoulakis2003,
katsoulakis_error_2006,katsoulakis_numerical_2008,kalligiannaki_multilevel_2011,chaimovich2011, bilionis2013} and nonequilibrium~\cite{katsoulakis2003} settings. 

% and the formulation of the coarse-scale poteni
% effect on the predictive uncertainty as well as accuracy 
% different definitions of the coarse potential have.
% Different coarse potentials by the meaning of the maximal interaction length considered
% and the order of interaction in the coarse potential.
% Not only the question of model selection is addressed but also the question
% of the uncertainty introduced by describing the
% fine-scale degrees of freedom with various amount of coarse-scale DOFs. 
% Additionally the dependency of the predictive
% uncertainty with respect to limited data is the focus of the following section.
% The observables we use for predictive purpose and for comparison
% is the magnetization $m(\bq)$ and the correlation between the sites $i$
% and $j$, $R_{ij}(\bq)$.
% 
% In the context of the relative entropy CG method, lattice systems were furthermore assessed
% in~\cite{chaimovich2011, bilionis2013}. A review on CG lattice systems
% in non-equilibrium settings is given in~\cite{katsoulakis2003}.

% \subsubsection{Problem Definition}

% \paragraph{Fine-Model}
We consider a periodic, one-dimensional lattice consisting of $n\aaa = 64$  sites. Each site $i$ is associated with a binary variable $x_i, i=1,\dots,n\aaa$ which takes values $\pm 1$.  The $n\aaa-$dimensional vector $\bq=\{ x_i\}_{i=1}^{n\aaa}$  follows $\pf(\bq) \propto \exp\{-\beta U\aaa(\bq)\}$
with the fine-scale potential given by:
\be
 U\aaa(\bq) = -\frac{1}{2} \sum^{L\aaa}_{k=1} J_k \left(
 \underset{
 |i-j|=k
 }{\sum } x_{i} x_{j} 
 \right)
 - \mu \sum^{n\aaa}_{i=1} x_{i}.
\ee
% \be
%  U\aaa(\bq) = -\frac{1}{2} \sum^{L_f}_{m=1} J_k \left(
%  \underset{
%  \begin{subarray}{c}
%  |i-k|=m \\
%  |j-l|=m
%  \end{subarray}
%  }{\sum \sum} x_{ij} x_{kl} 
%  \right)
%  - \mu \sum^{n_f^{1/2}}_{i,j=1} x_{ij}.
% \ee
The expression $|i-j|=k$ implies a summation over all lattice sites $i,j$ that are  $k-$sites apart  (periodic boundary conditions are assumed). The parameter $L\aaa$ expresses the  maximal interaction length. Following ~\cite{katsoulakis2003,trashorras2010,are2008}, we use a decaying interaction
strength $J_k$ with,
\be
  J_k = \frac{K}{k^{a} },
\ee
and the normalization,
\be
 K = \frac{J_0 }{L\aaa^{a-1} \sum_{k=1}^{L\aaa} k^{-a}}.
\ee
Finally, the parameter  $\mu$ denotes the external field.

The values $A=25$, $\alpha=0.15$, and $\rho= 0.75$ were used for the Robbins-Monro
updates (\refeqq{eqn:RMupdate}) based on suggestions given in~\cite{bilionis2013}.
We used $m=170$
samples for the MCMC estimates of the gradients in Eqs.~\eqref{eq:gradientmcest} and~\eqref{eq:gradmc}.

% The values $\alpha=0.15$, and $\rho=0.75$ for the Robbins-Monro updates,
% given in~\refeqq{eqn:RMupdate}, are used in the
% following numerical studies.
% The choice of $A$ arises from estimating the maximal iterations
% until convergence is expected. Here, we expect the optimization scheme to
% converge after maximal $250$ iterations and use $10\%$ of this
% value for $A=0.10 \cdot 250=25$, based on suggestions given in~\cite{bilionis2013}.
% In this example, we use $m=170$ samples for the MCMC estimates of the gradients
% in Eqs.~\eqref{eq:gradientmcest} and~\eqref{eq:gradmc}.

\subsubsection{Observables}
As pointed out previously, the framework proposed readily allows for reconstructions of the whole fine-scale description and therefore probabilistic predictions can be computed for any observable. For comparative purposes, we focus on two  such quantities. The first one is the magnetization $m(\mu)$ and its dependence on the external field parameter $\mu$.  This is associated with the following observable:
\be
 a^{(m)}(\bq) = \frac{1}{n\aaa} \sum_{i} x_{i},
 \label{eqn:obsMag}
\ee
i.e. $m(\mu)=\E_{p\aaa(\bq)}[ a^{(m)}(\bq)]$. The second quantity is the correlation  $R(k)$ at various separation distances $k$ which captures second-order statistical information of the fine-scale configurations. The corresponding observable is:
\be
 a^{(R)}(\bq; k) = \frac{1}{n\aaa} 
 \underset{
 \begin{subarray}{c}
 |i-j|=k% \\
 %|j-l|=p
 \end{subarray}
 }{\sum} x_{i} x_{j},
%  a^{R_{op}}(\bq) = \frac{1}{n_f} \sum_{i} 
%  \underset{
%  \begin{subarray}{c}
%  |i-j|=k% \\
%  %|j-l|=p
%  \end{subarray}
%  }{\sum \sum} x_{i} x_{j}.
  \label{eqn:obsCorr}
\ee
i.e. $R(k)=\E_{p\aaa(\bq)} [a^{R}(\bq; k)]$.

\subsubsection{Coarse-variables $\bQ$ and coarse-to-fine map}
While the framework proposed offers great flexibility in the definition of the coarse
variables $\bQ$, in this work we make perhaps the most intuitive choice by assuming that
$\bQ$ are (also) binary and have a {\em local} dependence on $\bq$. 
This offers a direct  appraisal on the level of coarse-graining as well as a natural,
visual interpretation of the coarse variables and their role.

\begin{figure}[tb]
\centering
\includegraphics[width=0.7\textwidth]{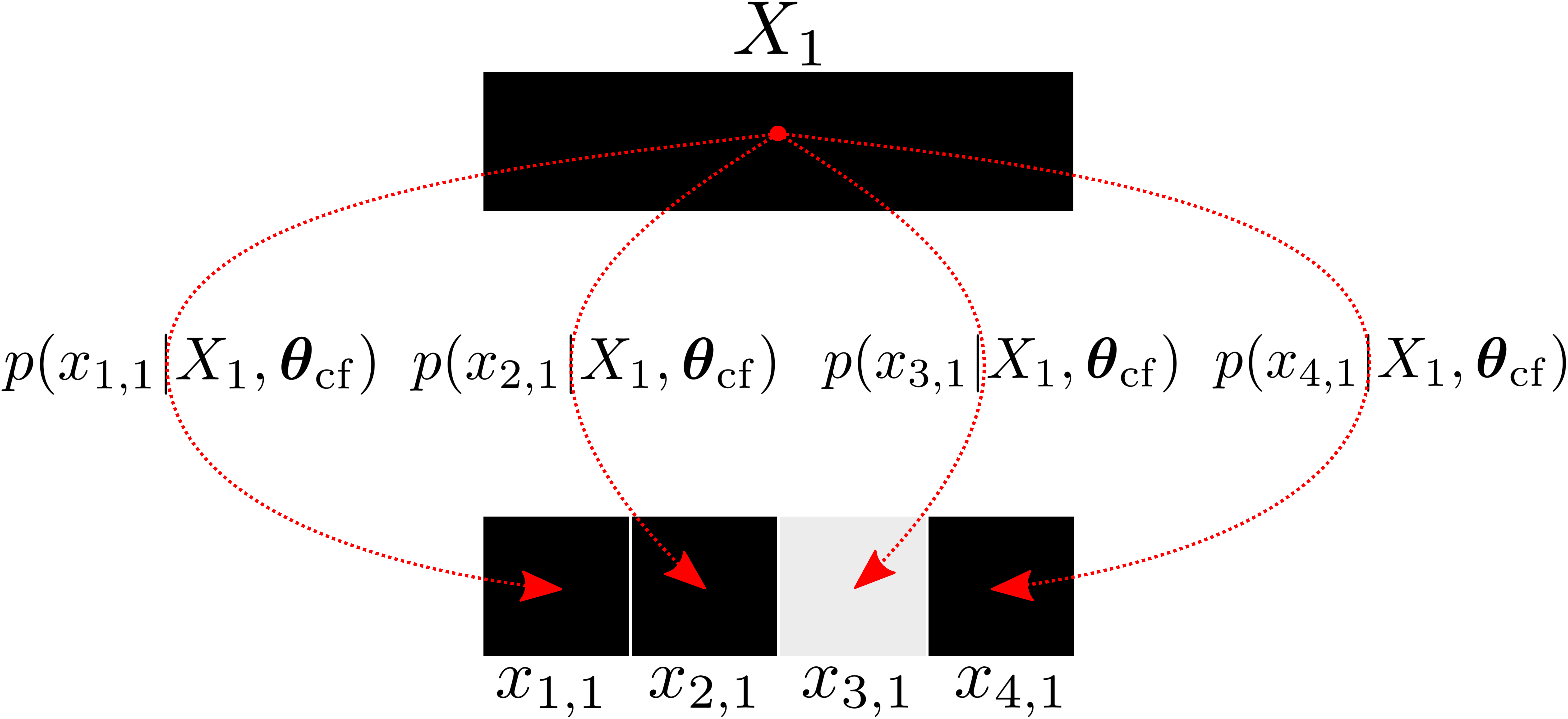}
\caption{Probabilistic coarse-to-fine map $p\cf(\bq|\bQ,\bgam\cf)$. The coarse-variable $X_1$
is e.g. associated with $x_{1\dots 4,1}$ fine-scale variables through the probabilistic coarse-to-fine map $p\cf$ (\refeqq{eq:p0}).
Each $x_{s,1}$ is \emph{conditionally} independent from the other.}
\label{fig:ising_pcf}
\end{figure}

In particular, we assume that each coarse variable $X_I, I=1,\ldots,n\cg$ is associated with a one-dimensional lattice  that is a coarser version of the fine-scale one, i.e. with $n\cg<n\aaa$ sites (\reffig{fig:ising_pcf}). We can construct such descriptions by regularly coarsening  by a factor of 2 such that $n\cg= n\aaa/{2^a}$, with $a=1,\ldots,A$. We assume that each $X_I$ (parent) is associated with $S=\frac{n\aaa}{n\cg}$ fine-scale variables (children) denoted by $x_{(I-1)S+s}=x_{s,I}$ (where $s=1,\ldots,S$ , \reffig{fig:ising_pcf}).
We define a  coarse-to-fine map of the form:
%\be
\begin{align}%{ll}
 p\cf(\bq | \bQ, \bgam\cf) & = \prod_{I=1}^{n\cg} \prod_{s=1}^S p(x_{s,I}|X_I, \bgam\cf) \nonumber \\
 & = \prod_{I=1}^{n\cg} \prod_{s=1}^S p_0^{\frac{1+x_{s,I} X_{I}}{2}} ~(1-p_0)^{\frac{1-x_{s,I} X_{I}}{2}} \nonumber \\
 & = p_0^{ \sum_{I=1}^{n\cg} \sum_{s=1}^S \frac{1+x_{s,I} X_{I}}{2} } ~(1-p_0)^{\sum_{I=1}^{n\cg} \sum_{s=1}^S \frac{1-x_{s,I} X_{I}}{2} }.
\label{eq:p0}
\end{align}
%\ee
The expression above implies that each  $x_{s,I}$ is {\em conditionally}  independent and follows a Bernoulli distribution with probability $p_0$ of being of the same value as its parent $X_I$, and probability $(1-p_0)$ of having the opposite spin.
We emphasize that this does not imply that  $x_{s,I}$ are also independent. In fact they will be correlated as a result of the dependencies between the  coarse variables $\bQ$ induced by the coarse model $p\cg$   which is discussed in the next subsection. 
The density $p\cf$ above belongs to the exponential family (Section \ref{sec:expon}) and is controlled by a single parameter, $p_0 \in [0,1]$. Given the symmetry of the model, we restrict $p_0 \in [0.5,1]$. To ensure that it  stays within this interval during the MC-EM updates (Algorithm \ref{alg:bayesianCG}), we operate instead on $\theta\cf \in \RR$ defined as follows:
\be
p_0=\frac{1}{2} (1+\frac{1}{ 1+e^{-\theta\cf} }).
\ee
The derivatives needed for the updates of the EM-scheme in~\refeqq{eq:gradmc} and~\refeqq{eq:hessian} are: 
%\be
\begin{align}%{ll}
 \frac{\pa \log p\cf}{\pa \theta\cf } & = \frac{\pa \log p\cf}{\pa p_0} \frac{\pa p_0}{\pa \theta\cf}, \nonumber \\
 \frac{ \pa^2 \log p\cf}{\pa \theta\cf^2} & =\frac{ \pa^2 \log p\cf}{\pa p_0^2} \left( \frac{\pa p_0}{\pa \theta\cf} \right)^2+\frac{\pa \log p\cf}{\pa p_0}  \frac{\pa^2 p_0}{\pa \theta\cf^2},
\end{align}
%\ee
where:
%\be
\begin{align}%{ll}
 \frac{\pa \log p\cf}{\pa p_0} & = \frac{\psi(\bq, \bQ)}{p_0}-\frac{1-\psi(\bq, \bQ)}{1-p_0}, \nonumber \\
 \frac{ \pa^2 \log p\cf}{\pa p_0^2} & =-\frac{\psi(\bq, \bQ)}{p_0^2}-\frac{1-\psi(\bq, \bQ)}{(1-p_0)^2},
\end{align}
%\ee
and  $\psi(\bq, \bQ)=\sum_{I=1}^{n\cg} \sum_{s=1}^S \frac{1+x_{s,I} X_{I}}{2}$. 
% Following the proposed predictive coarse-graining scheme, we need to define a probabilistic mapping fine variables $\bq$
% from the coarse-scale to the fine-scale $\pcf(\bq|\bQ,\bgam\cf)$. By specifying the mapping
% $\pcf(\bq|\bQ,\bgam\cf)$, the coarse variables are defined. The coarse-variables $\bQ$ follow the
% statistics of $\pc(\bQ|\bgam\cg)$ and describing the state of $n_c$ coarse-scale lattice sites.
% 
% The mapping from coarse to fine follows a Bernoulli distribution.
% A coarse variable $X_r$ with $r \in \{1,\dots,n_c\}$ corresponds to $S = n_f/n_c$ fine variables
% $x_{r,s}$ with $s \in \{1, \dots, n_f/n_c\}$
% (see Figure \ref{fig:ising_pcf}).
% We denote the level of coarse-graining as $l_c$ which is the ratio between
% the amount of fine-scale variable to the coarse-scale variables:
% \be
%  l_c = \frac{n_f}{n_c}.
% \ee
% The mapping assumes that the fine-variables $x_{r,s}$
% belonging to one and the same coarse-scale variable $X_r$
% are conditionally independent and thus,
% \be
%  p(x_{r,s}|X_r, \bgam\cf) = \theta\cf^{\frac{1+x_{r,s} X_{r}}{2}} (1-\theta\cf)^{\frac{1-x_{r,s} X_{r}}{2}}.
% \ee
% Therefore the distribution of the fine-scale variables $\bq_r =(x_{r,1}, \dots, x_{r,S})$
% of the corresponding coarse-variable $r$ is,
% \be
%  \pcf(\bq_r|\bQ_r, \bgam\cf) = \prod_s^S  p(x_{r,s}|X_r, \bgam\cf),
% \ee
% and the whole configuration $\bq$ given a coarse-configuration $\bQ$ follows to,
% \be
%  \pcf(\bq|\bQ, \bgam\cf) = \prod_r^{n_c} \pcf(\bq_r|\bQ_r, \bgam\cf).
% \ee
% By prescribing the probabilistic mapping, the coarse-variables $\bQ$ are
% defined, whereas the mapping is parametrized by $\bgam\cg$ and optimized
% with respect it.

\subsubsection{Coarse model}

The coarse potential $U\cg(\bQ, \bgam\cg)$ employed includes first-, second- and third-order interactions with 
various interaction lengths. In particular, we prescribe: 
\begin{align}
 U\cg(\bQ,\bgam\cg) = &-\frac{1}{2} \bigg\{
 \theta^{(1)}\cg \sum_{i} X_{i}  
 + \sum_{i}  X_{i}  \sum_k^{L\cg^{(2)}}
 \theta^{(2)}_{\mathrm{c},k}  X_{i \pm k} + \sum_{i}  X_{i}
  \underset{
 \begin{subarray}{c}
 k=1 \\
 l=1
 \end{subarray}
 }{\sum^{L\cg^{(3)}}}
  \theta^{(3)}_{\mathrm{c},kl} X_{i\pm k} X_{i\pm k \pm l}
\bigg\} \nonumber  \\
  &- \mu \sum_{i} X_{i}.
  \label{eqn:isingUc}
\end{align}
The parameters $L\cg^{(2)}$ and $L\cg^{(3)}$ denote the maximal second- and third- order interactions, respectively. With superscripts (1), (2),  (3) we distinguish between the coarse potential parameters $\bgam\cg$ that are associated with the first, two-body and third-body interactions, respectively. These parameters determine also the number of $\bgam\cg$ which is equal to $1+L\cg^{(2)}+(L\cg^{(3)})^2$.
% The maximal considered interaction length in the coarse potential is denoted as $L_c$
% with a superscript specifying the order of interaction ((1) linear, (2) two-body or (3) three-body).
% The same  superscript-denotation is applied for the parameters $\bgam\cg$.

% Respectively we use $N_{data}$ samples for fine-scale simulation, 
% $N_c$ samples from the coarse-distribution and $N_q$ samples for approximating
% $q(\bQi|\bqi, \bgam\cg, \bgam\cf)$ for each distribution $i$ to the corresponding data-point $i$.
% The samples are collected after dropping the burn-in period and also dropping
% % samples in between for ensuring samples with low correlation.
% 
% 
% 
% Results of the proposed predictive coarse-graining method with predictions
% of the relative entropy method and with the prediction of the fine-scale simulations,
% which we refer as the 'truth' at various $\mu_i$ estimated by 2500 samples from $p\aaa(\bq)$.

% \subsubsection{Comparison with Relative Entropy CG}

In order to compare the  proposed method  with the relative entropy method, 
as briefly summarized  in Section \ref{sec:esm}, a deterministic fine-to-coarse mapping $\calR(\bq)$ is needed.  We note that in~\cite{chaimovich2011, bilionis2013} such efforts have been made by ``coarse-graining" the interactions rather than the degrees
of freedom i.e. $\bq\equiv \bQ$. In order to truly assess the performance in cases where the coarse variables are of lower dimension and of the same type as in this study (i.e. binary), we prescribe the following map:
\be
    X_I= 
\begin{cases}
    +1,&  \frac{1}{S}\sum_s^S x_{s,I} \ge 0\\
    -1,&  \frac{1}{S}\sum_s^S x_{s,I} < 0. %\\
 %   U(-1,+1),              & \text{otherwise}.
\end{cases}
\label{eq:fine_to_coarse}
\ee
This implies a ``majority rule" where the label of the parent $X_I$ is determined by the majority of the children. The same model as in~\refeqq{eqn:isingUc} was used for the coarse potential. In order to reconstruct the fine configurations $\bq$  and estimate the observables of interest from the coarse description $\bQ$, a consistent sampling was performed from the conditional in~\refeqq{eq:cond} for the $\calR$ above.
% 
% The mapping from fine-scale to coarse-scale assumes the coarse variable $X_r$ takes the value
% of the majority of the corresponding fine-variables $x_{r,s}$. If in one coarse
% cell $X_r$ are the same amount of fine-variables having positive and negative spin, we assign
% the coarse-variable $X_r$ uniformly with $X_r \sim U(-1,+1)$.
% The novel part of the predictive CG method is the ability of reconstructing fine-scale
% realizations for a given coarse state $\bQ$. For comparing the predictions on the fine-scale
% with the relative entropy method we introduce a pseudo mapping from the coarse-scale back to the
% fine-scale.  Assumed that $\dim(\bq)/\dim(\bQ)=2$, for a given coarse variable
% $X_r$ the reconstruction follows $\{ x_{r,1},x_{r,2} \} \sim U(\{ X_r, X_r\}, \{ X_r, -X_r\})$.

% % The relative entropy method
% does not provide a posterior distribution $p(\bgam\cg, \bgam\cf|\bqd)$
% but the algorithm for calculating observables given by
% Algorithm \ref{alg:calcObservable} is still applicable while skipping the sampling
% from $p(\bgam\cg, \bgam\cf|\bqd)$.

\subsubsection{Results}

\begin{figure}
\centering
\includegraphics[width=0.97\textwidth]{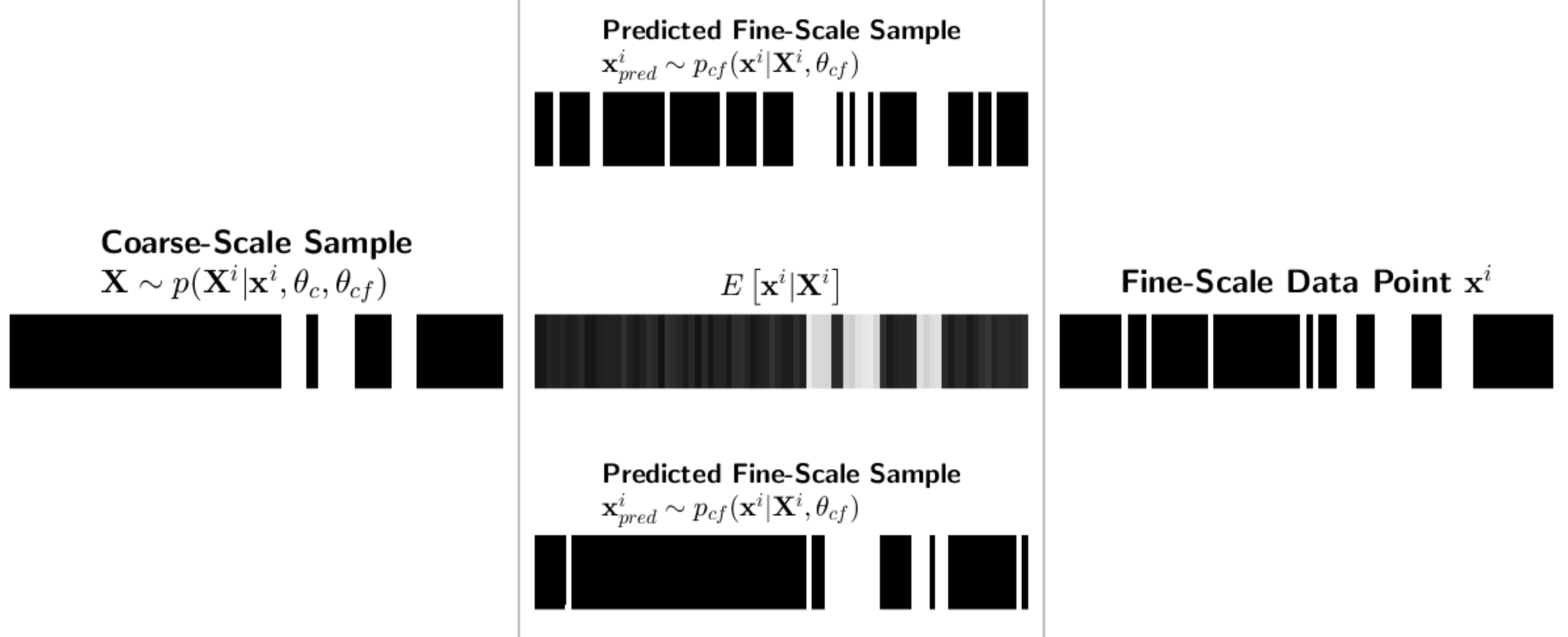}
\caption{For the  FG datum $\bqi$ (right), the image on the left shows a sample from the posterior of  the CG  $\bQ^{(i)}$ (upon convergence of the Algorithm \ref{alg:bayesianCG}) i.e. one of the possible {\em pre-images} of $\bqi$. The three images in the center illustrate the predictions/reconstructions of the fine-scale: the top and bottom are samples drawn from the $p\cf$ and the center is the expected FG configuration according to $p\cf$.}
\label{fig:ising_anim}
\end{figure}

\begin{figure}
\centering
\includegraphics[width=0.6\textwidth]{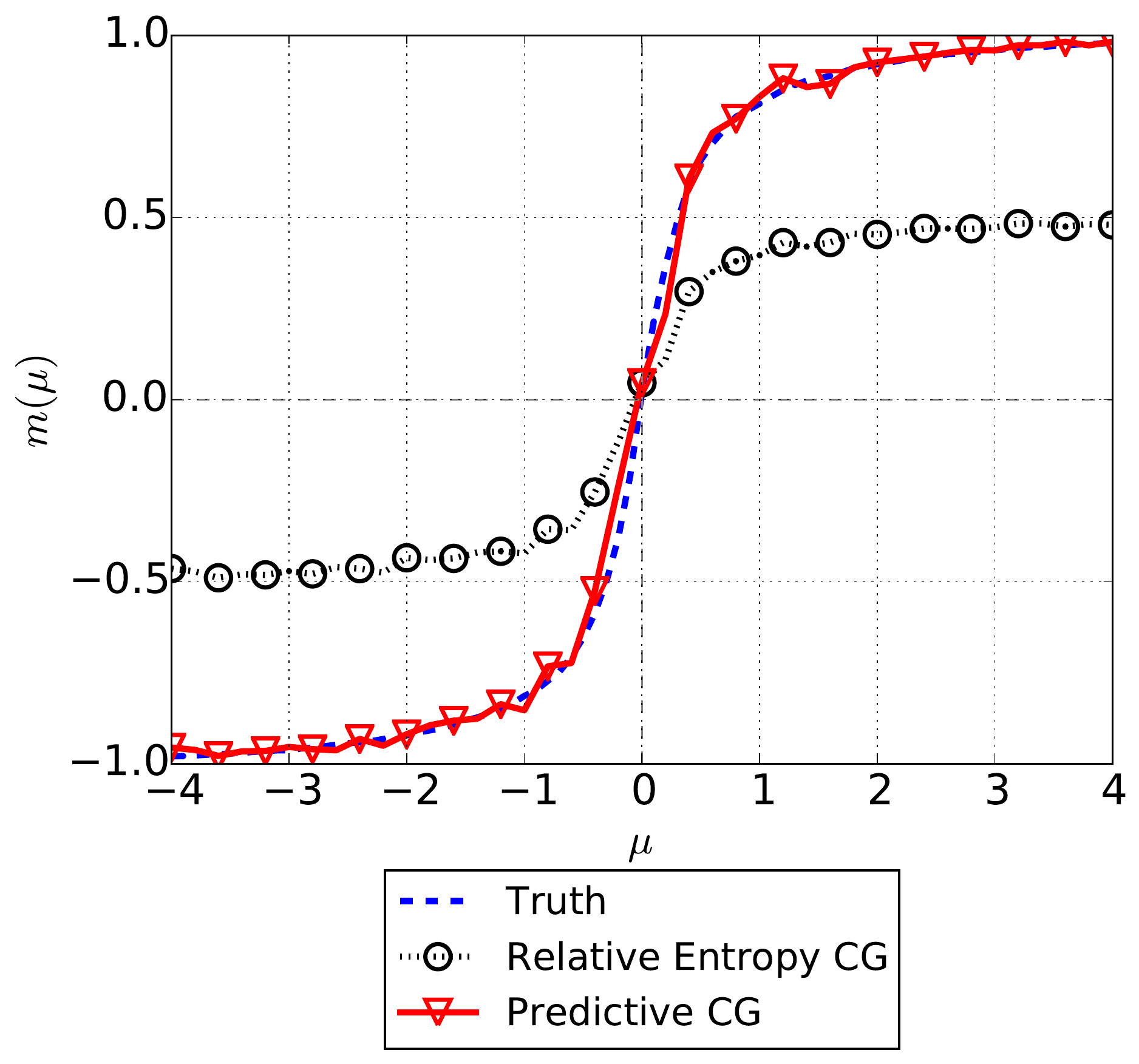}
\caption{Comparison of the reference magnetization (computed with the FG configuration) with posterior mean of predictive CG and relative entropy CG.
$N=20$, $\frac{n\aaa}{n\cg}=2$, $L\cg^{(2)} = 15$, $L\cg^{(3)} = 3$.}
\label{fig:ising_shell_combined}
\end{figure}

The ensuing  results are based on the following values for the fine-scale potential:
$J_0 = 1.5$, $a=0.8$, $L\aaa=8$, $\beta=0.3$, $n\aaa=64$.
We generated data from the fine scale model for each of   $41$ values of the external field $\mu$,  equidistantly  distributed  within $[ -4, 4 ]$.
A different CG model is trained for every  $\mu$ value considered. One could also envision introducing a dependence of the CG model's components  on $\mu$ which would allow a single model to be inferred and to be used for making predictions even for values of $\mu$ not contained in the data.
Figure~\ref{fig:ising_anim} provides some insight on the role of the CG variables, their posterior and their ability to represent/reconstruct the FG configuration.
% Fine-scale, coarse-scale, as well as the posterior
% distribution  $q(\bQi|\bqi, \bgam\cg, \bgam\cf)$ 
% occurring in the expectation-maximization scheme (see Algorithm \ref{alg:bayesianCG})
% are approximated by the MC estimate using a sequential Gibbs-sampling (\cite{geman1984}) simulator.
Figure~\ref{fig:ising_shell_combined} compares point-estimates of the predicted magnetization
as obtained with the proposed method (red) and  the relative entropy method
(for  fine-to-coarse mapping
as given in ~\refeqq{eq:fine_to_coarse}).
% \red{
% The proposed method has the ability of predicting fine-scale configurations with the 
% probabilistic reconstruction or coarse-to-fine mapping $p\cf(\bq|\bQ, \btheta\cf)$ within the
% generative model (defined in \refeqq{eq:p0}),
% reconstructive abilities are not provided in relative entropy CG which operates fully on the coarse-scale.
% In order to compare the observable  $a^{(m)}(\bq)$ (see \refeqq{eqn:obsMag})
% on the fine-scale with relative entropy CG, a consistent sampling of fine-scale configurations was introduced.
% }
While one can claim that better results can be obtained with a different set of CG variables (\refeqq{eq:fine_to_coarse}),
the point in this comparison is to demonstrate the information loss that takes place which can lead to poor predictions
when not quantified.
Given the same amount of training data $N$, the information loss in the relative entropy method is driven by
the not adjusted map in the consistent density of the fine-scale variables $p_{\calR}(\bq)$ denoted in \refeqq{eq:pR}
compared to PCG.
While in PCG the probabilistic map $p\cf(\bq|\bQ,\bgam\cf)$
(\refeqq{eq:p0}) is parametrized and optimized within the parametric family of $p\cf$.
We note further that the relative entropy method can lead to good approximations of the potential of mean force, and as a result, accurate estimates (as shown earlier) of  expectations of observables that depend solely on $\bs{X}$.
We could therefore select $\bs{X}$ in such a way that the magnetization is only a function of $\bs{X}$ in which case the result of the relative entropy method would probably be good.
If however another expectation was sought (that does not depend on the current $\bs{X}$) a new set of $\bs{X}$ would need to be defined and a new CG model would need to be retrained. 

\begin{figure}[htbp]
\centering
\begin{floatrow}
\ffigbox[\FBwidth]
{
\centering
\subfloat[Parametrization of two-body interactions $\theta_{\mathrm{c},k}^{\mathrm{(2)}}$]{\includegraphics[width=0.45\textwidth]{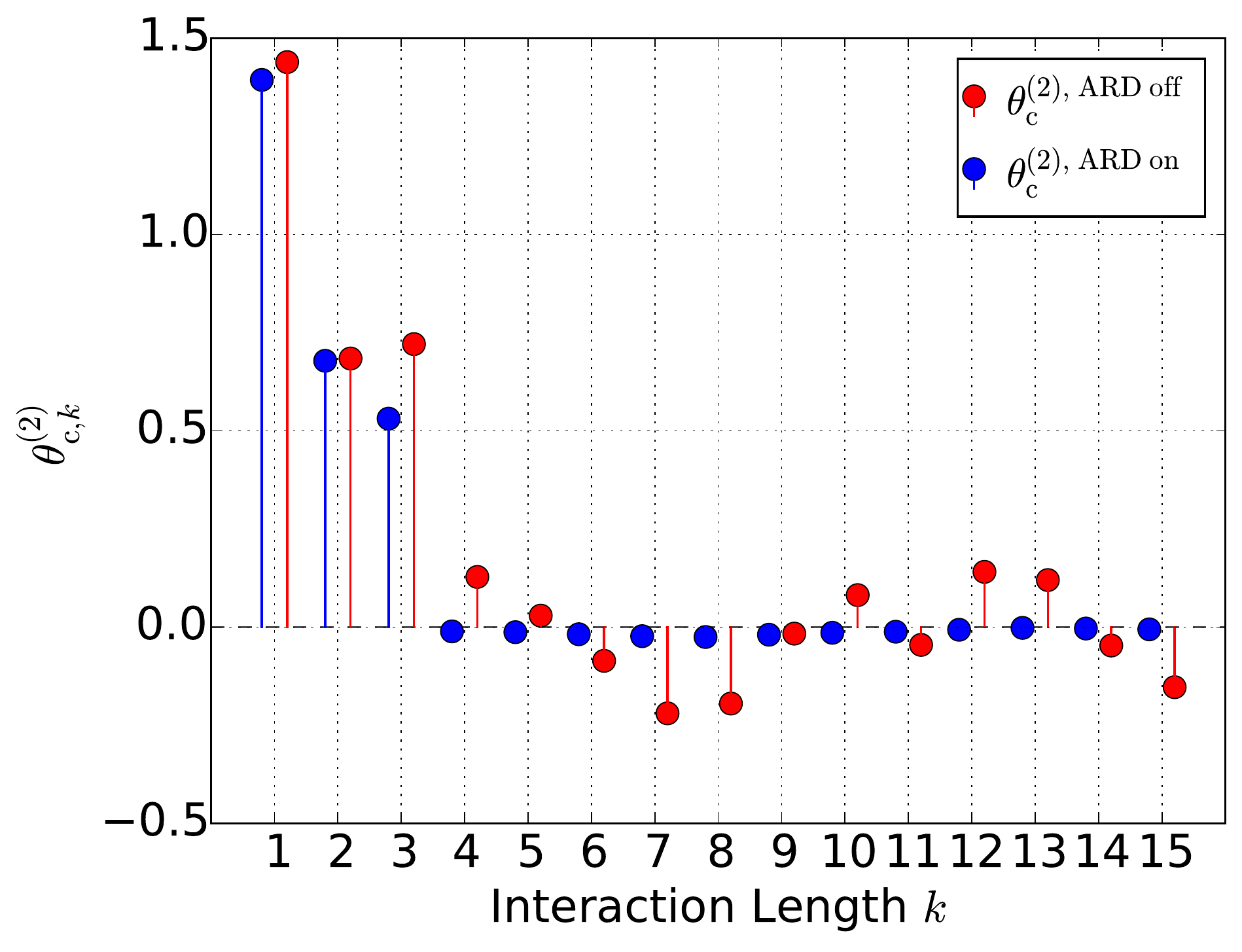}
\label{fig:ising_par_twoBody}
}
\quad
\centering
\subfloat[Parametrization of three-body interactions $\theta_{\mathrm{c},kl}^{\mathrm{(3)}}$]{\includegraphics[width=0.45\textwidth]{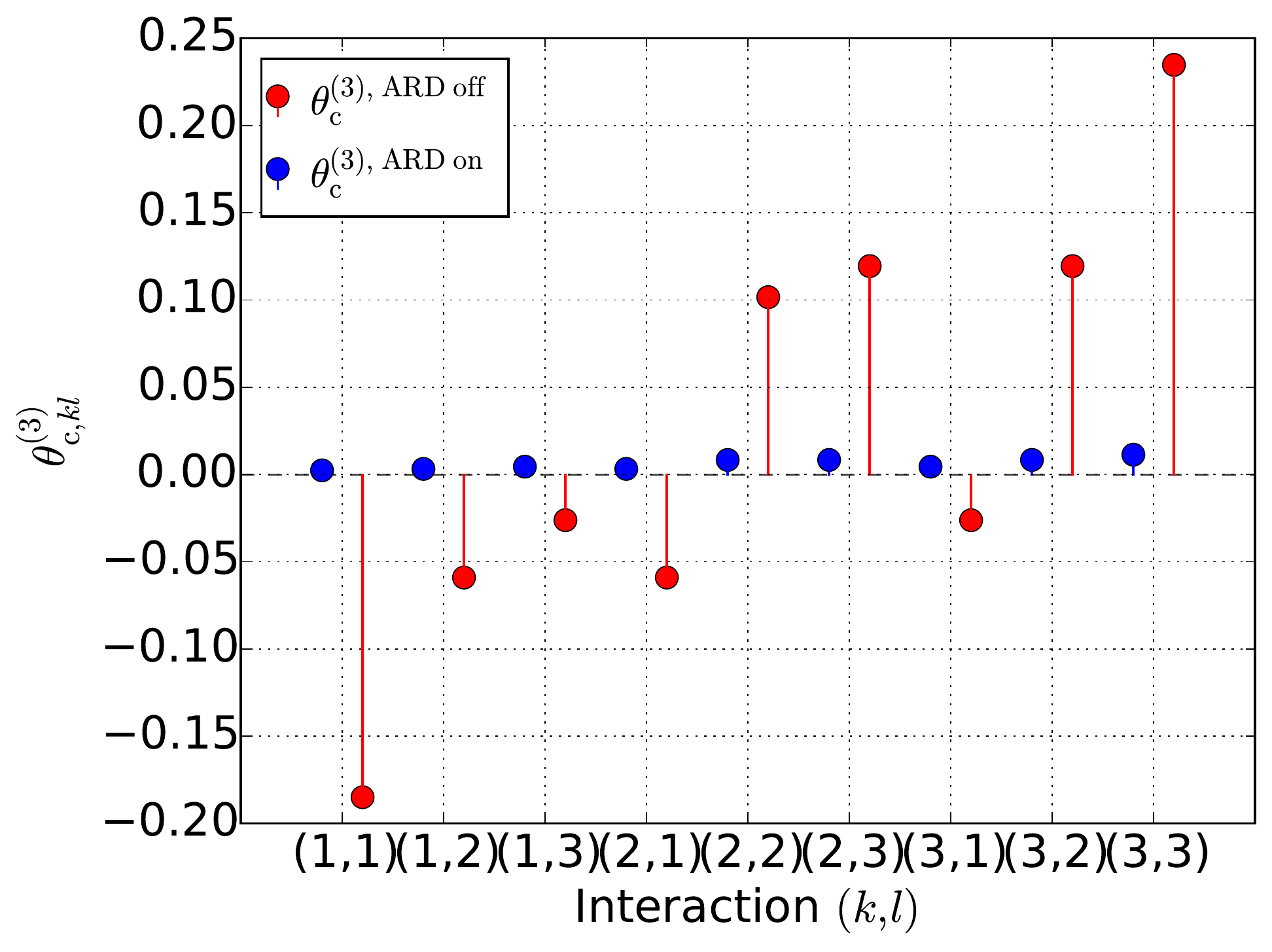}
%original version found under {figures/water/10_dp_lj_cs_50_slow_0001_ard_5_f_1_20/plt_par_freq_sin}
\label{fig:ising_par_threeBody}
}
}{
\caption{Parametrization of two- and three-body interactions at $\mu=0.0$, obtained with and without
ARD prior. A sparse solution is obtained with active ARD prior at the same 
predictive accuracy (see \reffig{fig:ising_mag_ard}).
$N=20$, $\frac{n\aaa}{n\cg}=2$, $L\cg^{(2)} = 15$, $L\cg^{(3)} = 3$.}
\label{fig:ising_par}
}
\end{floatrow}
\end{figure}

\begin{figure}
 \includegraphics[width=0.6\textwidth]{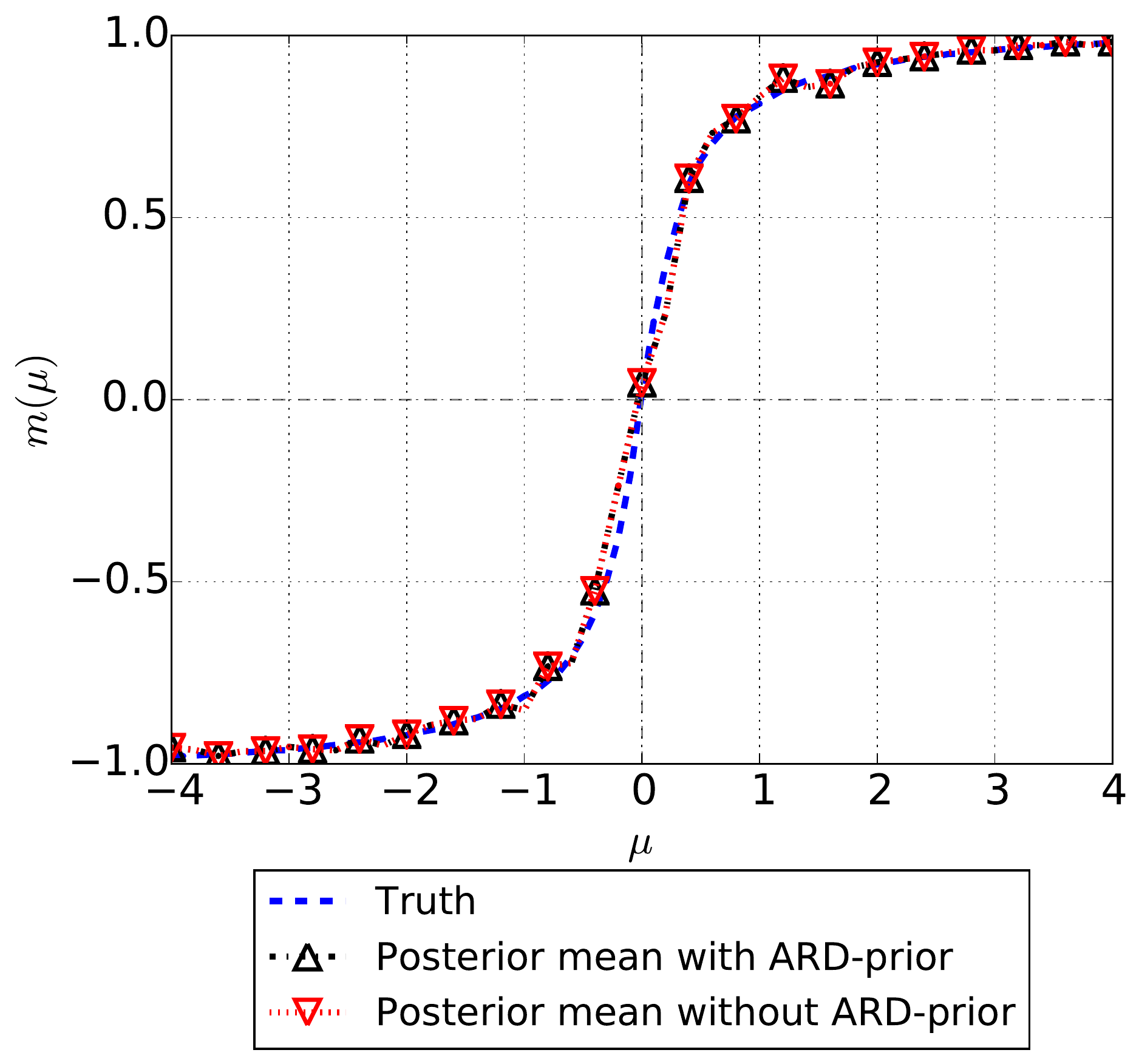}
 \caption{Comparison of predicted magnetization with and without ARD prior.
 $N=20$, $\frac{n\aaa}{n\cg}=2$, $L\cg^{(2)} = 15$, $L\cg^{(3)} = 3$.}
 \label{fig:ising_mag_ard}
\end{figure}

When $\frac{n\aaa}{n\cg}=2$, $L\cg^{(2)}=15$, $L\cg^{(3)}=3$, the total number of unknowns parameters $\bgam\cg$ in the potential $U\cg$ is  $1+L\cg^{(2)}+(L\cg^{(3)})^2=25$.  This is not a particularly large number, but we demonstrate nevertheless the effect of the sparsity enforcing prior in  \reffig{fig:ising_par} when $N=20$ data points are used. 
In the absence of the ARD prior  (\refeqq{eqn:ardPrior}), all $\bgam\cg$ are non-zero and the corresponding feature functions are all active (\refeqq{eqn:isingUc}). On the contrary, when the ARD prior is employed, the learning scheme identifies only 3 non-zero $\bgam\cg$.   Interestingly these are associated with two-body interactions up to separation 3 whereas all
other terms corresponding to two- and three-body interactions are found to be unnecessary,
despite having equal predictive accuracy as shown in \reffig{fig:ising_mag_ard} where point estimates of the magnetization are plotted (with and without the ARD prior).
% 
% Using an ARD-prior, as defined in~\refeqq{eqn:ardPrior}, is advantageous
% since the salient coarse potential features are detected as depicted in
% \reffig{fig:ising_par}. A sparse solution is obtained for two-body
% (\reffig{fig:ising_par_twoBody})
% and three-body interactions (\reffig{fig:ising_par_threeBody}) are not of importance
% for describing the data
% in the given setting.

\begin{figure}
\begin{minipage}{.5\linewidth}
\centering
\subfloat[$N=10$]{\label{fig:isingmag_10dp}\includegraphics[width=\textwidth]{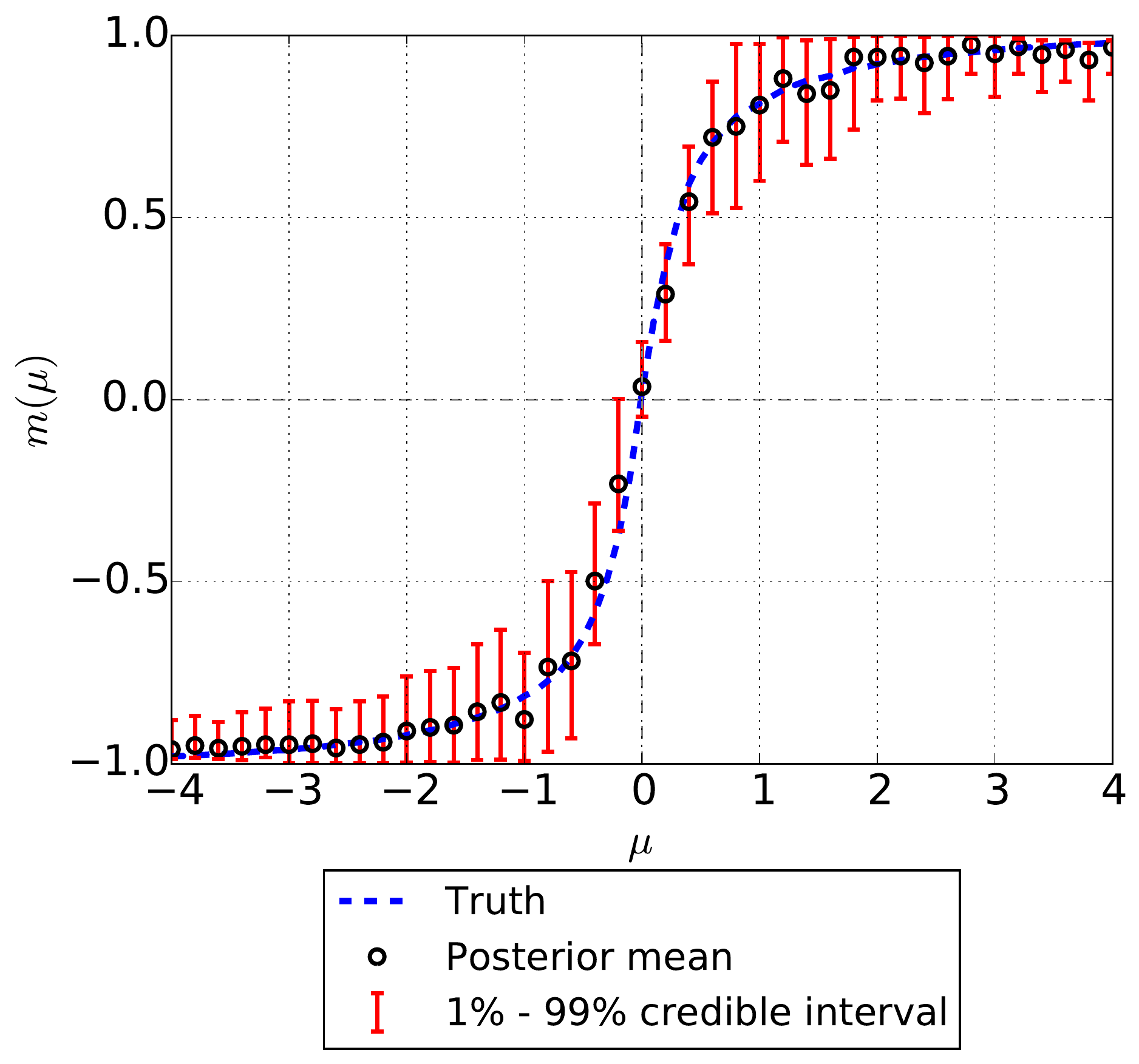}}%ising_VarMu_Sf_10_levCG_2_pexp_1_Lc_15_LcE_1_LcL_1_LcT_3_plt_mag.pdf}}
\end{minipage}%
\begin{minipage}{.5\linewidth}
\centering
\subfloat[$N=20$]{\label{fig:isingmag_20dp}\includegraphics[width=\textwidth]{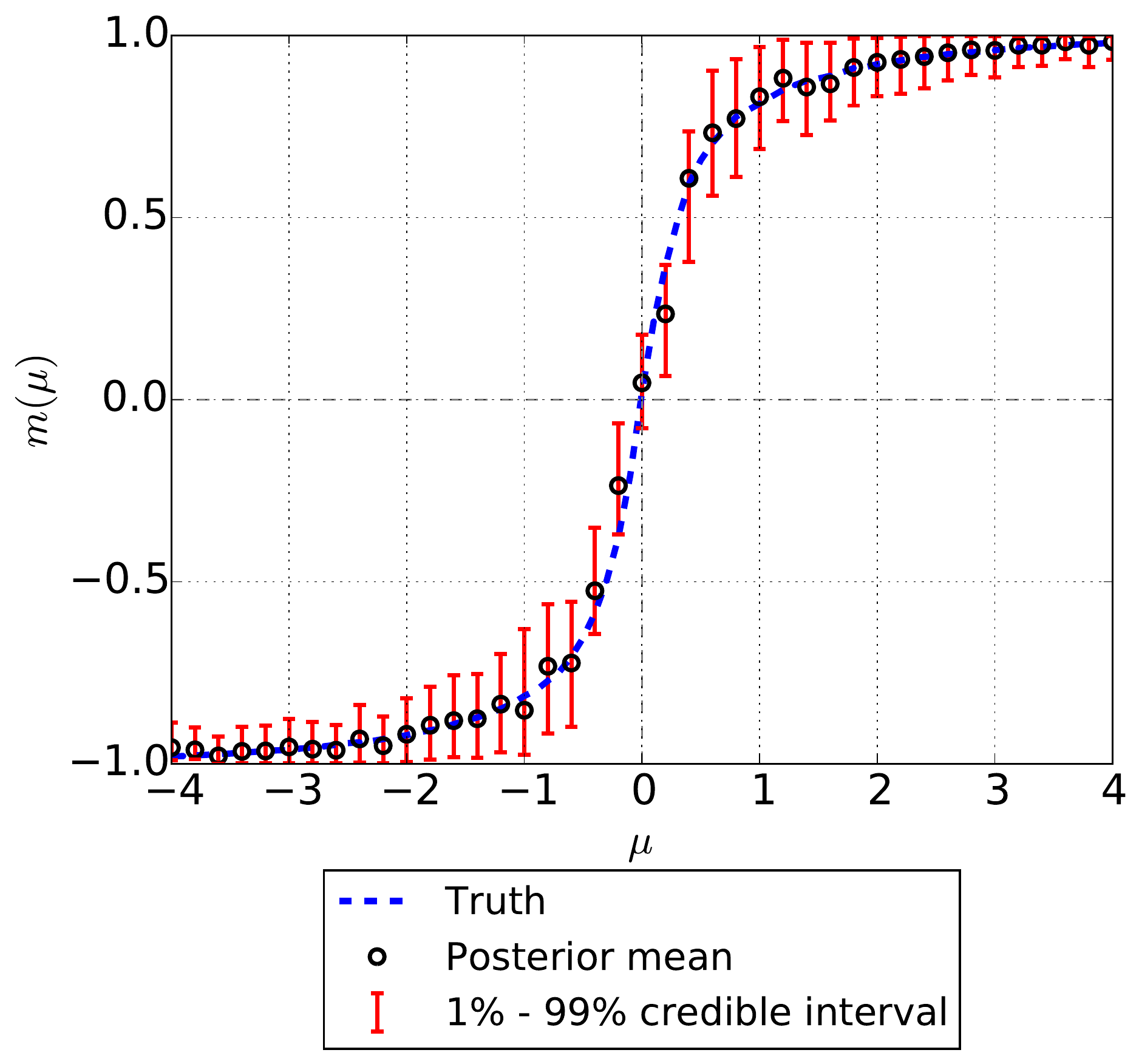}}
\end{minipage}\par\medskip
\centering
\subfloat[$N=50$]{\label{fig:isingmag_50dp}\includegraphics[width=0.5\textwidth]{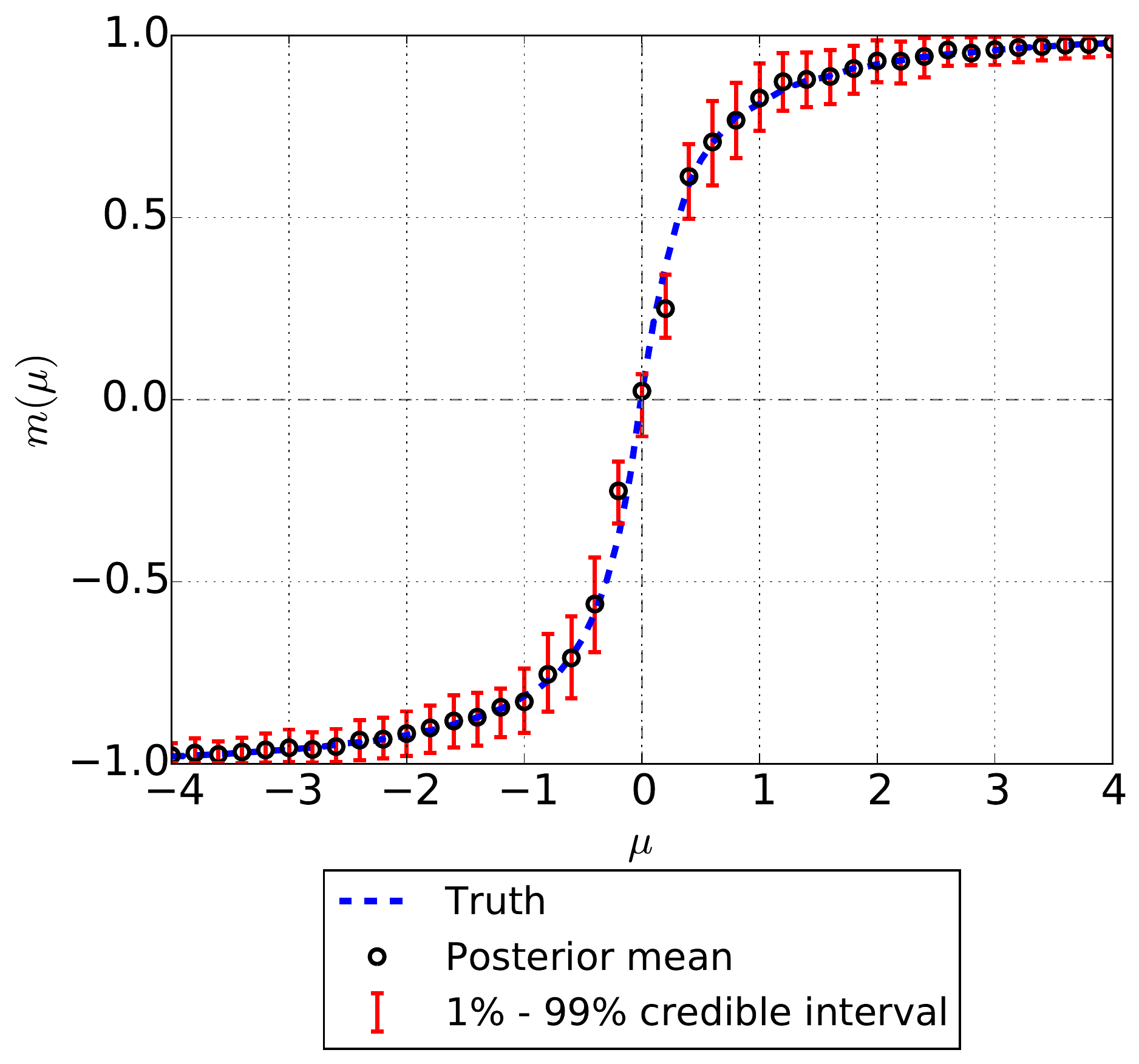}}
\caption{Comparison of the reference magnetization (computed with the FG configuration) with posterior mean and credible intervals corresponding to $1\%$ and $99\%$ posterior quantiles.
 $N=20$, $\frac{n\aaa}{n\cg}=2$, $L\cg^{(2)} = 15$, $L\cg^{(3)} = 3$.}
\label{fig:isingmag}
\end{figure}

\begin{figure}
\begin{minipage}{.5\linewidth}
\centering
\subfloat[$N=10$]{\label{fig:isingcorr_10dp}\includegraphics[width=\textwidth]{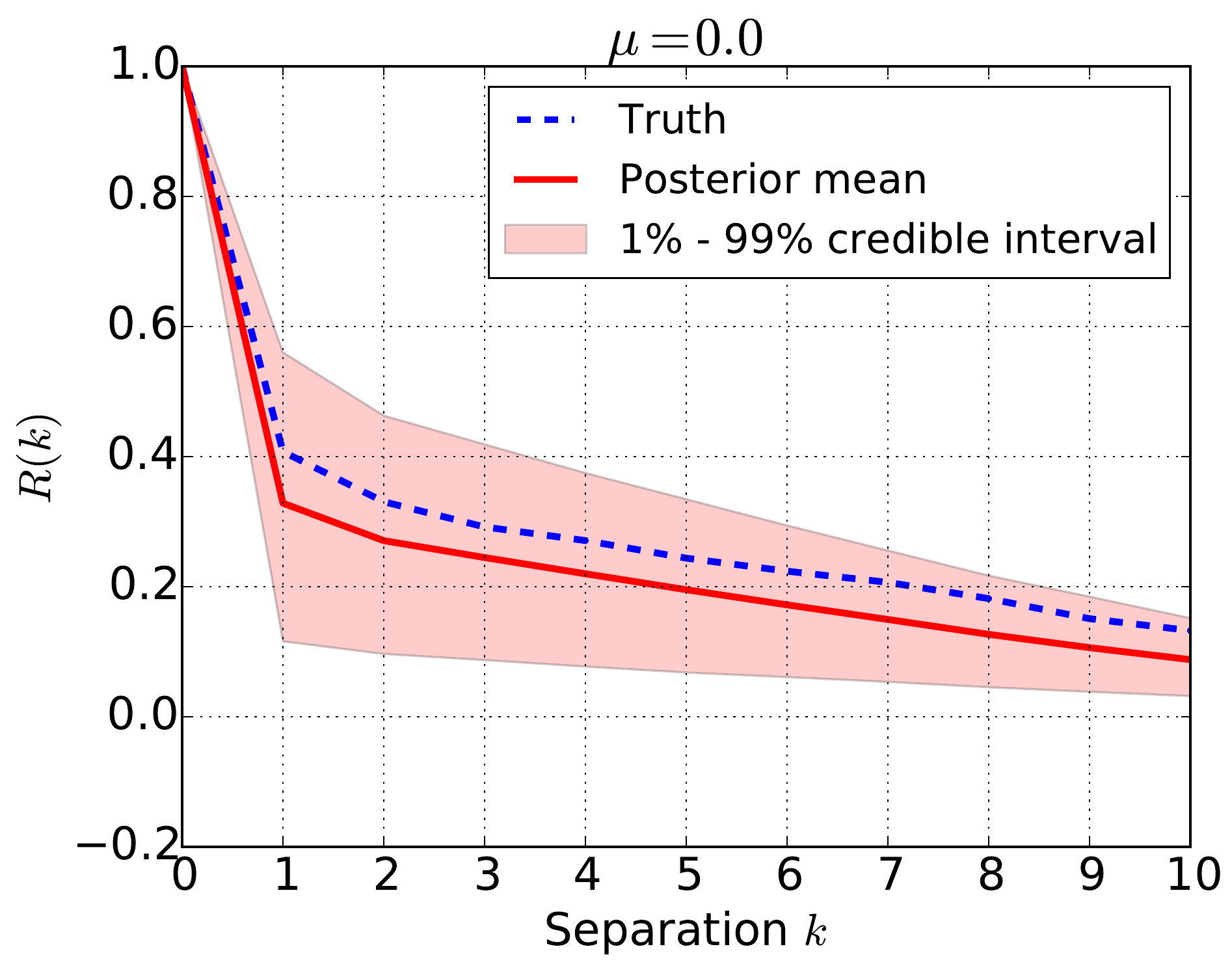}}
\end{minipage}%
\begin{minipage}{.5\linewidth}
\centering
\subfloat[$N=20$]{\label{fig:isingcorr_20dp}\includegraphics[width=\textwidth]{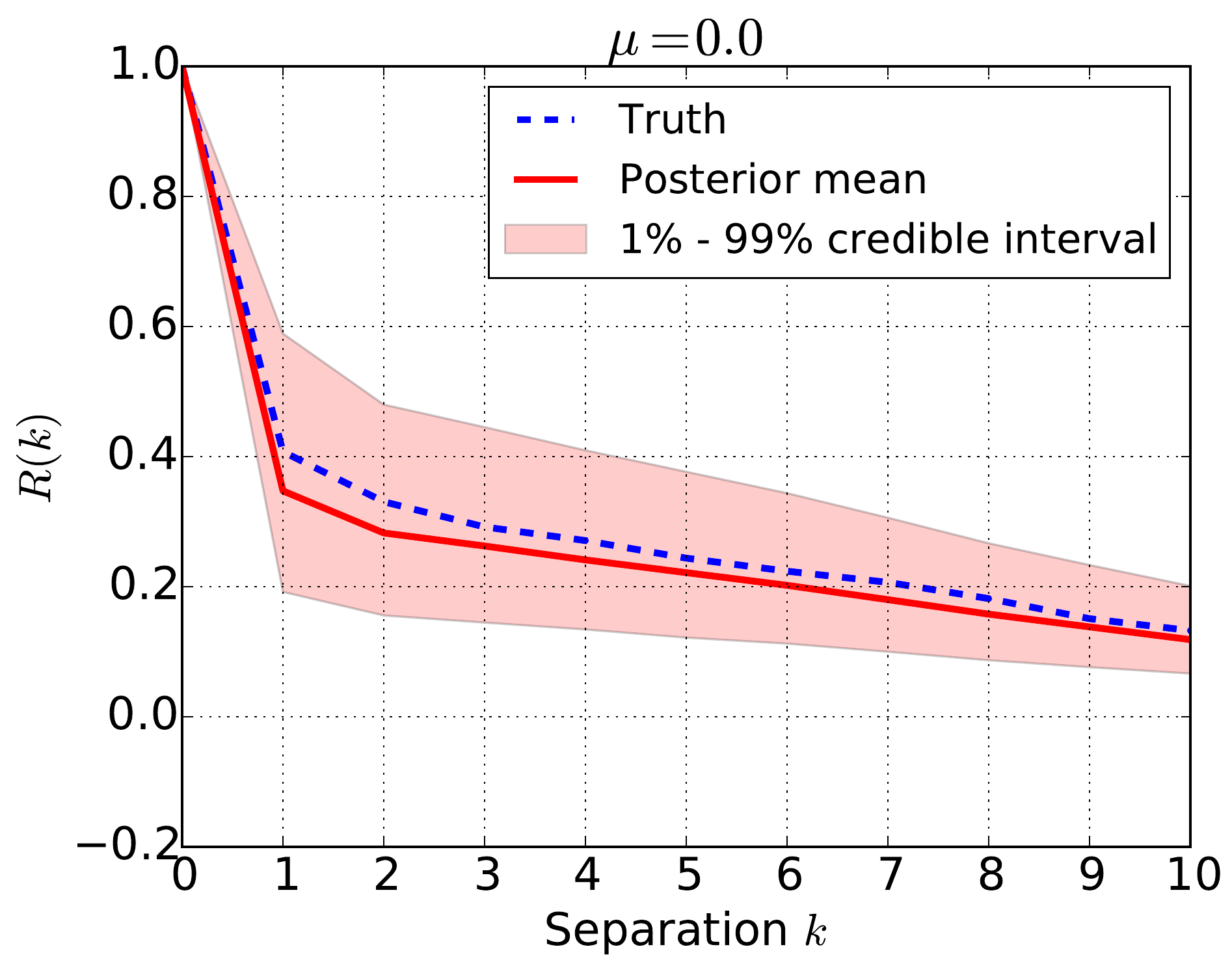}}
\end{minipage}\par\medskip
\centering
\subfloat[$N=50$]{\label{fig:isingcorr_50dp}\includegraphics[width=0.5\textwidth]{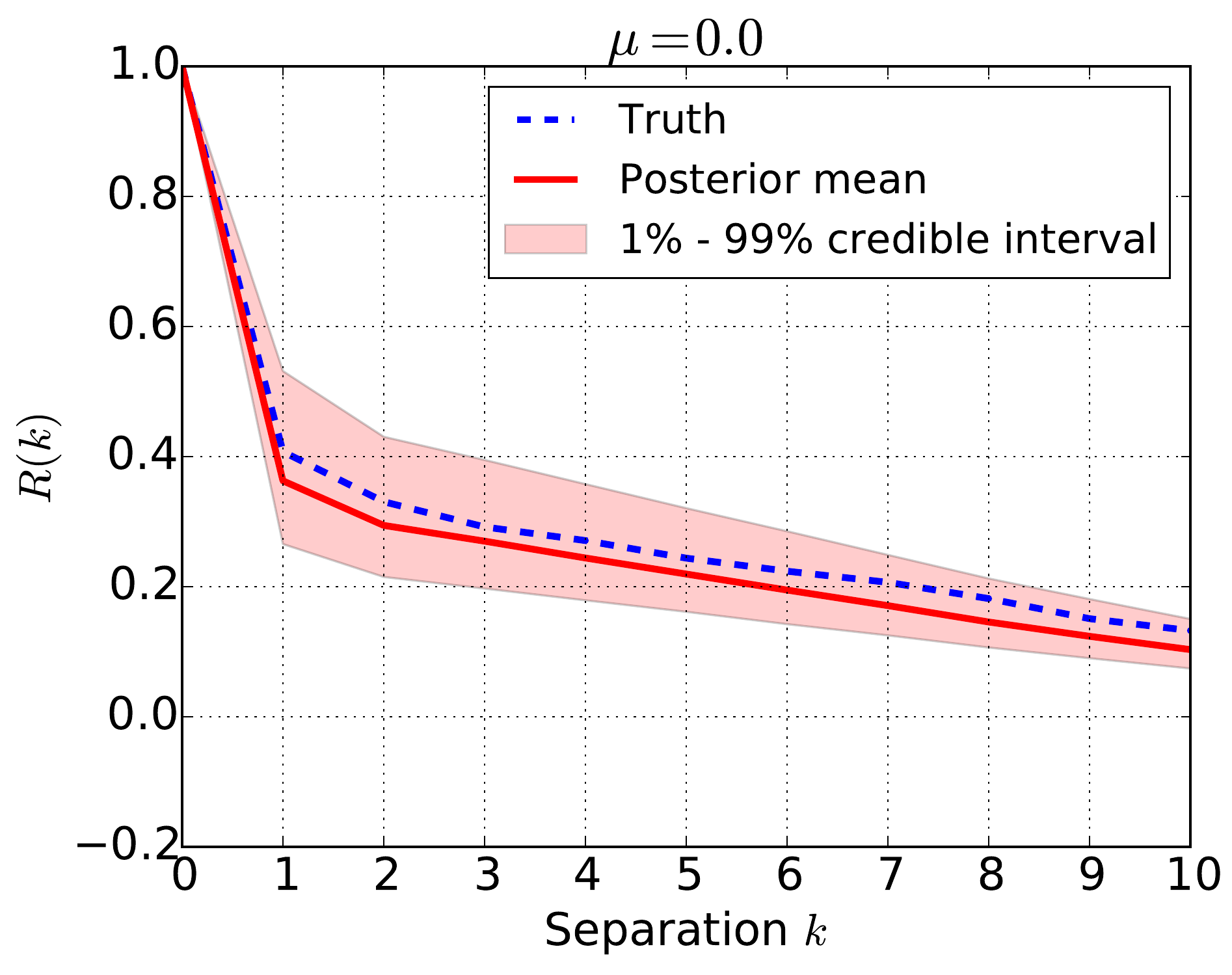}}
\caption{Comparison of the reference correlation (computed with the FG configuration) with posterior mean and credible intervals corresponding to $1\%$ and $99\%$ posterior quantiles.
 $N=20$, $\frac{n\aaa}{n\cg}=2$, $L\cg^{(2)} = 15$, $L\cg^{(3)} = 3$.
 }
\label{fig:isingcorr}
\end{figure}

\reffig{fig:isingmag} depicts the effect of adding more training data $N$ in the predictive posterior estimates
for the magnetization at various $\mu$ values.
One observes that as $N$ increases, not only the posterior mean estimates approach the reference solution, but more importantly,
the posterior credible intervals shrink around it reflecting the fact that the model becomes more confident.
Credible intervals are obtained by sampling the (approximate) posterior distribution $p(\btheta|\bqd)$ (\refeqq{eq:post}) 
and determining the observable for each sample $\btheta^{(i)}$ with the predictive estimator $\hat a(\btheta^{(i)})$ (\refeqq{eqn:propPrediction}).
We use the predictive samples $\hat a(\btheta^{(i)})$ to determine desired quantiles (see \ref{app:credInt} for more details).
The same observations can be made when attempting to predict second-order statistics of the fine-scale i.e. the correlation at various separations $k$ (\reffig{fig:isingcorr}). 
% The predictive quality increases with an increasing amount of training data, as
% shown in  for the magnetization (based on ~\refeqq{eqn:obsMag}) and in \reffig{fig:isingcorr}
% for the correlation (defined in~\refeqq{eqn:obsCorr}).

\begin{figure}
\centering
\includegraphics[width=0.45\textwidth]{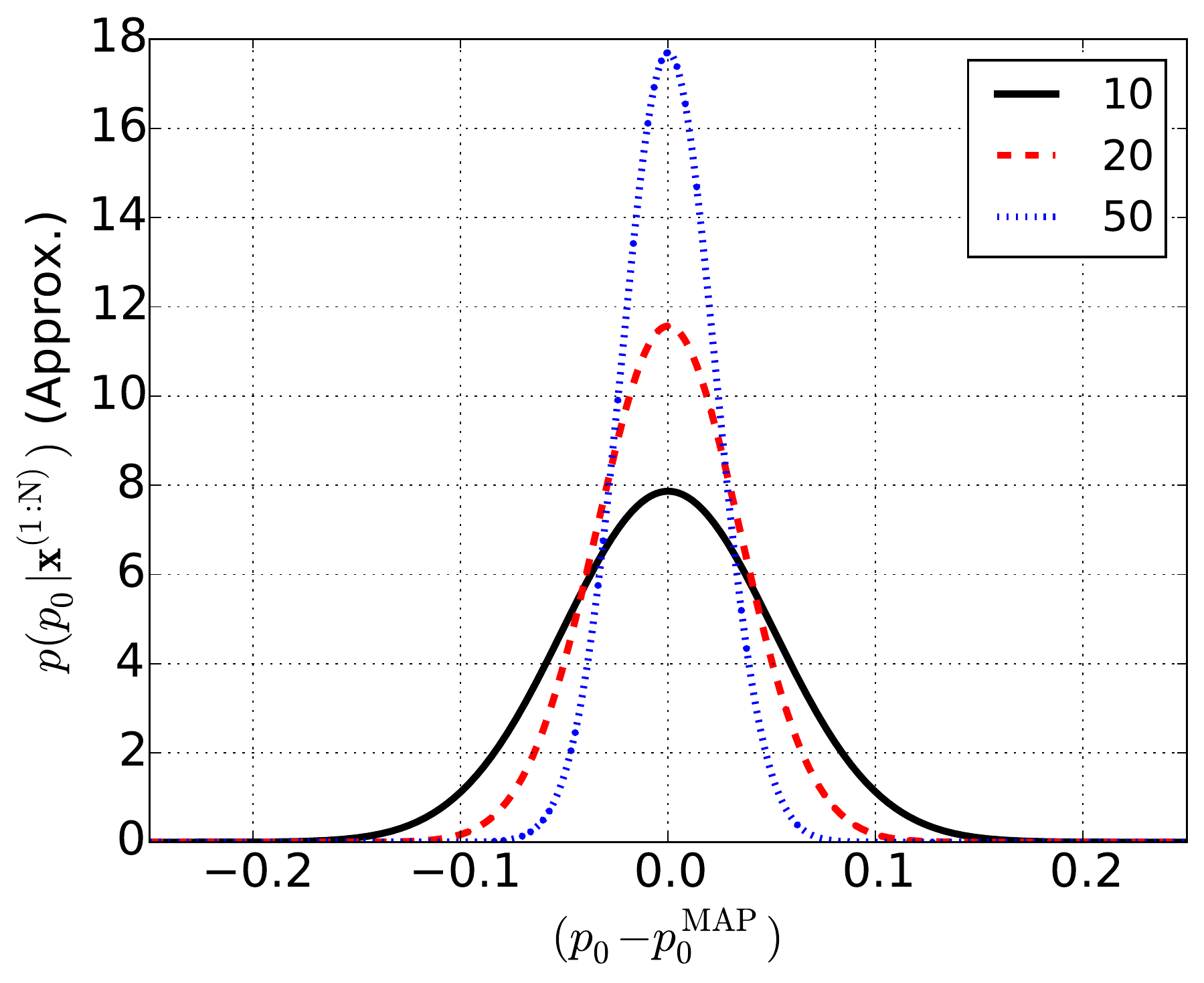}
\caption{Posterior $p (p_0| \bqd)$ at $\mu=0.0$ for $N=10,20,50$.  $\frac{n\aaa}{n\cg}=2$, $L\cg^{(2)} = 15$, $L\cg^{(3)} = 3$.}
\label{fig:ising_posterior_p0}
\end{figure}

The decreasing variance for increasing  $N$  can also be observed in the model parameters e.g. the coarse-to-fine mapping parameter $p_0$ (\refeqq{eq:p0}), the (approximate) posterior of which is shown in \reffig{fig:ising_posterior_p0}.

\begin{figure}[htbp]
\centering
\begin{floatrow}
\ffigbox[\FBwidth]
{
\centering
\subfloat[$\frac{n\aaa}{n\cg}=2$]{\includegraphics[width=0.45\textwidth]{figures/ising_VarMu_Sf_20_levCG_2_pexp_1_Lc_15_LcE_1_LcL_1_LcT_3_plt_mag_errbar.pdf}
\label{fig:ising_mag_lcg_2}
}
\quad
\centering
\subfloat[$\frac{n\aaa}{n\cg}=8$]{\includegraphics[width=0.45\textwidth]{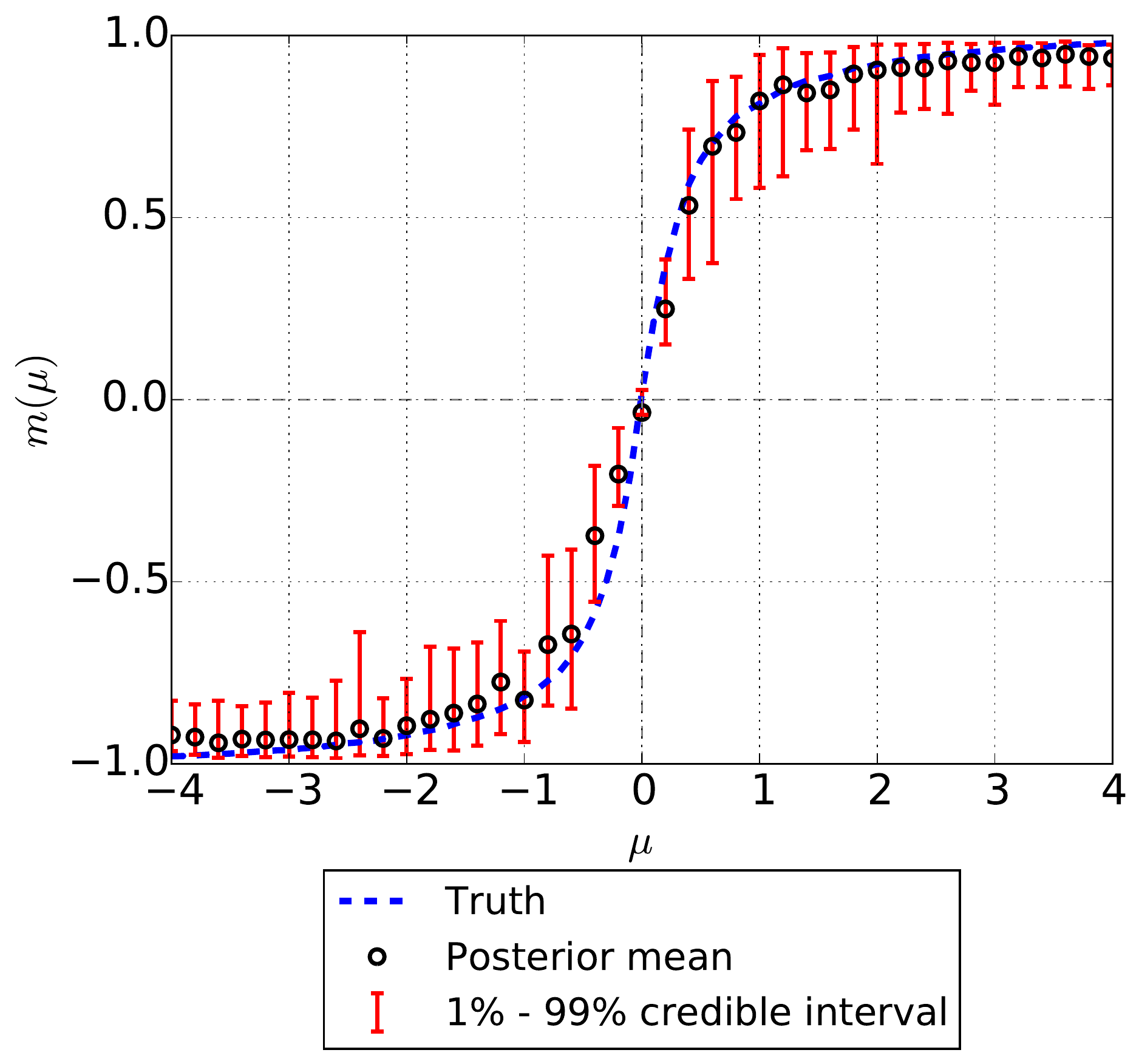}
%original version found under {figures/water/10_dp_lj_cs_50_slow_0001_ard_5_f_1_20/plt_par_freq_sin}
\label{fig:ising_mag_lcg_4}
}
}{
\caption{Magnetization for different level of coarse graining,
i.e. ratio of the amount fine/coarse variables
$\frac{n\aaa}{n\cg}=2$ ($L\cg^{(2)} = 15$, $L\cg^{(3)} = 3$) and
$\frac{n\aaa}{n\cg}=8$ ($L\cg^{(2)} = 3$, $L\cg^{(3)} = 1$). 
Both models were trained with the same data $N=20$.
}
\label{fig:ising_mag_lcg}
}
\end{floatrow}
\end{figure}

\begin{figure}[htbp]
\centering
\begin{floatrow}
\ffigbox[\FBwidth]
{
\centering
\subfloat[$\frac{n\aaa}{n\cg}=2$]{\includegraphics[width=0.45\textwidth]{figures/ising_VarMu_Sf_20_levCG_2_pexp_1_Lc_15_LcE_1_LcL_1_LcT_3_plt_corr_stepMu_3.pdf}
\label{fig:ising_corr_lcg_2}
}
\quad
\centering
\subfloat[$\frac{n\aaa}{n\cg}=8$]{\includegraphics[width=0.45\textwidth]{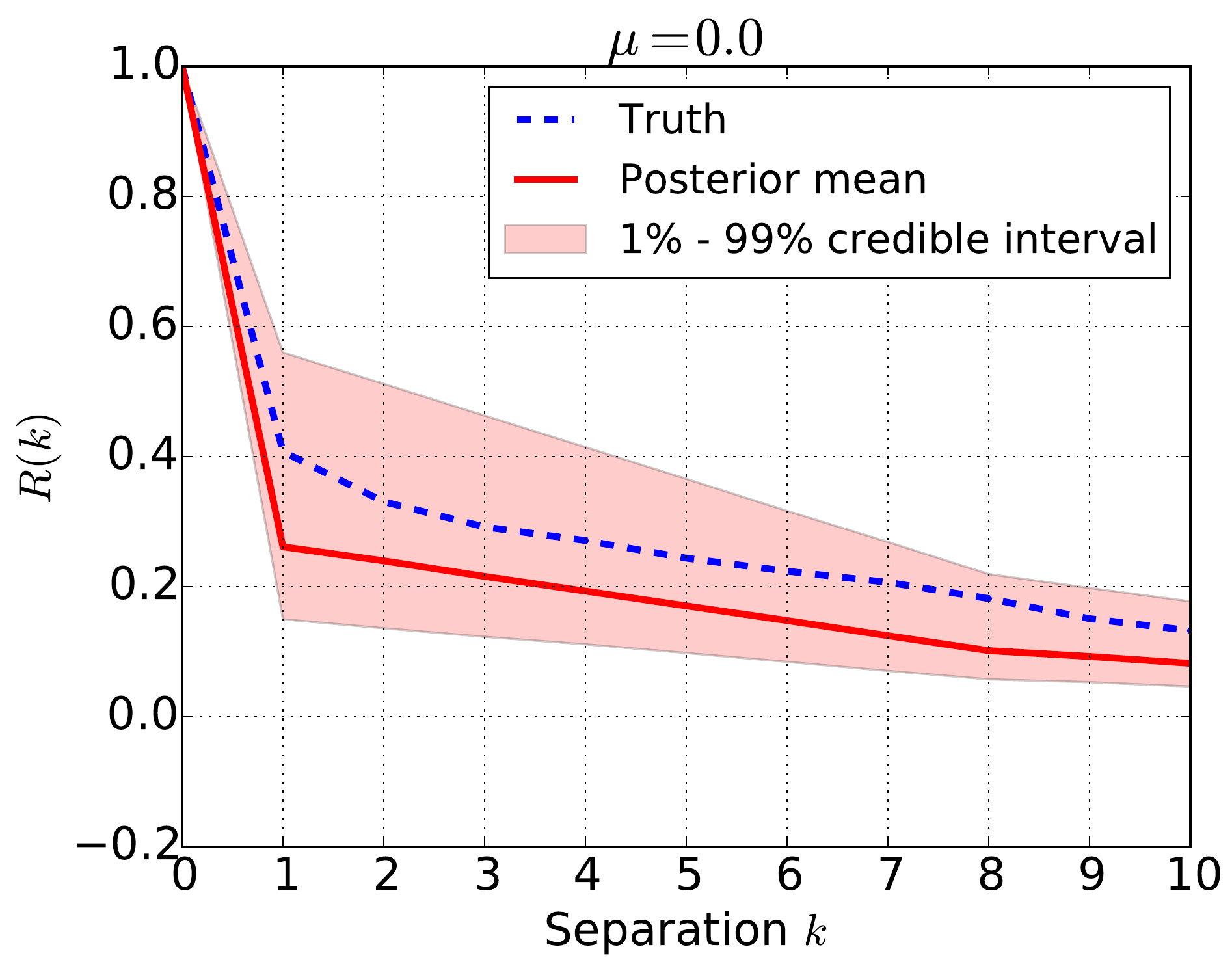}
\label{fig:ising_corr_lcg_8}
}
}{
\caption{Correlation for different level of coarse graining,
i.e. ratio of the amount fine/coarse variables
$\frac{n\aaa}{n\cg}=2$ ($L\cg^{(2)} = 15$, $L\cg^{(3)} = 3$) and
$\frac{n\aaa}{n\cg}=8$ ($L\cg^{(2)} = 3$, $L\cg^{(3)} = 1$).
Both models were trained with the same data $N=20$.
}
\label{fig:ising_corr_lcg}
}
\end{floatrow}
\end{figure}

Finally in Figs. \ref{fig:ising_mag_lcg} and \ref{fig:ising_corr_lcg}, the predictive ability of the model is compared for different levels of coarse-graining. In the formulation adopted, this is quantified by the ratio between the dimension of  fine $\bq$ and coarse $\bQ$ descriptions i.e. $\frac{n\aaa}{n\cg}$. We consider two cases i.e. $\frac{n\aaa}{n\cg}=2,8$. As one would expect, the posterior mean estimates are superior when $\frac{n\aaa}{n\cg}=2$ but also the predictive posterior uncertainty increases as the coarse-graining becomes more pronounced. This is easily understood by the fact that the fewer CG variables used, the higher the information loss becomes. It is important to note though that even when $\frac{n\aaa}{n\cg}=8$, the predictive  posterior's credible intervals always include the reference solution. 
% The predictive quality for a level of CG $\frac{n_f}{n_c}=2$ is better as 
% for i.e. $\frac{n_f}{n_c}=8$ (see \reffig{fig:ising_mag_lcg} for the magnetization and
% \reffig{fig:ising_corr_lcg} for the correlation), while the considerably more efficient description
% still includes the reference magnetization within the predicted credibility intervals.

\subsection{Coarse-Graining SPC/E water}
\label{sec:num_water}
The second example addresses the coarse-graining of a water model which is
described at the atomistic scale by oxygen and hydrogen atoms.  Water has been the focus of several studies in coarse-graining as it plays the role of the  solvent in  various biological and chemical systems and as a result it can take up to $80\%$
of the  total simulation time~\cite{noid2013}. Furthermore there exist several well-documented properties which can serve as a measure of comparison.
% 
% chemical systems. In that system the simulation of water takes up to 80\%
% of the system's total simulation time and thus potentially
% promises the largest increase of computational efficiency~\cite{noid2013}.
% This example is chosen since water poses significant challenges, even at the fine level,
% while well-documented reference properties exist to compare with.
% Furthermore water takes the role as solvent for various biological and
% chemical systems. In that system the simulation of water takes up to 80\%
% of the system's total simulation time and thus potentially
% promises the largest increase of computational efficiency~\cite{noid2013}.
In this study, we employ the  Simple Point Charge/Extended (SPC/E) water model
introduced in~\cite{berendsen1987,kusalik1994} for the  FG (all-atom) description. 
 In the context of the relative entropy method, coarse-graining of the the SPC/E water is addressed in~\cite{bilionis2013,shell2014,kremer2009, Erban20150556}.
%In~\cite{bilionis2013} the procedure for producing fine-scale was adopted  from~\cite{kremer2009}. 
 In particular,  we consider a system of $M=100$ water molecules at a temperature of $T=300\,\textrm{K}$, and a pressure of
$p=1.0\,\text{bar}$. The equilibrium box length is $l_{\mathrm{box}} = 14.56\,\textrm{\AA}$
and a time step of $\Delta t = 2.0\,\textrm{fs}$ is used. Periodic boundary conditions 
are applied in every dimension while ensuring the NVT ensemble by the
Nos\'e-Hoover thermostat~\cite{hoover1985, nose1984}. The $\bq$ vector contains  the coordinates of the 100 oxygen and 200 hydrogen atoms i.e. $\dim(\bq)=900$. The fine-scale potential 
$U\aaa(\bq)$ under the SPC/E model  consists of a Lennard-Jones (LJ) potential for non-bonded interactions and a Coulomb  potential  for long-range interactions.
Parameters for the LJ potential,
\be
U^\mathrm{LJ}\aaa(\bq) =
\frac{1}{2} \sum_{j \neq k}
4\epsilon \left( \left(\frac{\sigma}{R_{ij}(\bq)}\right)^{12}
- \left(\frac{\sigma}{R_{ij}(\bq)}\right)^{6} \right),
\ee
are $\sigma=3.166\,\textrm{\AA}$ and $\epsilon = 0.650\,\frac{\textrm{kJ}}{\textrm{mol}}$,
with the distance between particle $i$ and $j$ denoted as $R_{ij}$.

The electric load of Hydrogen (H) and Oxygen (O) atoms are given by $q_\mathrm{O} = -0.8476\,e$,
$q_\mathrm{H} = +0.4238\,e$ where $e$ represents the elementary charge.  The SPC/E model assumes the bonded interaction
to be rigid with a bonding angle defined between the two H-atoms and the central O-atom as
$\omega_\mathrm{HOH}=109.47^\circ$. The  bond-length used in this study is
$l_\mathrm{OH}=1.0\,\textrm{\AA}$.
The equilibration for the  NVT 
ensemble  was performed as in~\cite{bilionis2013,kremer2009}. For both fine- and coarse-scale simulations the molecular dynamics software package LAMMPS~\cite{plimpton1995} was used. Further details are contained in   \ref{app:spceProd}.

The values $A=9$, $\alpha=0.05$, and $\rho= 0.60$ were used for the Robbins-Monro
updates (\refeqq{eqn:RMupdate}) based on suggestions given in~\cite{bilionis2013}.
We used $m=160$
samples for the MCMC estimates of the gradients in Eqs.~\eqref{eq:gradientmcest} and~\eqref{eq:gradmc}.

% {\color{red}
% The values $\alpha=0.05$, and $\rho=0.60$ for the Robbins-Monro updates,
% given in~\refeqq{eqn:RMupdate}, seem promising for the
% following numerical studies.
% $A$ is selected as $3\%$ of the $300$ maximal expected iterations until convergence,
% thus $A=0.03 \cdot 300=9$, based on suggestions given in~\cite{bilionis2013}.
% In the following, we use $m=160$ samples for the MCMC estimates of the gradients
% in Eqs.~\eqref{eq:gradientmcest} and~\eqref{eq:gradmc}.
% }

\subsubsection{Observables}
The first macroscopic observable of interest is the Radial Distribution Function (RDF) $g(r)$  which represents a characteristic and well-studied property in water models.
Several  computational and experimental results related to the RDF  are described in~\cite{clark2010}.
As a pair correlation function,  $g(r)$
depends on the statistics of the distances $r_{jk}$ between each pair of molecules  $j$, $k$. To compute these distances, we employ the coordinates of the center of mass of each water molecule $\hat{\bq}_j$:
\be
\hat{\bq}_j=\frac{ \bq_{\mathrm{O},j} m_\mathrm{O}+\bq_{\mathrm{H},j_1}m_\mathrm{H} +\bq_{\mathrm{H},j_2}m_\mathrm{H}}{m_\mathrm{O}+2m_\mathrm{H}},
\label{eq:centermass}
\ee
where $\bq_{\mathrm{O},j}$  are the coordinates of the oxygen atom of molecule $j$, $\bq_{\mathrm{H},j_1}, \bq_{\mathrm{H},j_2}$ are the coordinates of the two hydrogen atoms of the same molecule, and $m_\mathrm{O}, m_\mathrm{H}$ are 
the masses of oxygen and hydrogen atoms, respectively (see  \ref{app:spceProd}).
If $r_{jk}=|\hat{\bq}_j-\hat{\bq}_k|$, then the corresponding observable of interest is~\cite{molinero2014}:
\be
 a^{\mathrm{RDF}}(\bq) = \frac{V}{M^2} \sum_j^M \sum_{j \neq k}^M \delta(r-r_{jk}),
 \label{eq:obsrdf}
\ee
where $V$ denotes the volume of the simulation box ($14.56^3\,\textrm{\AA}^3$) and  $M=100$ the  number of molecules in the system. Additional details can be found in \ref{app:spceRDF}.

% 
% and not directly on the absolute coordinates $\bq$. 
% For predictive purposes we use the framework defined in 
% Algorithm \ref{alg:calcObservable}
% with an observable $a^g(\bq)$ (see Ref.~\cite{molinero2014})
% \be
%  a^g(\bq) = \frac{V}{N^2} \sum_i \sum_{i \neq j} \delta(r-r_{ij}),
% \ee
% with $r_{ij}= || \bq_j - \bq_i ||$, the euclidean distance. $V$ denotes the volume of the
% considered simulation box and $N$ the amount of molecules in the system.
% A more detailed definition of the RDF is provided in the \ref{app:spceRDF}.

The second property of interest involves the tetrahedral structure of water.
Neighboring water molecules temporarily build such tetrahedral clusters due to the
hydrogen bonds. Several measures of tetrahedrality have been proposed which relate to the deviation  from the perfect
tetrahedral structure $\omega_{0} = 109.471^\circ$
\cite{shell2014, wang2009}.
In this work, we employ  the angular distribution function which considers 
the eight closest neighbors $n_c=8$ for a given molecule $j$.
It is defined as follows:
\be
a^{\mathrm{tetra}}(\bq; \omega)  = \frac{1}{M n_{\omega}} %\left\langle
\sum^M_{j=1} \sum^{n_c}_{k=1} \sum^{n_c-1}_{l \neq j}
\delta (\omega - \omega_{jkl}),
% \right \rangle_{p(\bq)},
\label{eq:obstetra}
\ee
with  $\omega_{jkl}$ the angle between molecules $j,k,l$, with the central molecule
$j$, (as computed using the centers of mass $\hat{\bq}$  in \refeqq{eq:centermass}) 
and $n_\omega= \left( \begin{array}{l} 
 n_c \\3 \end{array} \right) =56$. 
The product $(M n_\omega )$ normalizes $a^{\mathrm{tetra}}$ with respect to the
considered angular triplets.

We  note that since the observables of interest depend
only on the centers of mass $\hat{\bq}=\hat{\bq}(\bq)$,
it suffices to use a  coarse-to-fine map that relates the coarse variables
$\bQ$ directly with $\hat{\bq}$ (\refeqq{eqn:propPrediction}).

\subsubsection{Coarse-variables $\bQ$ and coarse-to-fine map}
\begin{figure}[h]
\centering
\includegraphics[width=0.3\textwidth]{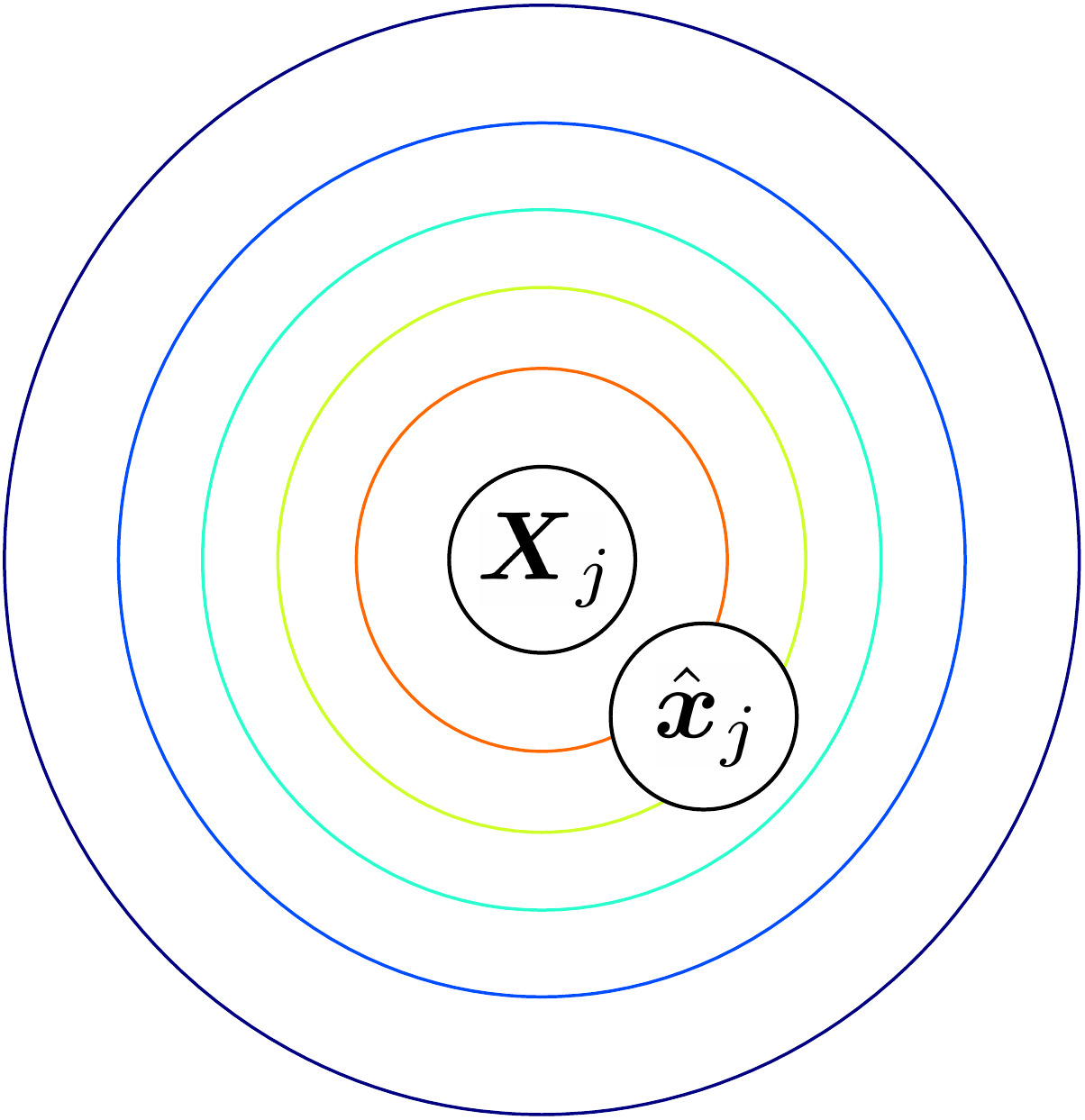}
\caption{Probabilistic mapping $p\cf(\hat \bq_j |\bQ_j,\bgam\cf)$, with mean $\bQ_j$ and predicted fine-scale variable $\hat \bq_i$.
The countours depict the isotropic Gaussian distribution of \refeqq{eqn:waterPcf} with mean $\bQ_j$ and variance $\sigma^2$.}
\label{fig:pcf_water}
\end{figure}

Since the  observables of interest depend on the centers of mass $\hat{\bq}$ (\refeqq{eq:centermass}), the coarse-to-fine probabilistic map assumes the form $p\cg(\hat{\bq} | \bQ)$. As frequently done in CG studies of water, each molecule $j$ is represented by a CG variable $\bQ_j \in \RR^3$. We then prescribe a $p\cf$ of the following form:
\be 
\pcf(\hat{\bq}|\bQ, \bgam\cf) = \prod_{j=1}^M \mathcal{N}(\hat{\bq}_j|  \bQ_j, ~\sigma^2 \bs{I}), 
 \label{eqn:waterPcf}
\ee
where $\bs{I}$ is the $3\times 3$ identity matrix. This suggests that each $\bQ_j, j=1,
\ldots, M$ determines the center of mass $\hat{\bq}_j$ up to an isotropic Gaussian with mean $\bQ_j$ and variance $\sigma^2$ (see \reffig{fig:pcf_water}).
The latter quantifies the uncertainty in the prediction of the fine-scale (up to centers of mass) from the CG description. Large values of $\sigma^2$ imply that $\bQ$ provides an imprecise reconstruction of $\hat{\bq}$ and vice versa. 
Hence there is only one parameter in the coarse-to-fine map i.e. $\sigma^2 \ge 0$. In order to ensure non-negativity during updates we operate instead on $\bgam\cf = -\log  \sigma^2$ which leads to the following derivatives
needed in Eqs. \eqref{eq:gradmc} and \eqref{eq:hessian}: 
%\be
\begin{align}%{ll}
 \frac{ \pa \log p\cf}{\pa \theta\cf} & =\frac{3M}{2} -\frac{1}{2\sigma^2} \sum_{j=1}^M | \hat{\bq}_j-\bQ_j|^2, \nonumber \\
 \frac{ \pa^2 \log p\cf}{\pa \theta\cf^2} & =-\frac{1}{2\sigma^2} \sum_{j=1}^M | \hat{\bq}_j-\bQ_j|^2.
\end{align}
%\ee
Naturally, more complex descriptions involving an anisotropic covariance or a mixture of Gaussians could be used. 

% 
% We assume the center of mass of the water molecule $i$ with the coordinates $\bq_i$
% is distributed as a Gaussian with the position of the coarse-variable of the
% corresponding coarse particle $\bQ_j$ as mean. Additionally we assume spacial
% independence and no correlation within the different directions. The mapping
% for all $M$ particles in the system follows to,
% \begin{align}
%  \pcf(\bq|\bQ, \bgam\cf) &= \prod_{m=1}^M\prod_{j=1}^3 \calN(x^{m}_j|X^{m}_j,\sigma^2) \\
%  & \propto \exp\{ -\frac{\sum_m^M || \bq_m - \bQ_m ||^2}{2 \sigma^2} \},
%  \label{eqn:waterPcf}
% \end{align}
% given the coordinate of particle $m$ in dimension $j$.
% With $\bgam\cf = \sigma^2$ we optimize with respect to the variance of the
% given probabilistic mapping.
% The distribution
% $\pcf(\bq|\bQ, \bgam\cf)$ accounts for an eventual lack of expressiveness
% of the coarse potential $U\cg(\bQ,\bgam\cg)$. The mapping
% $\pcf(\bq|\bQ,\bgam\cf)$, denotes the probability of finding the 
% center of mass of the water molecule given the coarse-variable
% $\bQ$. 

\subsubsection{Coarse model}

The coarse potential  $U\cg(\bQ; \bgam\cg)$ employed consists of two- and three-body interactions. It assumes the form:
\be
U\cg(\bQ; \bgam\cg)= \underbrace{U^{\mathrm{SW}}(\bQ)}_{\text{fixed}}+\tilde{U}(\bQ; \bgam\cg),
\label{eq:waterUc}
\ee
where $U^{\mathrm{SW}}(\bQ)$ is a fixed term described below and $\tilde{U}(\bQ; \bgam\cg)$ represents the ``correction" that is learned from the data using the framework advocated.
In particular, the fixed term $U^{\mathrm{SW}}(\bQ)$ is given by (a variation of)  the Stillinger-Weber (SW) potential proposed in~\cite{stillinger1985} and discussed in \ref{app:sw}.
The remaining part $\tilde{U}(\bQ; \bgam\cg)$ consists only of two-body interaction terms i.e.
\be
 \tilde{U}(\bQ; \bgam\cg)= \frac{1}{2}\sum_{j\ne k}  u^{(2)}(R_{jk}; \bgam\cg),
 \label{eq:waterUc1}
 \ee
 where $R_{jk}=|\bQ_j-\bQ_k|$ and the pairwise potential $ u^{(2)}(R; \bgam\cg)$ is parametrized as follows:
 \be
 u^{(2)}(R; \bgam\cg)=u^{\mathrm{LJ}}(R; \bgam\cg^{\mathrm{LJ}} )+\sum_{k=1}^K \theta_{\mathrm{c},k}^{\mathrm{cor}} \phi_k(R), \quad R>0.
 \label{eq:water2nd}
 \ee
 In the equation above, 
 $u^{\mathrm{LJ}}(R; \bgam\cg^{\mathrm{LJ}} )$ is a Lennard-Jones potential and the feature functions $\bs{\phi}=\{\phi_k(R)\}_{k=1}^K$
 are a combination of sines and cosines truncated in the interval $I\cg=[R_\mathrm{{min}}=2.0\,\mathrm{\AA},~R_{\mathrm{max}}=6.0\,\mathrm{\AA}]$. The bounds  $R_\mathrm{{min}}, R_{\mathrm{max}}$ define an   effective window where the LJ potential is corrected to capture the associated CG interactions. In particular:
 \be
 \phi_k(R) = \left\{  \begin{array}{ll}
                       1_{I_c}(R)~\sin 2\pi \nu_k R, & ~k=\text{odd}, \\
                       1_{I_c}(R)~\cos 2\pi \nu_k R, & ~k=\text{even}, \\
                      \end{array} \right.
 \label{eq:sincos}
 \ee
 where $1_{I\cg}(R)$ is the indicator  function of the interval $I\cg$.
 The wave-numbers $\nu_k$ offer a Fourier-like decomposition of the second-order potential and were defined as follows:
 \be
%  \textrm{\color{red} WHAT ARE THE $\nu_k=??$}
\nu_{2k'}=\nu_{2k'+1}=1+\frac{19}{K/2} k', \quad k'=0,2,\ldots,K/2-1,
\ee
i.e. at a uniform grid in $[1,20]$.
By increasing the total number $K$ of these terms, one can potentially learn  finer fluctuations of this potential. Naturally one would want to use as many feature functions as possible in order to ensure greater flexibility of the model, which gives rise to the need for sparsity-enforcing priors for $\theta_{\mathrm{c},k}^{\mathrm{cor}}$ as discussed previously.
In this study, $K=100$ was used. 
 
 The superimposed  LJ potential ensures that $\lim_{R \rightarrow 0}  u^{(2)}(R; \bgam\cg) = \infty$ and is of the form:
 \be
 u^{\mathrm{LJ}}(R; \bgam\cg^{\mathrm{LJ}} )= 4\epsilon \left( \left(\frac{\sigma_{\mathrm{LJ}}}{R}\right)^{12}
  - \left( \frac{\sigma_{\mathrm{LJ}}}{R} \right)^6 \right),
  \label{eqn:UcLJ}
 \ee
where $\bgam\cg^{\mathrm{LJ}}=(\sigma_{\mathrm{LJ}},\epsilon)$.  
The total number of parameters associated with the two-body term was $K+2=102$ and consists of  $\bgam\cg=(\bgam\cg^{\mathrm{LJ}}, \bgam\cg^{\mathrm{cor}})$.
The ARD prior is employed only for $\bgam\cg^{\mathrm{cor}}$ and an (improper) uniform prior is employed for the rest $\bgam\cg^{\mathrm{LJ}}$. We note that due to the LJ part,
the corresponding distribution $p\cg$ is not in the exponential
family anymore (Section \ref{sec:expon}) and the possibility of multiple local maxima cannot be excluded.

\subsubsection{Results}

\begin{figure}[htbp]
\centering
\begin{floatrow}
\ffigbox[\FBwidth]
{
\centering
\subfloat[Evolution of $u^{(2)}(R; \bgam\cg)$ in \ref{eq:water2nd} at various iterations of the Algorithm \ref{alg:bayesianCG}.
Darker lines correspond to more proceeded iteration steps in
the optimization scheme.
The solid line shows the converged solution $u^{(2)}(R; \bgam_\mathrm{c,MAP})$.]
{\includegraphics[width=0.45\textwidth]{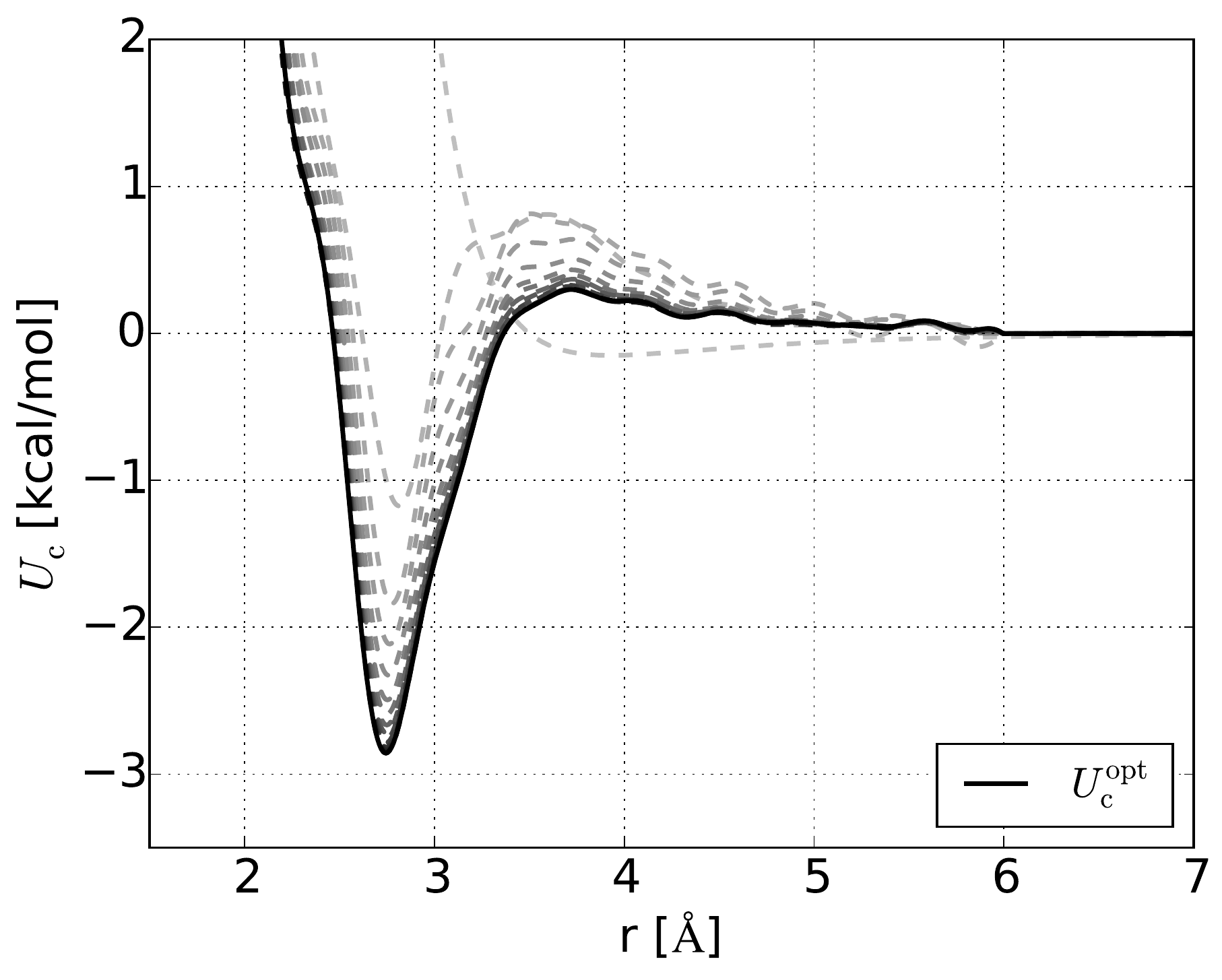}
\label{fig:20_three_body_Uc_evolution}
}
\quad
\centering
\subfloat[Comparison of $u^{(2)}(R; \bgam\cg)$ identified with the proposed method (red) with the  two-body potential computed using the 
relative entropy  method  in~\cite{molinero2014} (dashed blue) and the
LJ part of the fine-scale SPC/E model  $U\aaa^\text{LJ\ SPC/E}$.]
{\includegraphics[width=0.45\textwidth]{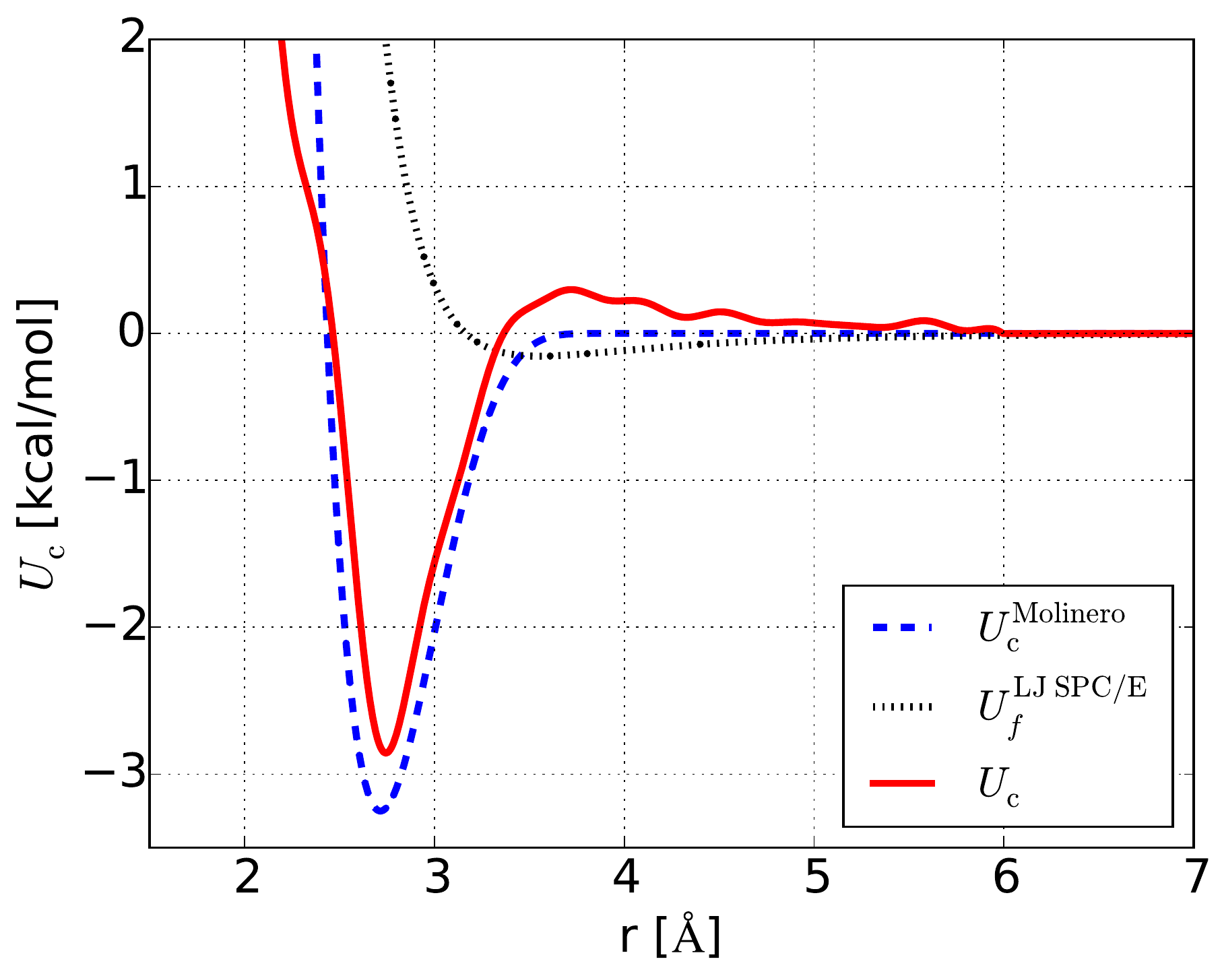}
%original version found under {figures/water/10_dp_lj_cs_50_slow_0001_ard_5_f_1_20/plt_par_freq_sin}
\label{fig:20_three_body_Uc_compare}
}
}{
\caption{Coarse-graining SPC/E water using $N =20$ training data. Computed two-body, coarse-scale potential  $u^{(2)}(R; \bgam\cg)$ and comparisons.}
\label{fig:20_three_body_Uc}
}
\end{floatrow}
\end{figure}

We first run the proposed algorithm for $N=20$ fine-scale (all-atom) realizations. Figure~\ref{fig:20_three_body_Uc_evolution} depicts the evolution of the inferred coarse-scale potential $u^{(2)}(R; \bgam\cg)$ (\refeqq{eq:water2nd})
at various iterations of the EM-scheme. We initialize with $\bgam\cg^{\mathrm{cor}}=\bs{0}$ and
$\bgam\cg^{\mathrm{LJ}}=(\epsilon=0.15\,\frac{\mathrm{kcal}}{\mathrm{mol}}, \sigma_{\mathrm{LJ}}=3.5\,\mathrm{\AA})$. After 194 iterations, the converged result $u^{(2)}(R; \bgam_{\mathrm{c,MAP}})$ is depicted with a solid black line. In Fig.~\ref{fig:20_three_body_Uc_compare}, we compare this converged result  (red)
 with the two-body potential computed in~\cite{molinero2014} (dashed blue)  using the relative entropy method and the LJ part (black) of the fine-scale SPC/E model.
 The former two exhibit similarities but also differences which stem from the different structure of these two models. These differences persist even if more training data $N$ are used.

\begin{figure}[htbp]
\centering
\begin{floatrow}
\ffigbox[\FBwidth]
{
\centering
% \subfloat[Without ARD-prior]{
% %\resizebox{0.8\textwidth}{!}{%
% \setlength\figureheight{.4\textwidth} 
% \setlength\figurewidth{.45\textwidth}
% \input{figures/water/10_dp_cs_50_no_ard/plt_par_freq_sin_cos.tex}
% 
% %}
\subfloat[Without ARD prior]{\includegraphics[width=0.45\textwidth]{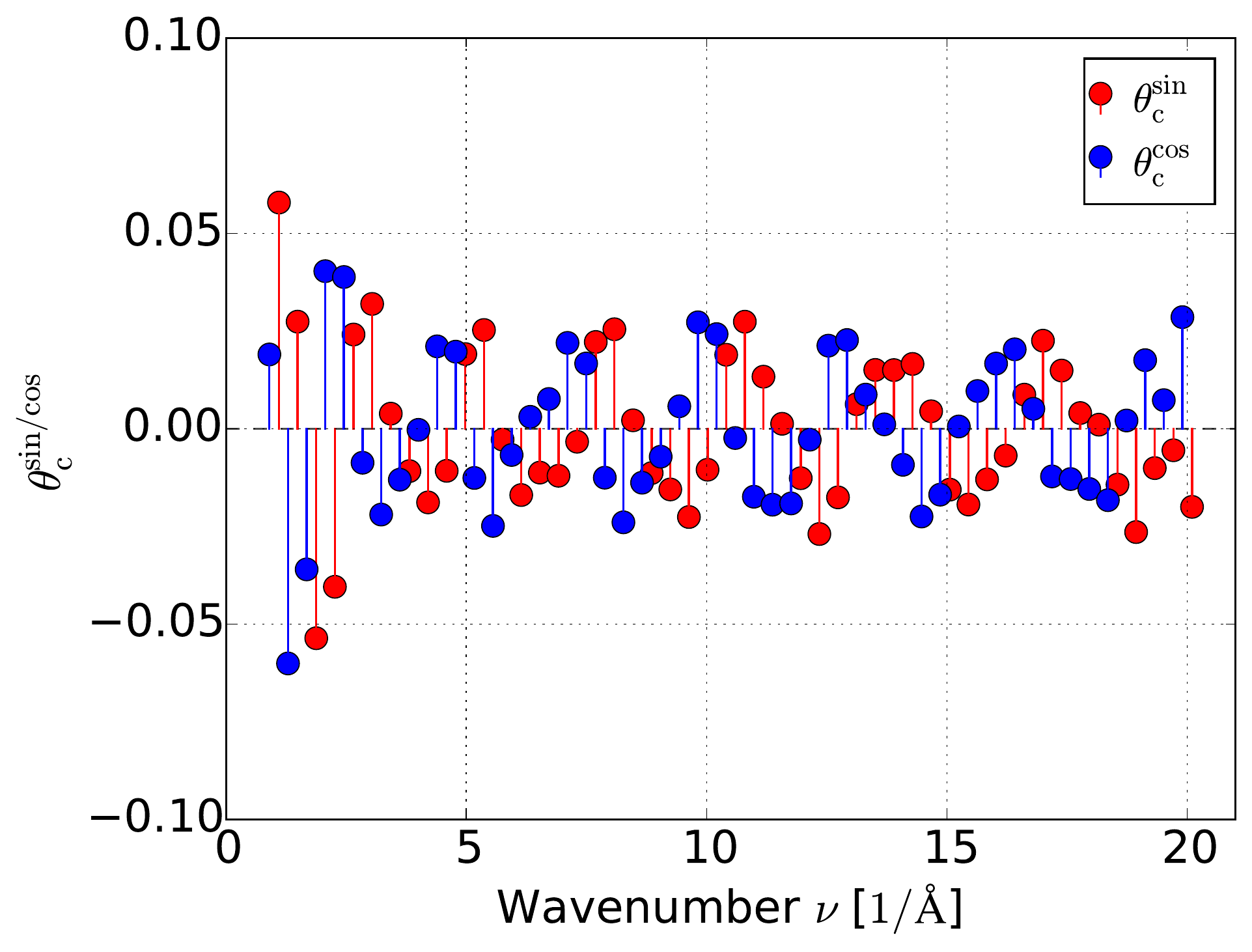}
\label{fig:10_dp_cs_50_no_ard}
}
\quad
\centering
% \subfloat[With ARD-prior]{
% %\resizebox{0.48\textwidth}{!}{%
% %  \pgfplotsset{ticklabel style={font=\small\sffamily},
% %                 legend style={font=\small}}
% \setlength\figureheight{.4\textwidth} 
% \setlength\figurewidth{.45\textwidth}
% \input{figures/water/10_dp_cs_50_ard/plt_par_freq_sin_cos.tex}
% %}
\subfloat[With ARD prior]{\includegraphics[width=0.45\textwidth]{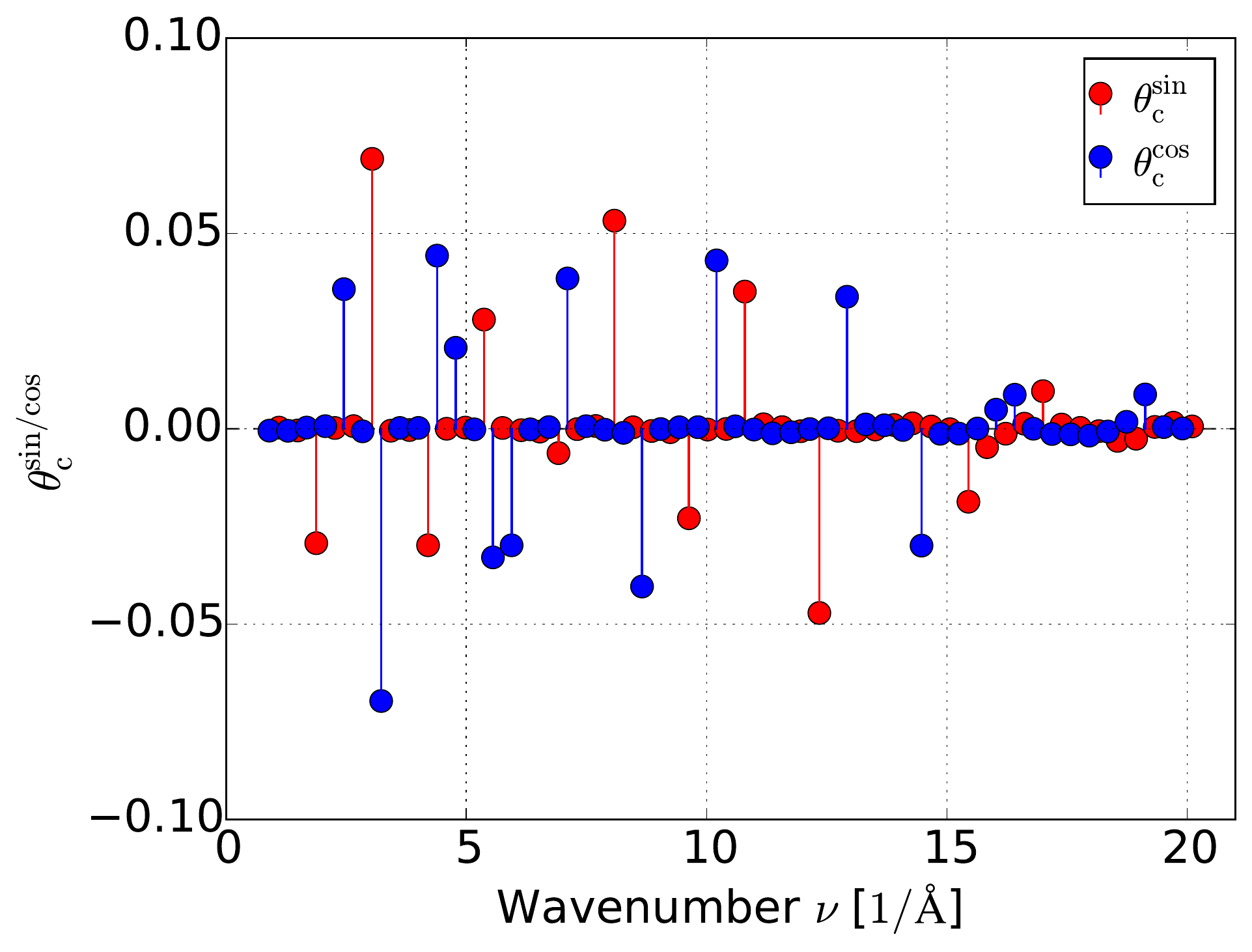}
%original version found under {figures/water/10_dp_lj_cs_50_slow_0001_ard_5_f_1_20/plt_par_freq_sin}
\label{fig:10_dp_cs_50_ard}
}
}{
\caption{$\bgam_\mathrm{c,MAP}^{\mathrm{cor}}$ without and with the ARD prior with respect to the wavenumber $\nu_k$ (\refeqq{eq:sincos}).
Superscripts sin (red) and cos (blue) indicate whether the corresponding  $\theta_{\mathrm{c},k}^{\mathrm{cor}}$ (\refeqq{eq:water2nd}) is associated with a sine or cosine feature function respectively. }
\label{fig:10_dp_cs_50_par}
}
\end{floatrow}
\end{figure}

\begin{figure}[htbp]
\centering
\begin{floatrow}
\ffigbox[\FBwidth]
{
\centering
\subfloat[Joint posterior $p(\sigma_\mathrm{LJ},\epsilon|\bqd)$.]{\includegraphics[width=0.6\textwidth]{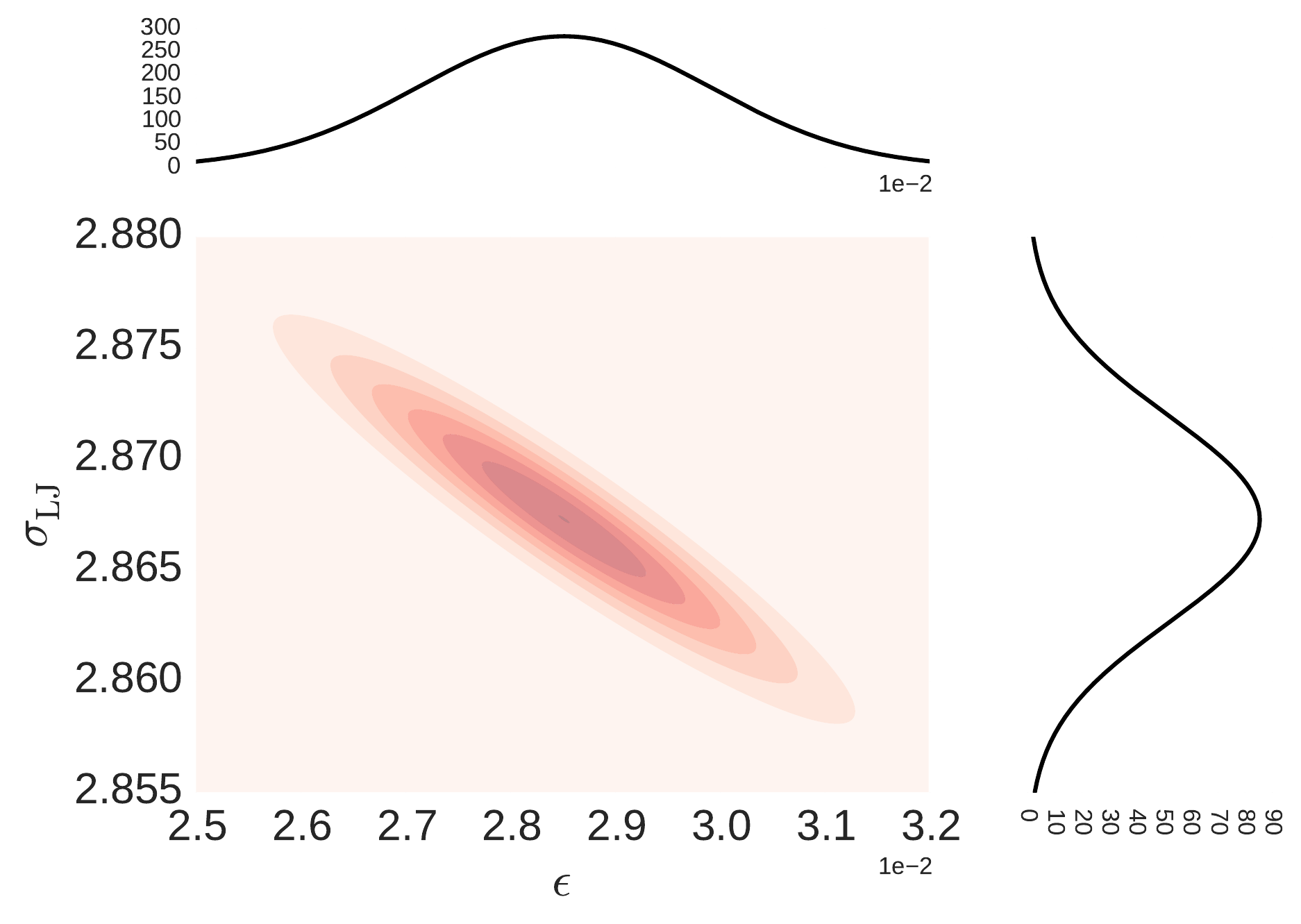}
\label{fig:ljjoint}
}
\quad
%\\
\centering
\subfloat[Posterior $p(\sigma^2|\bqd)$.]{\includegraphics[width=0.35\textwidth]{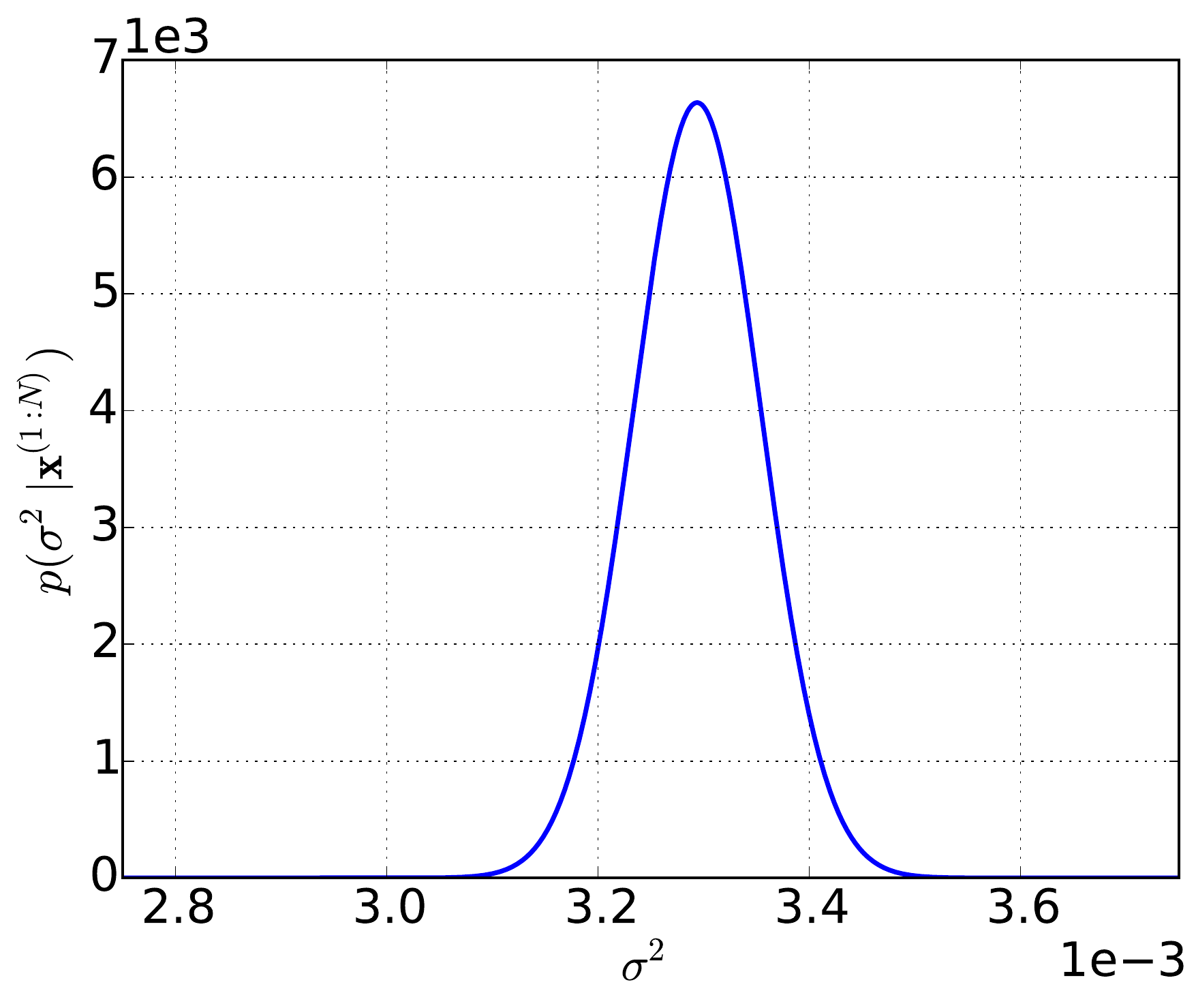}
%original version found under {figures/water/10_dp_lj_cs_50_slow_0001_ard_5_f_1_20/plt_par_freq_sin}
\label{fig:sigma2}
}
}{
\caption{Posterior of $\bgam\cg^{\mathrm{LJ}}=(\sigma_{\mathrm{LJ}},\epsilon)$ in \refeqq{eqn:UcLJ} and $\sigma^2$ in \refeqq{eqn:waterPcf} for $N=20$. }
\label{fig:postLJ}
}
\end{floatrow}
\end{figure}

Figure~\ref{fig:10_dp_cs_50_par} depicts the effect of the ARD prior on $\bgam\cg^{\mathrm{cor}}$.
%which in this case if of dimension $100$ as explained earlier.
One observes in
\reffig{fig:10_dp_cs_50_no_ard} that if no such prior is used (instead a uniform was employed) almost all $\bgam\cg^{\mathrm{cor}}$ are non-zero and as a result almost all the  corresponding feature functions  $\phi_k(R)$ in \refeqq{eq:water2nd} are active and the model is unable to distinguish their relative importance (unless $N$ becomes very large).
In contrast, the inclusion of the ARD prior in \reffig{fig:10_dp_cs_50_ard} leads to a sparse solution in which most $\phi_k(R)$ are deactivated (roughly 80 out of 100 in this case). It can be clearly seen as well that feature functions (sines/cosines) with high wave-numbers (small wave-lengths) are largely unnecessary for the description of the coarse potential. Although not demonstrated in this run, we envision that this modeling feature will eventually allow us to identify not only the most important terms in each potential term but also the most suitable order of interactions in the coarse potential.
Figure~\ref{fig:postLJ} depicts the (approximate) posterior obtained for $\bgam\cg^{\mathrm{LJ}}=(\sigma_{\mathrm{LJ}},\epsilon)$ (\refeqq{eqn:UcLJ}) and $\sigma^2$ (\refeqq{eqn:waterPcf}) for $N=20$.

% Comparing the resulting parametrization $\bgam\cg^{\text{sin/cos}}$
% obtained with active and deactivated ARD-prior in Figure \ref{fig:10_dp_cs_50_par},
% it clearly shows the a desirably sparse solution with 20 out of 100 active features 
% $\phi_i(\bQ)$ can be obtained by using an ARD-Prior. The obtained sparse
% solution is able to capture the important features of the flexible potential 
% and to turn off others not necessary describing the data.
% 
% Figure \ref{fig:10_dp_cs_50_ard} clearly shows that a sparse
% representation of the potential with recognizing the important
% features is found by using an ARD prior. Without sparsity enforcing prior
% all features are necessary for describing the underlying data (see Figure \ref{fig:10_dp_cs_50_ard})
% while using the ARD prior leads to a solution with $20$ out of 100 active
% features which are sufficient for describing the potential
% $U\cg(\bQ,\bgam\cg)$ given the data.

% \begin{figure}[htbp]
% \begin{minipage}{.25\linewidth}
% \vbox to -30mm{}
% \centering
% \subfloat[$\sigma_{\mathrm{LJ}}$]{\label{fig:sigma_20} \includegraphics[width=0.9\textwidth]{figures/sigma_LJ_20_posterior.pdf}}
% \end{minipage}%
% \begin{minipage}{.25\linewidth}
% \centering
% \subfloat[joint $(\sigma_{\mathrm{LJ}},\epsilon)$]{\label{fig:ljjoint} \includegraphics[width=0.9\textwidth]{figures/LJ_20.pdf}}\\
% 
% \subfloat[$\epsilon$]{\label{fig:eps_20}\includegraphics[width=0.9\textwidth]{figures/epsilon_LJ_20_posterior.pdf}}\\
% \end{minipage}
% 
% \begin{minipage}{.45\linewidth}
% \centering
% \vbox to -10mm{}
% \subfloat[$\sigma^2$]{\label{fig:sigma2}\includegraphics[width=0.99\textwidth,height=4cm]{figures/sigma2_20_posterior.pdf}}
% \end{minipage}%
% 
% \caption{Posterior of $\bgam\cg^{\mathrm{LJ}}=(\sigma_{\mathrm{LJ}},\epsilon)$ \eqref{eqn:UcLJ} and $\sigma^2$ \eqref{eqn:waterPcf} for $N=20$. }
% \label{fig:postLJ}
% \end{figure}

\begin{figure}[htbp]
\centering
\begin{floatrow}
\ffigbox[\FBwidth]
{
\centering
\subfloat[Realizations of $u^{(2)}(R; \bgam\cg)$ for random samples of $\bgam\cg$ drawn from the approximate posterior.]{\includegraphics[width=0.45\textwidth]{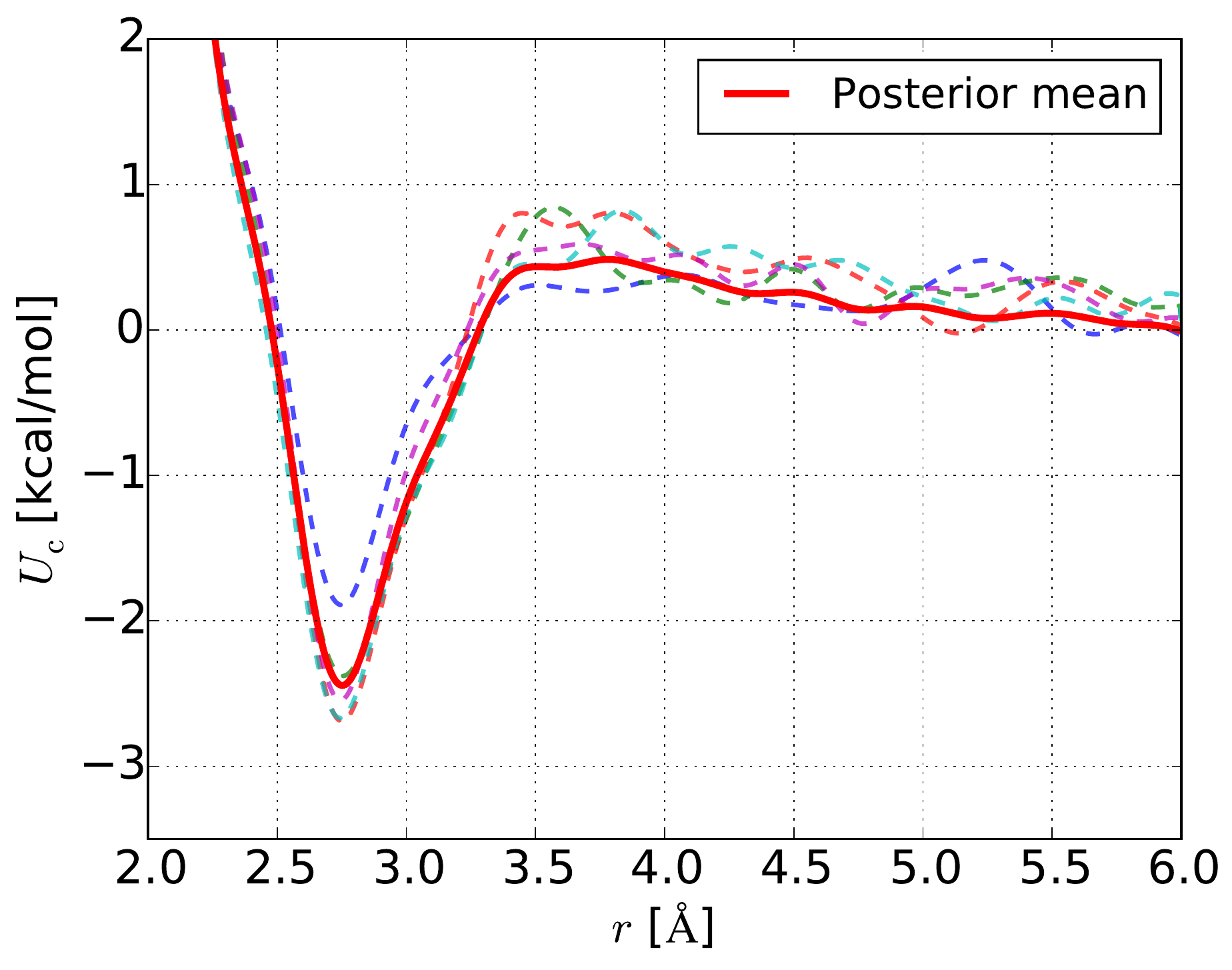}
\label{fig:UcRealizations}
}
\quad
\centering
\subfloat[Posterior mean and credible intervals for $u^{(2)}(R; \bgam\cg)$. We compare this with with the two-body potential computed in~\cite{molinero2014} (dashed blue)  using the relative entropy method.]{\includegraphics[width=0.45\textwidth]{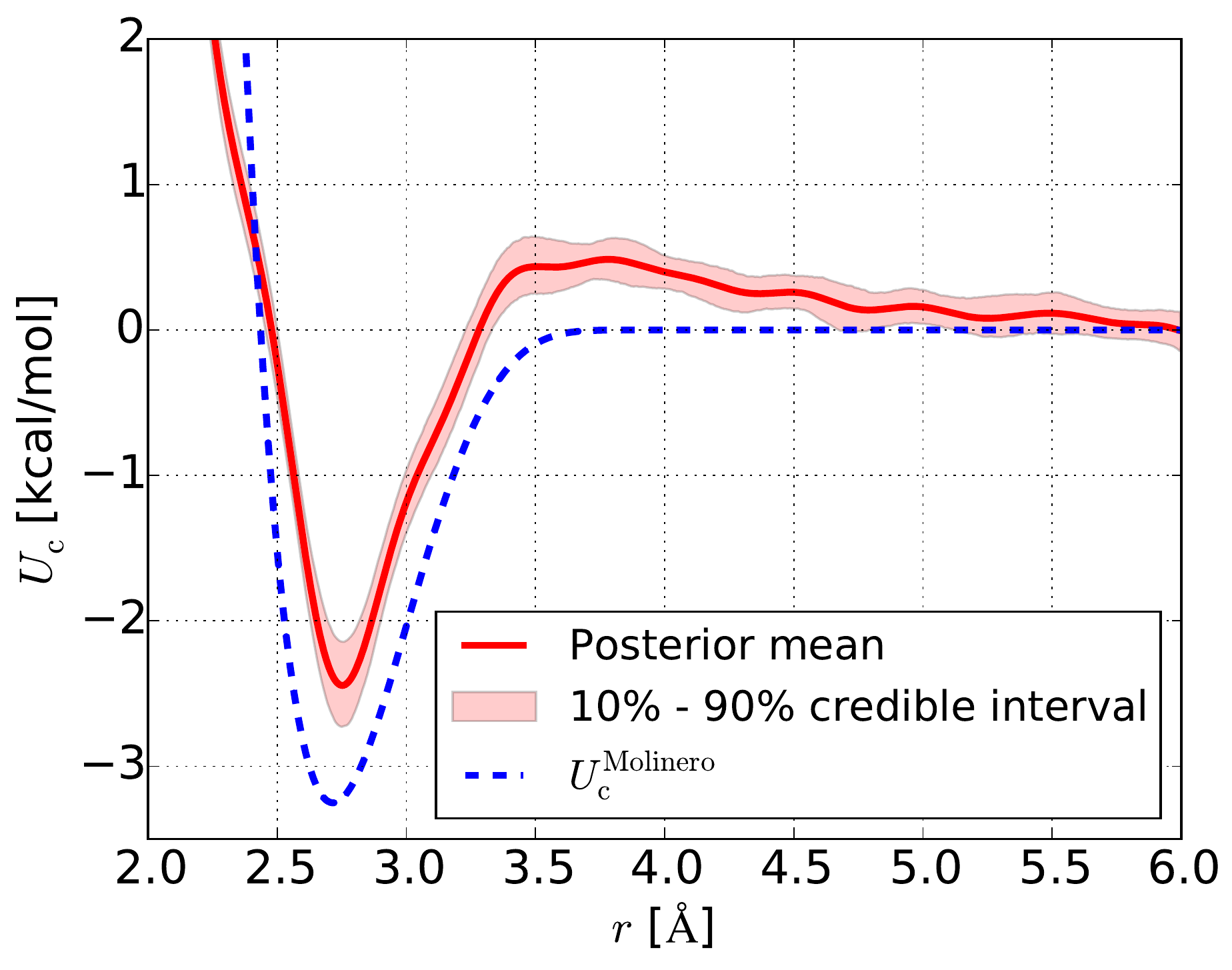}
\label{fig:UcCredible}
}
}{
\caption{Posterior of $u^{(2)}(R; \bgam\cg)$ for $N=20$. }
\label{fig:UcUQ}
}
\end{floatrow}
\end{figure}

Figure~\ref{fig:UcUQ} provides information with regards to the (approximate) posterior of $\bgam\cg$,  computed using the Laplace's approximation proposed, as reflected in the $u^{(2)}(R; \bgam\cg)$.
In particular in \reffig{fig:UcRealizations}, we plot sample realizations of $u^{(2)}(R; \bgam\cg)$ corresponding to different samples of $\bgam\cg$ from the (approximate) Gaussian posterior (Section \ref{sec::approxBayInf}).  We note that all  realizations suggest the same location for  the minimum of the potential. Variability is observed in the depth of this well as well as in its shape to the right of the minimum.  Figure~\ref{fig:UcCredible} depicts the posterior mean of $u^{(2)}(R; \bgam\cg)$ as well as credible intervals at $10\%$ and $90\%$ posterior  quantiles which reflect the inferential uncertainties discussed. 
% 
% Approximating the posterior distribution on the
% potential's parametrization $p(\bgam\cg|\bqi)$ and drawing realizations $j$
% with $\bgam\cg^j \sim p(\bgam\cg|\bqi)$, we show in Figure \ref{fig:UcRealizations}
% the corresponding realization for the coarse potential $U(\bQ,\bgam\cg^j)$.
% The 10\%-90\% credible interval of the potential $U\cg(\bq,\bgam\cg)$ is shown in
% Figure \ref{fig:UcCredible}.
% All realizations of the potential $U\cg(\bQ, \bgam\cg^j)$ lead to a very distinct minimum
% of the potential only a low variability in its corresponding pair distance $r$ can be observed.
% Whereas the gradient $\frac{\partial U\cg(r(\bQ), \bgam\cg^j)}{\parital r(\bQ)}$
% or the corresponding intermolecular force varies around the distinct minimum.

\begin{figure}
\begin{minipage}{.5\linewidth}
\centering
\subfloat[$N=10$]{\label{fig:rdf_10dp}\includegraphics[width=\textwidth]{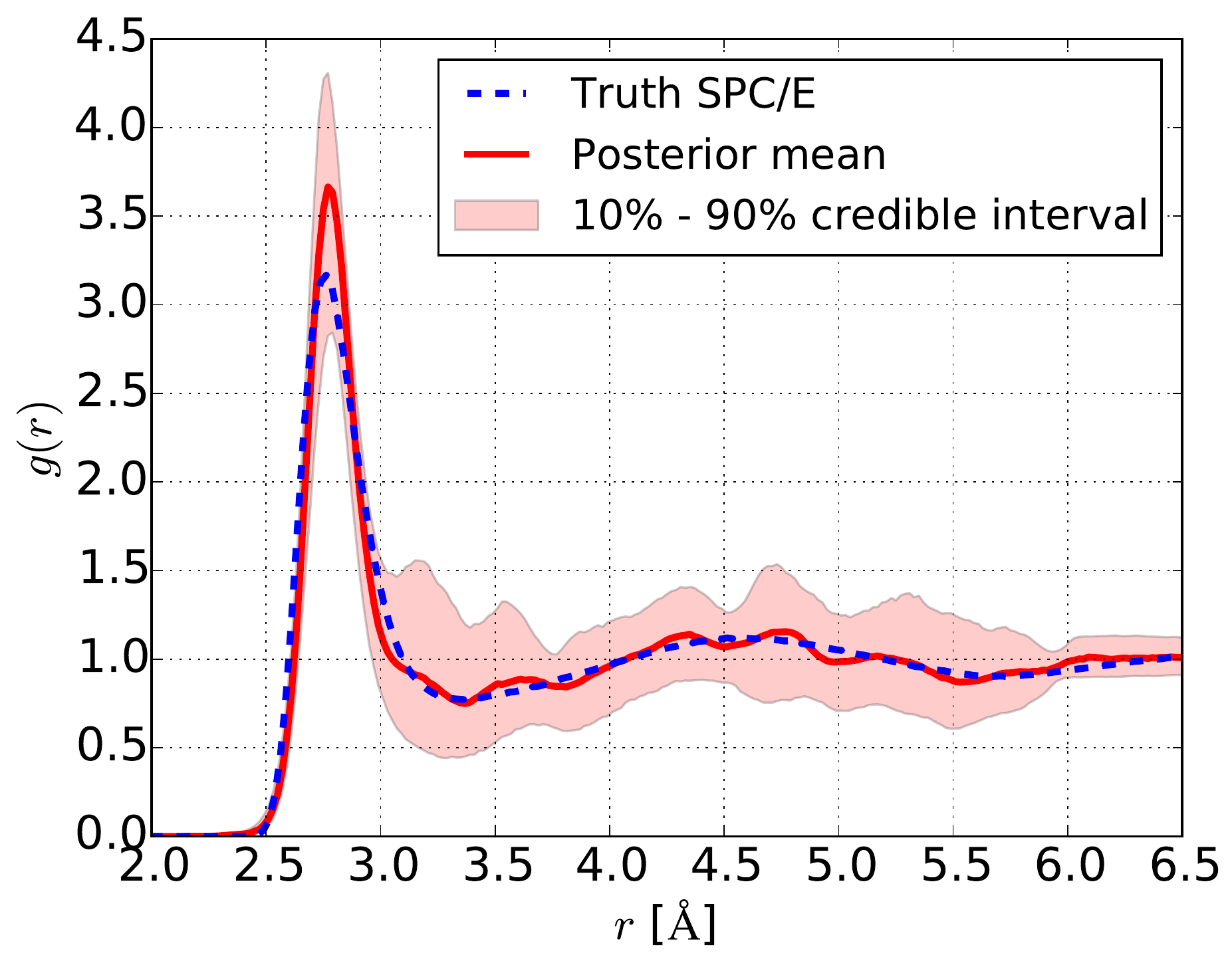}}
\end{minipage}%
\begin{minipage}{.5\linewidth}
\centering
\subfloat[$N=20$]{\label{fig:rdf_20dp}\includegraphics[width=\textwidth]{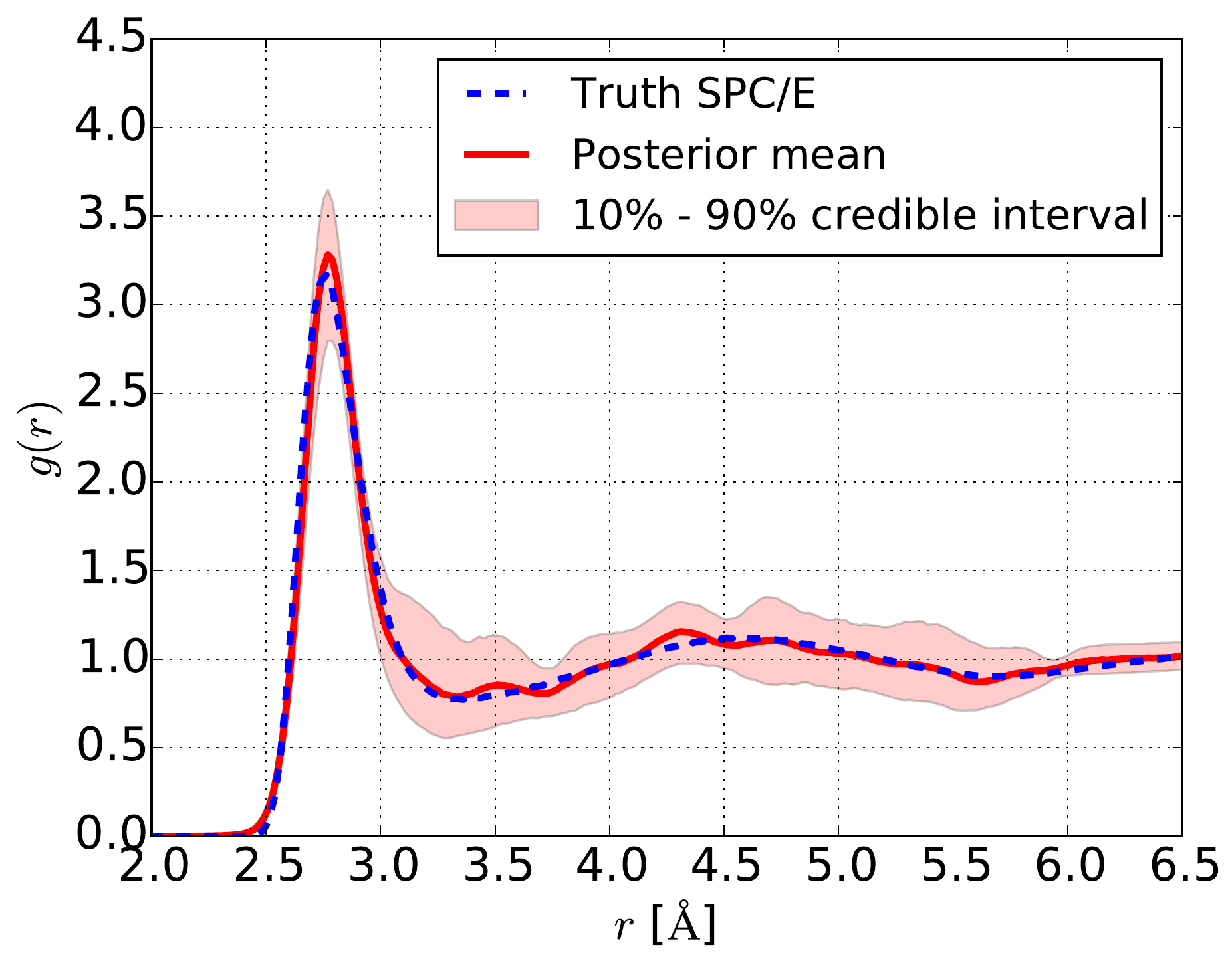}}
\end{minipage}\par\medskip
\centering
\subfloat[$N=100$]{\label{fig:rdf_100dp}\includegraphics[width=0.5\textwidth]{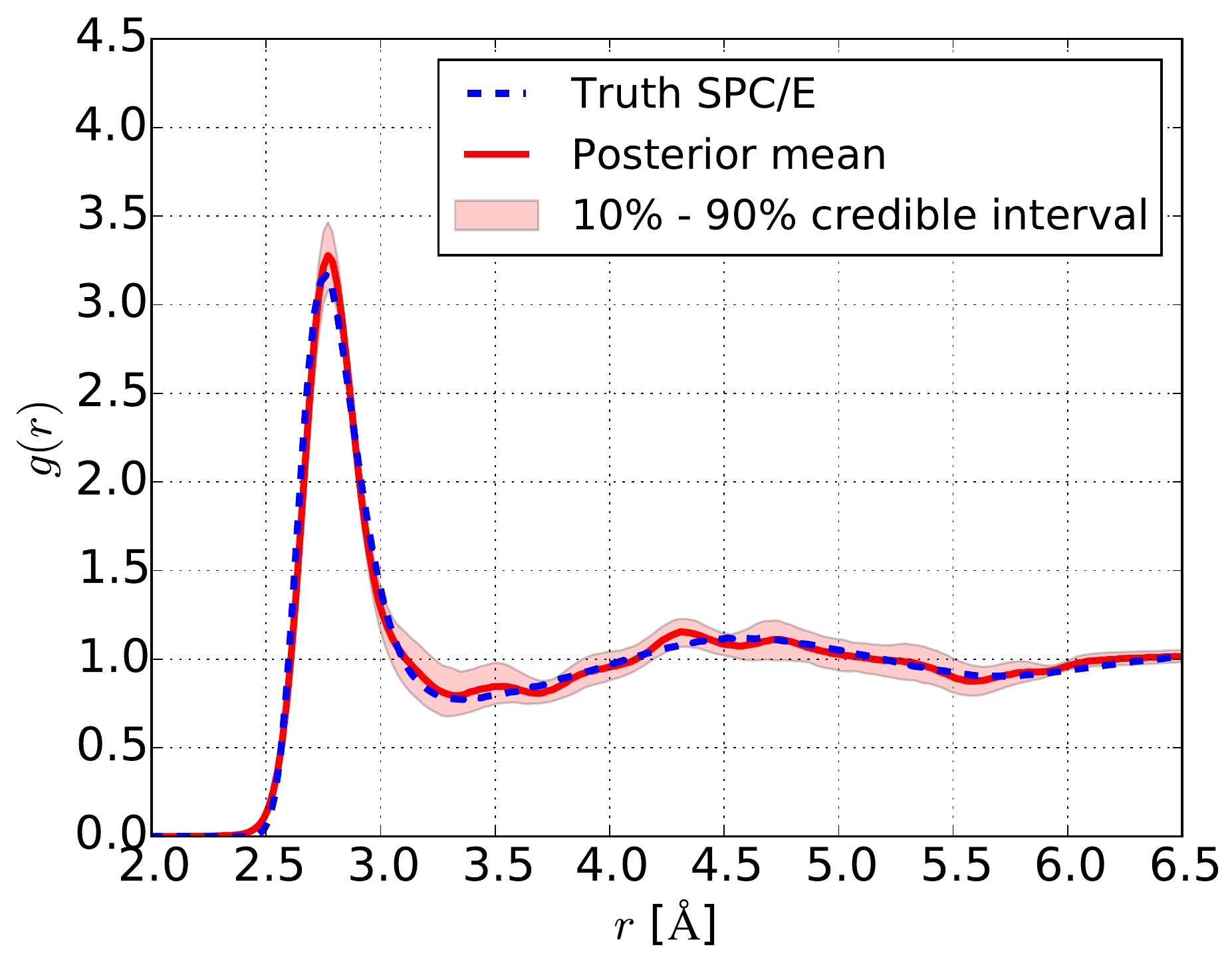}}
\caption{Comparison of the reference RDF $g(r)$ (computed with all-atom simulations using the SPC/E model) with posterior mean and credible intervals corresponding to $10\%$ and $90\%$ posterior quantiles. }
\label{fig:rdf_ndata}
\end{figure}

\begin{figure}
\begin{minipage}{.5\linewidth}
\centering
\subfloat[$N=10$]{\label{fig:adf_10dp}\includegraphics[width=\textwidth]{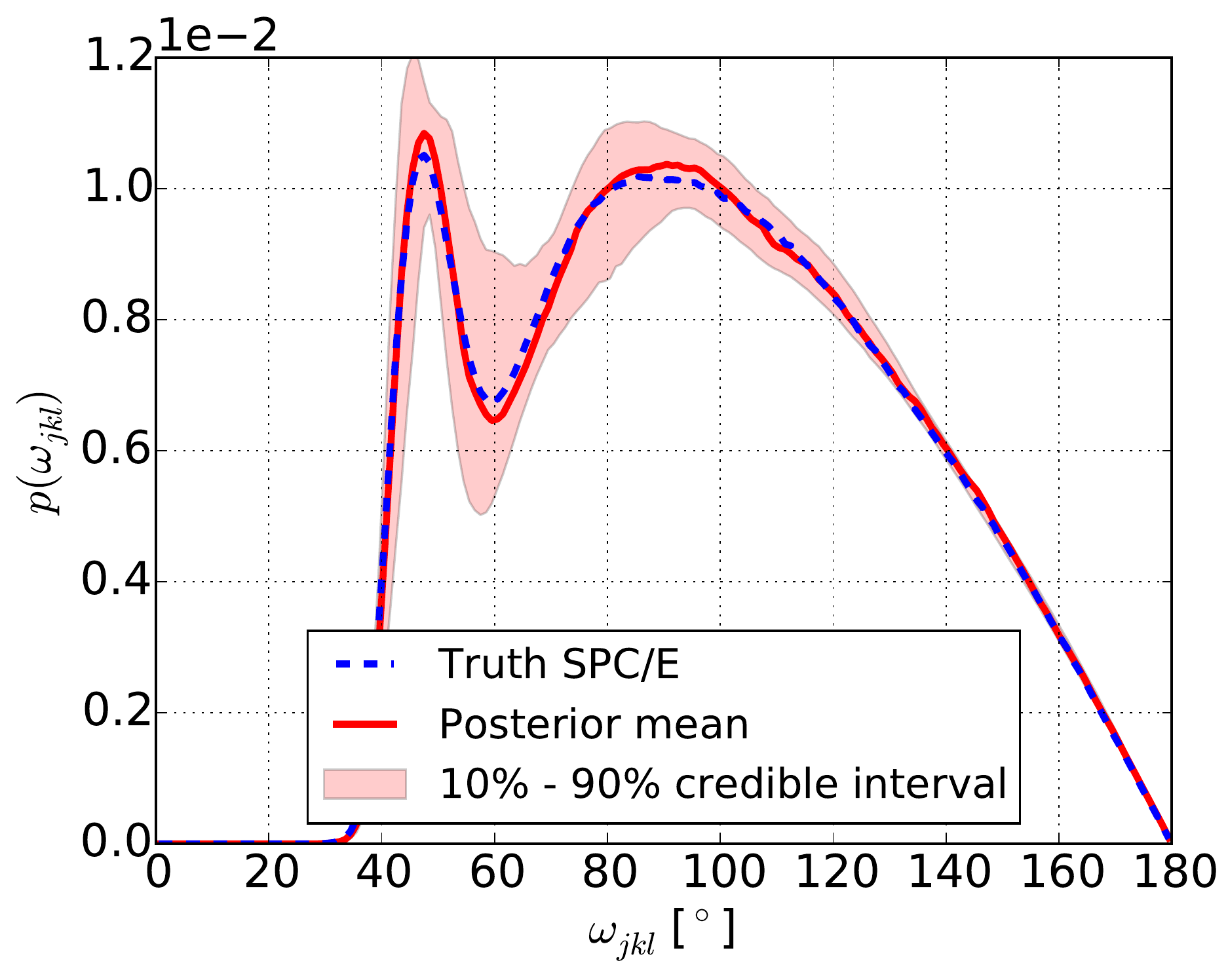}}
\end{minipage}%
\begin{minipage}{.5\linewidth}
\centering
\subfloat[$N=20$]{\label{fig:adf_20dp}\includegraphics[width=\textwidth]{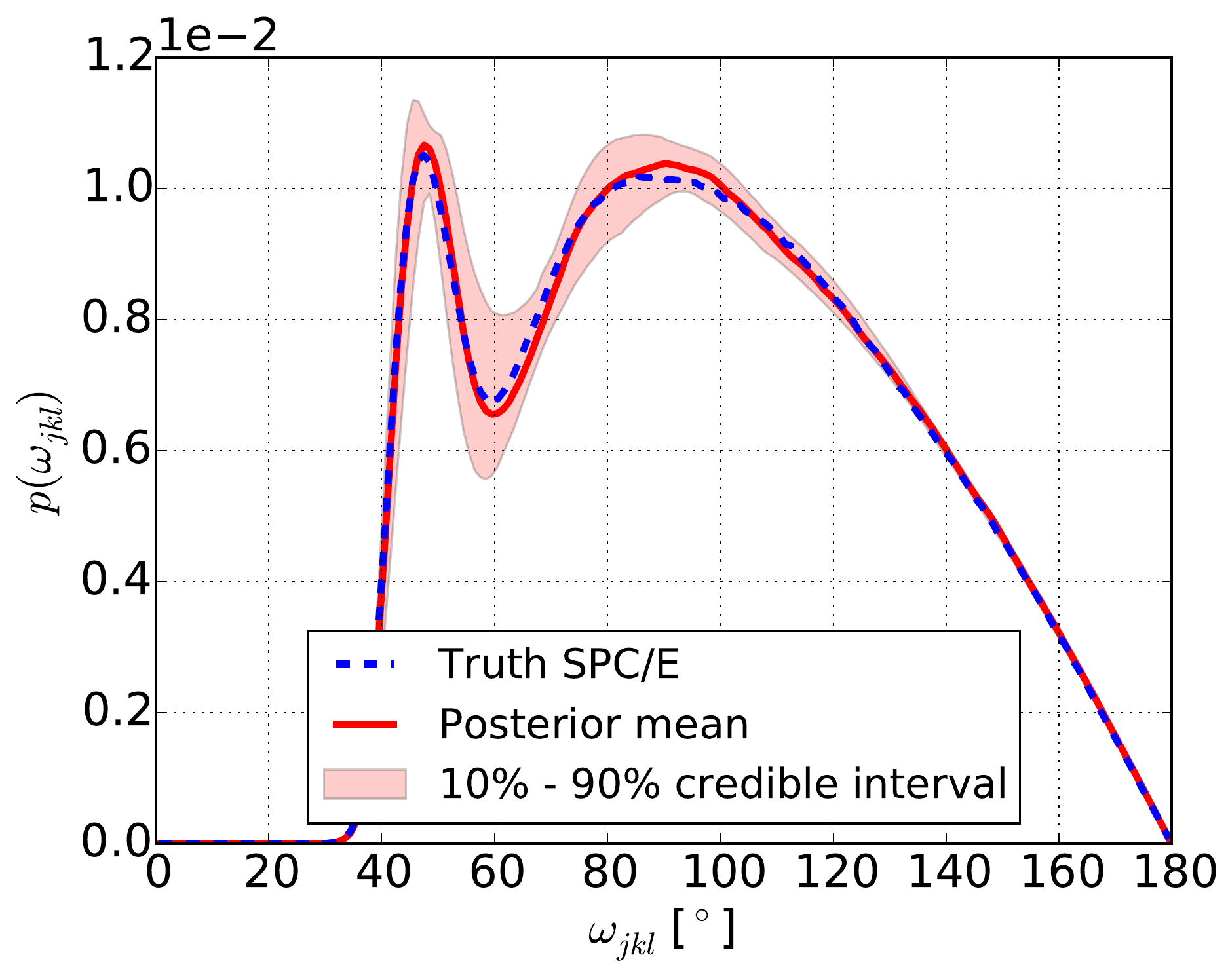}}
\end{minipage}\par\medskip
\centering
\subfloat[$N=100$]{\label{fig:adf_100dp}\includegraphics[width=0.5\textwidth]{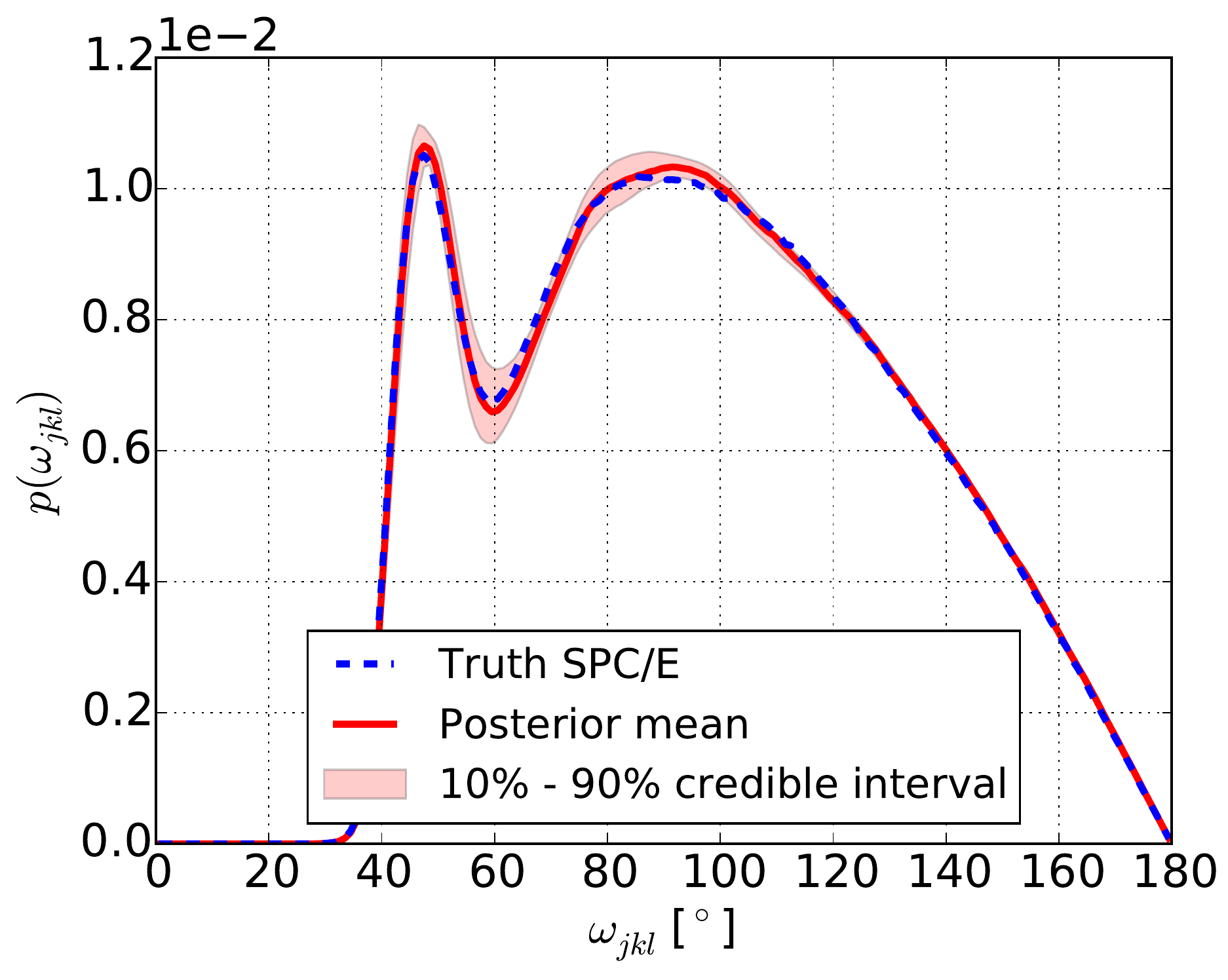}}
\caption{Comparison of the reference ADF $p(\omega)$ (computed with all-atom simulations using the SPC/E model) with posterior mean and credible intervals corresponding to $10\%$ and $90\%$ posterior quantiles. }
\label{fig:adf_ndata}
\end{figure}

We finally report results  illustrating the predictive capability of the model in terms of the macroscopic observables of interest i.e. the RDF and the angular distribution function discussed previously.
To that end, we consider three data settings with $N=10,20$ and $100$ fine-scale (all-atom) training data. While the MAP estimates do not exhibit prominent differences,
the advantage of the method proposed is the predictive posterior
that is furnished (\refeqq{eqn:propPrediction}) and quantifies the uncertainty in the predictions that the coarse-grained model produces.
Figures~\ref{fig:rdf_ndata}  and~\ref{fig:adf_ndata} depict the posterior means and credible intervals corresponding to $10\%$ and $90\%$ posterior quantiles for  the RDF $g(r)$
(i.e. the expected value of the observable in \refeqq{eq:obsrdf})   and the angular distribution function $p(\omega)$
(i.e. the expected value of the   observable in \refeqq{eq:obstetra}). In all cases, the posterior means are very close to the reference values obtained by simulating the all-atom SPC/E model.
It is interesting to point out that when only $N=10$ data were used, the posterior mean overestimates the first peak in the RDF (\reffig{fig:rdf_10dp}). Nevertheless the true solution is contained within the credible intervals computed. As one would expect, the breath of the credible intervals decreases as more training data $N$ is introduced, reflecting the reduction in the predictive uncertainty of the model. Details for the computation of these credible intervals can be found in \ref{app:credInt}.

% \begin{figure}[htbp]
% \centering
% \begin{subfloatrow}
% \ffigbox[\FBwidth]
% \centering
% \subfloat[$N=10$]{\includegraphics[width=0.48\textwidth]{figures/water/three_body/sampling_post_thetaC/10_dp/rdf_theta/plt_rdf.pdf}
% \label{fig:rdf_10dp}
% }
% \quad
% \centering
% \subfloat[$N=20$]{\includegraphics[width=0.48\textwidth]{figures/water/three_body/sampling_post_thetaC/20_dp/rdf_theta/plt_rdf.pdf}
% \label{fig:rdf_20dp}
% }
% \end{subfloatrow}%\hspace*{\columnsep}%
% \\
% \begin{subfloatrow}
% \ffigbox[\FBwidth]
% {
% %\quad
% \centering
% \subfloat[$N=100$]{\includegraphics[width=0.48\textwidth]{figures/water/three_body/sampling_post_thetaC/100_dp/rdf_theta/plt_rdf.pdf}
% \label{fig:rdf_30dp}
% }
% }{
% \caption{Comparison of the reference RDF $g(r)$ (computed with all-atom simulations using the SPC/E model) with posterior mean and credible intervals corresponding to $10\%$ and $90\%$ posterior quantiles. }
% \label{fig:rdf_ndata}
% }
% \end{subfloatrow}
% \end{figure}

% placement from
% http://tex.stackexchange.com/questions/64858/how-to-create-subfloat-figures-two-in-first-row-and-one-below

\FloatBarrier
\section{Conclusions}
\label{sec:summary}

We presented  a novel, data-driven coarse-graining scheme of atomistic ensembles in equilibrium. In contrast to existing techniques which  are based on a restriction, fine-to-coarse map, we adopt the opposite strategy by prescribing a {\em probabilistic coarse-to-fine} map.
This corresponds to a directed probabilistic model where the coarse variables play the role of latent generators of the fine scale (all-atom) data.  Such a model can readily quantify the  uncertainty due to the information loss that unavoidably occurs during the CG process.
We showed that  from an information-theoretic perspective, the  framework proposed broadens  the  relative entropy method. Furthermore, it can be readily extended to a  fully Bayesian model where various sources of uncertainties   are reflected in the posterior of the model parameters.
The latter can be used to produce not only point estimates of fine-scale reconstructions or macroscopic observables, but more importantly, predictive posterior distributions on these quantities. We show how these can quantify the confidence of the model as a function of the amount of data and the level of coarse-graining, i.e. the contrast in the dimension between fine and coarse descriptions.

A critical issue in all CG methods pertains to the form of the coarse model or coarse potential. On one hand, it is desirable   to introduce not only as many feature functions as possible  but also to capture interactions  of the highest-order possible.
On the other hand, such an intricate representation leads to a large number of unknown parameters, augmented computational cost and an increased possibility of overfitting. Such challenges can be readily addressed within the Bayesian framework adopted by the incorporation of appropriate prior models that promote the discovery of sparse solutions and are capable of revealing the most dominant features in the coarse potential. We demonstrated how such a hierarchical prior model, namely the ARD,  is capable of distinguishing the most prominent  feature functions.

The computational engine of the proposed framework is based on an MC-EM scheme that alternates between expectations with respect to the posterior of the latent variables and maximization with respect to the model parameters. This leads to MAP estimates of the model parameters which serve as the basis for the Laplace's model that approximates their posterior. We note that this represents
a very basic approximation that we intend to extend by exploiting advanced MCMC schemes~\cite{liang_double_2010} and/or variational inference schemes~\cite{wainwright_graphical_2008}. From a practical point of view, we note that the algorithm proposed is embarrassingly parallelizable with regards to the expectation step (which  is also the most expensive) and incremental variants  can be readily adopted leading to improvements in computational efficiency.

The generative definition of the CG variables through a probabilistic coarse-to-fine map allows for great flexibility in the type and number of CG variables used. For example in~\cite{dama2013}, the FG configuration space is partitioned and within each of these subdomains 
a different set of CG variables and CG models is learned. This is a reasonable strategy not only because  a {\em globally-}good set of CG variables is difficult to find, but also because the local CG variables  can be lower-dimensional as they need only to work on a limited subdomain.
In the context of the directed, probabilistic model
advocated, the same effect
can be readily achieved by using a mixture model~\cite{bishop_bayesian_2003}.
Consider for example augmenting the set of (latent) CG variables with a  discrete-valued variable,
$S$ which can take values between $1$ and $L$ (which is the number of partitions).
The (latent) variable $S$ characterizes a finite number of discrete states of the system. Depending on the value $S$ takes, the number and type of CG variables $\bQ$ can change by affecting the
two distributions making up the mode, i.e:
 \be
 p\cg(\bQ, S=s | \bgam\cg)= p\cg( \bQ |   \bgam\cg^s) p\cg(S=s),
 \ee
 where each $ p\cg( \bQ |   \bgam\cg^s)$ can be of the same or different form
 (e.g. exponential family) but with different parametrizations $\bgam\cg^s, s=1,\ldots L$.
 Similarly for the coarse-to-fine map, we can define:
 \be
 p\cf(\bq | \bQ, S=s, \bgam\cf)=p\cf(\bq | \bQ,  \bgam\cf^s),
 \ee 
 where again the parametrization can depend or not on $S$, $\bgam\cf^s, s=1,\ldots L$.
 Infinite  mixture models~\cite{antoniak_mixtures_1974,rasmussen_infinite_1999,Chen2015uq} based on Dirichlet process priors  could provide a rigorous strategy on determining the number $L$ of such hidden states needed to describe the atomistic ensemble.
 We note finally that, in nonequilibrium settings, by appropriate modeling of the time dependence of $S$ one would recover  Hidden Markov Models (HMM,~\cite{cappe_inference_2005}) which have been employed in coarse-graining frameworks~\cite{fischer_identification_2007,horenko_data-based_2007}.

 Another potentially powerful extension, involves the use of deep, hierarchical models. Deep learning tools have revolutionized various machine learning tasks~\cite{lecun_deep_2015} by stacking multiple layers of simple representations. In the context of coarse-graining, such a scheme could be materialized by augmenting the set of CG variables as $\bQ_1,\bQ_2, \ldots \bQ_L$ and the CG model as:
 \be
 p\cg(\bQ_1,\bQ_2, \ldots \bQ_L)= p_{\mathrm{c},1}(\bQ_1 | \bQ_2, \bgam\cg^{1}) \ldots p_{\mathrm{c},L-1}(\bQ_{L-1} | \bQ_L, \bgam\cg^{L-1}) p_{\mathrm{c},L}(\bQ_L | \bgam\cg^{L}).
 \ee
 If $\dim(\bQ_1) > \dim(\bQ_2)>\ldots > \dim(\bQ_L)$, then such a structure could provide a hierarchical decomposition of the CG picture, starting from a highly coarse description and gradually reaching the more detailed abstraction $\bQ_1$. 
 The coarse-to-fine map could be controlled by $\bQ_1$ as $p\cf(\bq | \bQ_1, \bgam\cg)$.
 
%  A final challenge involve sthe determination of the number of CG variables $\bQ$. Such a problem should unavoidably involve the $\bgam\cf$.

\section{Acknowledgments}
\noindent We acknowledge the support by the Hans Fisher Senior Fellowship
of Nicholas Zabaras of the Technical University of Munich -- Institute for Advanced Study,
funded by the German Excellence Initiative and the European
Union Seventh Framework Programme under grant agreement No. 291763.
Nicholas Zabaras also acknowledges support from the Computer Science
and Mathematics Division of ORNL under the DARPA EQUiPS program.

\begin{appendix}

\section{Methodology}

\subsection{Estimating credible intervals}
\label{app:credInt}
\noindent
This note summarizes necessary steps for estimating credible intervals.
The   Bayesian inference algorithms described  in Sections \ref{sec:inference_learning}
and  \ref{sec::approxBayInf}, lead to (Gaussian) approximations of the posterior  $p(\btheta|\bqd)$ (\refeqq{eq:post}).
% with,
% \be
% p(\btheta|\bqd) \propto \prod_{i=1}^N \left(\int_{} p\cf(\bq^{(i)} |\bQ^{(i)},\bgam\cf )~ p\cg(\bQ^{(i)} | \bgam\cg) d \bQ^{(i)} \right) ~p(\btheta).
% \ee
The credible intervals shown in Figs.~\ref{fig:ising_mag_lcg},~\ref{fig:ising_corr_lcg},~\ref{fig:rdf_ndata}, and~\ref{fig:adf_ndata} are constructed from Monte Carlo samples $\hat{a}(\btheta^{(i)} )$ of the observables of interest. These are generated on the basis of  \refeqq{eqn:propPrediction} as follows:
\begin{algorithm}[H]
\caption{\small Estimating Credible Intervals}
\begin{algorithmic}[1]
%\algsetup{linenosize=\tiny}
\small
\FOR{all $i=1,\dots,I$}
\STATE Obtain a posterior sample: $\btheta^{(i)} \sim p(\btheta|\bqd)$ (\refeqq{eq:post}).
\STATE Calculate the predictive estimate $\hat a(\btheta^{(i)})$ shown in \refeqq{eqn:propPrediction}:
 \be
 \hat{a}(\btheta^{(i)} ) = \left( \int a(\bq)~p\cf(\bq | \bQ, \bgam\cf^{(i)})~p\cg(\bQ |  \bgam\cg^{(i)})~d\bQ ~ d\bq \right).
 \ee
 The integrations  involved are performed with Monte Carlo sampling. We note that this requires simulating only the CG model as the mapping implied by $p_{cf}$ is straightforward. 
\ENDFOR
\STATE Compute desired quantiles with the given samples $\hat a(\btheta^{(1 \dots I)})$.
\end{algorithmic}
\end{algorithm}
\noindent
We note that the estimated quantiles of the corresponding predictive posterior are not
necessarily symmetric around its MAP estimate  $\hat a(\btheta_{\mathrm{MAP}})$, even in the
case of a symmetric posterior of the model's parameters $p(\bgam|\bqd)$ (\refeqq{eq:post}).

% \begin{enumerate}
%  \item Obtain a posterior sample: $\btheta^i \sim p(\btheta|\bqd)$.
%  \item Calculate the predictive estimate $\hat a(\btheta^i)$ shown in \refeqq{eqn:propPrediction}
%  with,
%  \be
%  \hat{a}(\btheta^i ) = \left( \int a(\bq)~p\cf(\bq | \bQ, \bgam\cf^i)~p\cg(\bQ |  \bgam\cg^i)~d\bQ ~ d\bq \right).
%  \ee
%  \item Repeat 1. and 2. until suffice amount of estimates reached.
%  \item Estimate desired quantiles from the given samples $\hat a(\btheta^i)$.
% \end{enumerate}

\subsection{Comparison of gradients between relative entropy method and PCG}
\label{app:compare_gradients_relEntr}
\noindent
This section compares the gradients with respect to the parameters of the coarse potential  $\bgam\cg$, between the proposed scheme and the relative entropy method. These are used for fitting the model parameters $\bgam\cg$. In our case, the gradient is given by:
\be
  \frac{\pa \mathcal{F} }{ \pa \theta_{c,k} }= {  \sum_{i=1}^N }~   \left( <\phi_k(\bQ^{(i)})>_{ {\color{red} q_i(\bQ^{(i)}) } } -<\phi_k(\bQ)>_{ \pc(\bQ|\bgam\cg) }  \right) \\,
  \label{eqn:app_grad_predCG}
\ee
whereas for the relative entropy method (when the objective $\mathcal{F}_{\mathrm{KL}}$ is 
given as in~\refeqq{eq:klre}):
%\be
\begin{align}%{ll}
  \frac{\pa \mathcal{F}_{\mathrm{KL}} }{ \pa \theta_{c,k} } & = \left( <\phi_k(\calR(\bq))>_{ {\color{red} p\aaa(\bq) } } -<\phi_k(\bQ)>_{ \pc(\bQ|\bgam\cg) }  \right) \nonumber \\
  & \approx \frac{1}{N} \sum_{i=1}^N  \left(  <\phi_k(\calR(\bq^{(i)})) -<\phi_k(\bQ)>_{ \pc(\bQ|\bgam\cg) }   \right).
  \label{eqn:app_grad_relEntr}
\end{align}
%\ee
In the latter case,
the expectations with respect to $p\aaa(\bq)$ are estimated using the fine-scale data
$\bq^{(i)}$ whereas in the former these involve averaging over the {\em posterior}
of the CG variables $\bQ$.
This emphasizes the role of the CG variables play in our model as latent (hidden) generators of the fine-scale.

\subsection{ARD Prior}
\label{app:ard}
\noindent
We adopt the Automatic Relevance Determination (ARD,~\cite{mackay1994}) which is formulated in the context of
hierarchical Bayesian models. The prior on the parameters $\btheta\cg$ is modeled as independent Gaussian for each $\theta_{\mathrm{c}, k}$
with zero mean and precision hyper-parameter $\tau_k$:
\be
p(\btheta\cg|\bs{\tau}) \equiv  \prod_k  \underbrace{\mathcal{N}(\theta_{\mathrm{c}, k}|0,\tau_k^{-1})}_{p(\theta_{\mathrm{c}, k}|\tau_k)}.
\ee
The precision (hyper-)parameters $\tau_k$ follow a Gamma distribution,
\be
 \tau_k \sim  Gamma( \tau_k |a_0,b_0).
\ee
Anytime derivatives of the log-prior are needed,  an  inner-loop Expectation-Maximization scheme can be employed which is based on the same ideas presented previously. In particular, for  any set of densities $q_k(\tau_k)$ we can obtain a lower bound on the the log-prior as follows :
%\be
\begin{align}%{ll}
 \log p( \btheta\cg ) &= \log \left( \prod_k \int p(\theta_{\mathrm{c}, k}|\tau_k) ~p(\tau_k|a_0,b_0) ~d\tau_k \right) \nonumber \\
 &= \sum_k \log \int q_k(\tau_k) \frac{p(\theta_{\mathrm{c}, k}|\tau_k) ~p(\tau_k|a_0,b_0)}{q(\tau_k)} ~d\tau_k \nonumber \\
 &\geq \sum_k \int q_k(\tau_k) \log \frac{p(\theta_{\mathrm{c}, k}|\tau_k) ~p(\tau_k|a_0,b_0)}{q_k(\tau_k)} ~d\tau_k \qquad \textrm{(Jensen's inequality)} 
%  &= \sum_k f^{(k)} (q_k(\tau_k), \theta_k). \\
\label{eq:emARD}
\end{align}
%\ee
 The optimal $q_k$ i.e. the posteriors $p(\tau_k|\theta_{\mathrm{c},k})$ (for which the lower bound becomes tight) can be analytically computed and are Gamma densities with parameters $a_k=a_0+\frac{1}{2},~b_k=b_0+\frac{\theta_{\mathrm{c},k}^2}{2}$ \cite{bishop_variational_2000},  where the current values of $\theta_{\mathrm{c},k}$'s are used.
This leads to the extremely simple iterations of the following form  \cite{bishop_variational_2000}:
\begin{itemize}
 \item E-step: evaluate:
 \be \left\langle \tau_k \right\rangle_{p(\tau_k|\theta_{\mathrm{c},k})} =\frac{a_k}{b_k}=
 \frac{ a_0 + \frac{1}{2} }{ b_0 + \frac{\theta_{\mathrm{c},k}^2}{2} }.
 \ee
 \item M-step: evaluate:
%\be
\begin{align}%{ll}
\frac{\partial \log p(\bgam\cg) }{\partial \theta_{\mathrm{c},k} } &= \frac{\partial}{\partial \theta_{\mathrm{c},k} } \int q_k(\tau_k) \log p(\theta_{\mathrm{c},k}|\tau_k) ~d\tau_k \nonumber \\
 &= - \int q_k(\tau_k) \tau_k ~d\tau_k ~\theta_{\mathrm{c},k} \nonumber \\
 &= - \left\langle \tau_k \right\rangle_{p(\tau_k|\theta_{\mathrm{c},k})}  \theta_{\mathrm{c},k}.
\label{eq:emARD-M}
\end{align}
%\ee
\end{itemize}

\section{Numerical Examples}
% \subsection{SPC/E Water}
% \label{app:spceProd}
% The molecular dynamics software package used in this publication is LAMMPS \cite{plimpton1995}.
% It is freely distributed under \url{http://lammps.sandia.gov}.
\subsection{SPC/E model, parameters and simulation details}
\label{app:spceProd}
\noindent
The following SPC/E parameters as given in~\cite{bilionis2013,kremer2009}
are used for producing the fine-scale data.
\begin{itemize}
 \item LJ-potential: $\sigma = 3.166\,\text{\AA}$, $\epsilon = 0.650\,\frac{\text{kJ}}{\text{mol}}$.
 \item Electrostatic load: $q_\mathrm{H} = +0.4238\,e$, $q_\mathrm{O} = -0.8476\,e$.
 \item Structural properties of rigid water model: bond-legnth $l_{\mathrm{OH}}=1.0\,\text{\AA}$ and
 bond-angle $\theta_{\mathrm{HOH}} = 109.47^\circ$.
 \item Masses:
 $m_\mathrm{O} = 15.994\,\frac{\text{g}}{\text{mol}}$ and
 $m_\mathrm{H} = 1.00794\,\frac{\text{g}}{\text{mol}}$.
\end{itemize}

\subsubsection{Simulation steps}
\noindent
In this work, we consider a system of $N_w=100$ water molecules at a temperature $T= 300\,\text{K}$.
The following steps for obtaining training data are performed:
\begin{enumerate}
 \item NPT simulation with $p=1\,\text{bar}$ and a timestep of $\Delta t = 2.0\,\text{fs}$. Simulate the system for $t=100\,\text{ns}$.
 \item Use last $t=80\,\text{ns}$ for calculating the equilibrium box size.
 We found $l_{\text{box}} = 14.5459665\,\text{\AA}.$
 \item Fix the box length to the one obtained from previous step. Simulate
 system in NVT ensemble for $t=45\,\text{ns}$ with a timestep of $\Delta t = 2.0\,\text{fs}$.
 Use the last $t=40\,\text{ns}$ and write the trajectory every $200$ steps.
\end{enumerate}

\subsection{Radial Distribution Function}
\label{app:spceRDF}
\noindent
The radial distribution function $g(r)$ is defined by,
\[
 g(r) =  \left\langle \frac{V}{N^2} a^{\text{RDF}}(r) \right\rangle.
\]
The discrete version follows with the number of bins $n_{\mathrm{bin}}$ and a bin size $\Delta r$:
\[
 g(r_1) = \frac{1}{N n_{\mathrm{bin}}} \frac{\left\langle  a^{\text{RDF}}(r_1) \right\rangle}{\rho_{\mathrm{ideal}}},
\]
with,
\[
 \rho_{\mathrm{ideal}} = N/V,
\]
\[
  a^{\text{RDF}}(r_1) = \frac{ n(r_1) }{ \Delta V } = \frac{\sum_{ij} \int_{r_1}^{r_1 + \Delta r} \delta(r_{ij} - r) dr }
 {\frac{4}{3} \pi ( (r_1 + \Delta r)^3 - r_1^3)}.
\]

\subsection{Stillinger-Weber (SW) potential}
\label{app:sw}
\noindent
The Stillinger-Weber (SW) potential originally proposed in~\cite{stillinger1985}  and extended in~\cite{molinero2014}, contained both two- and three-body interactions.
In this work, we make use only of the latter three-body contribution: 
% \paragraph{Three-Body Interactions}
% Not all properties as for example the tetrahedral structure of water is
% well predicted by two-body interactions \cite{kremer2009, molinero2014, shell2014}.
% Therefore it is of interest to explore capabilities of using a potential with 
% second and third order contributions in combination.
% The Stillinger-Weber (SW) potential introduced in \cite{stillinger1985}
% has contributions from second- and third order interactions with,
\be
 U^{\mathrm{SW}}(\bQ) = 
\sum_j \sum_{k \neq j} \sum_{l>k}
 \phi_3^{\mathrm{SW}}(r_{jk}, R_{jl}, \omega_{jkl} ),
\ee
where the three-body term $\phi_3^{\mathrm{SW}}(r_{jk}, r_{jl}, \omega_{jkl} )$ is given by:
 \be
\begin{array}{ll} 
 \phi_3^{\mathrm{SW}}(r_{jk}, r_{jl}, \omega_{jkl} ) &= \lambda \epsilon \left[ \cos\omega_{jkl} - \cos\omega_0  \right]^2
 \exp\left( \frac{\gamma \sigma}{r_{jk} - a_3 \sigma_{\mathrm{SW}}} \right)
 \exp\left( \frac{\gamma \sigma}{r_{jl} - a_3 \sigma_{\mathrm{SW}}} \right),
\end{array}
\label{eqn:molinero_threeB}
\ee
with $r_{jk}$ being the pairwise distances between molecules $j$ and $k$ and $\omega_{jkl}$ is the angle between molecules $j,k,l$.
The following values for the parameters were used \cite{molinero2014}:  $\lambda = 0.762$, $\epsilon = 83.5737$, $\cos \omega_0 = -0.487217$,
$\gamma = 0.291321$, $a_3 = 0.586097$, $\sigma_{\mathrm{SW}} = 6.4144$.

\end{appendix}

%\section*{References}

%\bibliography{ref,my_zote_zero_library}
\bibliography{predictive_cg}

\end{document}